\documentclass{article}
\usepackage{arxiv}

\usepackage[T1]{fontenc}
\usepackage[utf8]{inputenc}
\usepackage{amsmath}
\usepackage{amssymb}
\usepackage{amsfonts}
\usepackage{graphicx}
\graphicspath{{figures/}{./}}
\usepackage{booktabs}
\usepackage{array}
\usepackage{makecell}
\usepackage{multirow}
\usepackage[usenames,dvipsnames]{xcolor}
\usepackage{colortbl}
\usepackage{float}
\usepackage[version=3]{mhchem}
\usepackage{algorithm}
\usepackage{algpseudocode}
\usepackage{enumitem}
\usepackage{xspace}
\usepackage{caption}
\usepackage[numbers,sort&compress]{natbib}
\usepackage{hyperref}
\usepackage{url}

\newcommand{\pas}[1]{\mathcal{P}_{#1}}
\newcommand{\pasj}{\mathcal{P}_{\text{joint}}}

\newcommand{\bopas}[1]{\hat{\mathcal{P}}_{#1}}
\newcommand{\bopasj}{\hat{\mathcal{P}}_{\text{joint}}}
\newcommand{\paspred}{\mathcal{P}_{\text{pred}}}
\newcommand{\pastrue}{\mathcal{P}_{\text{true}}}
\newcommand{\params}{\mathbf{X}}
\newcommand{\point}{\mathbf{x}}
\newcommand{\hpoint}{\hat{\mathbf{x}}}
\newcommand{\pointnext}{\mathbf{x}^*}
\newcommand{\ucb}{\text{UCB}}
\newcommand{\str}{\text{STR}}
\newcommand{\rstr}{\text{STR}^*}
\newcommand{\jarex}{A_{\text{jarex}}}
\newcommand{\fei}{\text{EI}_\mathrm{feas}}
\newcommand{\obsd}{\textit{obsidian}\xspace}

\title{JAREX: An Acquisition Function for Multi-Objective Algorithmic Process Characterization}

\usepackage{authblk}

\author[1]{Xinyang Li\thanks{\texttt{xinyang.li@merck.com}}}
\author[1]{Kevin Stone}
\author[1]{Ajit Vikram\thanks{\texttt{ajit.vikram@merck.com}}}
\affil[1]{Pharmaceutical Analysis \& Digital Technologies, Merck \& Co., Inc., Rahway, New Jersey, 07065, United States}
\date{}
\renewcommand{\shorttitle}{JAREX for Multi-Objective Process Characterization}

\hypersetup{
  pdftitle={JAREX: An Acquisition Function for Multi-Objective Algorithmic Process Characterization},
  pdfauthor={Xinyang Li, Kevin Stone, Ajit Vikram},
}

\begin{document}
\maketitle

\begin{abstract}
        Pharmaceutical process characterization is central to Quality by Design because it defines how variations in process parameters affect the ability to meet product quality specifications, thereby supporting proven acceptable ranges and robust manufacturing.
        In practice, however, characterization still relies largely on factorial design of experiments (DOE) approaches, which are inefficient for resolving multivariate pass/fail boundaries in higher-dimensional spaces.
        While Bayesian optimization has transformed process optimization, adaptive methods for multi-objective process characterization remain lacking.
        Here, we introduce JAREX (Joint Acceptable Region EXploration), a Bayesian active-learning acquisition function for multi-objective process characterization.
        JAREX formulates characterization as a joint boundary-learning problem and adaptively selects experiments to recover the joint pass region defined by simultaneous satisfaction of threshold criteria across multiple objectives.
        JAREX combines an optimistic joint-feasibility mask with a multi-objective extension of randomized straddle, focusing sampling on the joint edge of failure.
        On a simulated kinetic model and two in silico pharmaceutical case studies with up to 13 inputs and 7 objectives, JAREX recovers the joint pass region more accurately than factorial DOE, I-optimal designs, and space-filling sampling, and in most settings than greedy objective-wise strategies.
        For batched experimentation, it reduces the number of sequential experimental rounds by more than half, at the cost of more total experiments.
        Implemented in the open-source obsidian package, JAREX provides a modular framework for adaptive, data-efficient multi-objective algorithmic process characterization, supporting sample-efficient range finding in high-dimensional spaces.
\end{abstract}

\section{Introduction}

Pharmaceutical manufacturing under the Quality-by-Design (QbD) paradigm depends on a quantitative understanding of how process parameters (e.g., temperature, pressure, time, solvent ratio, etc.) impact in-process attributes and critical quality attributes (CQAs) such as identity, purity, and yield.~\cite{ICH2009,ICHQ11,Yu2014,Rathore2009,Lionberger2008,YuKopcha2017,Grangeia2020,amEnde2007}
Two complementary approaches build this understanding: \textit{process optimization} and \textit{process characterization}.
Process optimization identifies optimal operating conditions that maximize desirable outcomes while minimizing undesirable ones.
Process characterization, however, is not about identifying an optimum but about measuring the operating ranges of the process: how CQAs respond across the parameter space and identifying where the \textit{edge of failure} lies within relevant operational windows, beyond which specifications can no longer be met.
In current practice, process characterization most commonly results in establishing proven acceptable ranges (PARs), informed by univariate windows varying one factor at a time (OFAT).~\cite{ICH2009,Yu2014,Rege2024}
PARs alone, however, do not constitute a \textit{design space}, the multidimensional region demonstrated to deliver quality assurance against joint parameter excursions rather than OFAT variations.~\cite{ICH2009,ICHQ11,Peterson2010}
Exploring a design space therefore requires multivariate characterization of the parameter space, which is costly and data-intensive in practice, and this is the problem this work addresses.

Traditionally, process characterization workflows rely heavily on factorial design of experiments (DOE) or other statistical screening designs.~\cite{Montgomery2017,Weissman2015,Rege2024,amEnde2007}
These methods provide systematic but non-adaptive sampling of the parameter space and scale poorly with dimensionality.
Alternatively, adaptive, data-driven strategies that iteratively select informative experiments based on prior observations may be useful to explore design space efficiently.
A useful precedent comes from process optimization, where Bayesian optimization (BO) has emerged as a transformative tool.~\cite{Shahriari2016,Frazier2018}
BO combines probabilistic surrogate models, typically Gaussian process (GP) models, with acquisition functions such as Expected Improvement,~\cite{Jones1998} Upper Confidence Bound,~\cite{Auer2002,Srinivas2010} and Probability of Improvement~\cite{Kushner1964} to balance exploration and exploitation.
Open-source frameworks such as BoTorch~\cite{Balandat2020} have made these methods broadly accessible.
An analogous adaptive paradigm is needed for process characterization, but the gap between theory and practice remains substantial.
While mathematical frameworks for this type of problem do exist,~\cite{Bryan2005,Gotovos2013,Boukouvala2012,Metta2021} their translation into practical, application-ready tools for pharmaceutical process characterization is still limited.
Adaptive identification of regions that satisfy several specifications has been studied in engineering design~\cite{Garbuglia2021,Qing2022} and in multi-objective experimental design.~\cite{Malkomes2021}
To our knowledge, however, no acquisition function has been tailored to multi-objective process characterization (MOPC), even though nearly all pharmaceutical processes involve multiple CQAs simultaneously.

Here, we present JAREX (Joint Acceptable Region EXploration), a new acquisition function designed for characterizing high-dimensional processes with multiple objectives.
JAREX adaptively targets the joint edge of failure, where acceptability is decided by all objectives together, while standard outputs such as PARs and multi-factor interaction plots are recovered from the fitted surrogate via post-processing.
On a simulated kinetic model, JAREX substantially outperforms the other methods in sample efficiency and boundary recovery.
On two pharmaceutical case studies with up to 13 inputs, it is more accurate than designed experiments and space-filling sampling in every setting and more accurate than a greedy objective-wise strategy in most, without the objective order the greedy strategy needs.
The gain is largest where much of each objective's boundary lies outside the joint pass region and the surrogates can recognize those parts.
The method is available in the open-source \obsd\ package~\cite{obsidian} and brings to pharmaceutical process characterization the kind of adaptive data-driven guidance that Bayesian optimization has brought to process optimization.

\section{Methodology}

Process characterization is fundamentally a \textit{global learning} task: the goal is to understand how quality attributes respond across an entire parameter space, with particular emphasis on where they cross their acceptable quality limits.
This differs sharply from optimization, which seeks a single optimal point (or, in multi-objective settings, a Pareto surface).
Consider a process with parameter space $\params \subseteq \mathbb{R}^d$, where $d$ is the number of process parameters, and let $\point \in \params$ denote a specific set of operating conditions.
For each of $m$ quality attributes, we define an objective $O_i(\point)$ for $i = 1, \ldots, m$, together with a threshold $h_i$ that encodes the pass/fail specification.
Without loss of generality, we take ``pass'' to mean $O_i(\point) \ge h_i$, so the pass region for attribute $i$ is
\begin{align}
  \label{eq:pass-region}
  \pas{i} = \{\point \in \params : O_i(\point) \ge h_i\}.
\end{align}
Our characterization task is to recover these pass regions from a surrogate model fitted on a finite sample budget.

There are two natural ways to frame this task algorithmically, and they motivate distinct acquisition strategies.
The first is a \textit{feasibility} perspective.
At every candidate $\point$, query the likelihood of the objective passing, and sample where this likelihood is most informative.
Ierapetritou and coworkers~\cite{Metta2021} took this route by adapting Expected Improvement into a feasibility-oriented acquisition function (feasibility EI) that favors candidates near the threshold with high model uncertainty.
The full definition and interpretation are given in Section~S1.1 of the Supplementary Information.
The second is a \textit{boundary} perspective: rather than reasoning pointwise about feasibility, treat the surface $\{\point : O_i(\point) = h_i\}$ itself, the ``edge of failure'' in the ICH Q8(R2) guideline~\cite{ICH2009}, as the object to recover.
This is precisely the \textit{level set estimation} (LSE) problem studied in applied mathematics.
Two main algorithm families have emerged for LSE: stepwise uncertainty reduction,~\cite{Bect2012,Chevalier2014} which reduces global uncertainty about the level set, and straddle-based methods,~\cite{Bryan2005,Gotovos2013,Inatsu2024} which target points that are simultaneously near the threshold and uncertain.
Our method builds on the boundary perspective via the straddle family, for reasons that will become clear once we introduce the multi-objective setting.

\subsection{Joint Pass Region: the Target of Multi-Objective Characterization}

Nearly all real pharmaceutical processes involve multiple quality attributes, and characterizing each objective independently ignores that a set of conditions is acceptable only if all attributes pass together.
The relevant target is the \textit{joint pass region}, the intersection of all individual pass regions,
\begin{align}
  \label{eq:joint-pass}
  \pasj = \bigcap_{i=1}^{m} \pas{i} = \{\point \in \params : O_i(\point) \ge h_i, \forall i\}.
\end{align}
Outside $\pasj$, at least one attribute fails its specification, and the identity of the failing objective is often the critical piece of information for process understanding.

$\pasj$ makes precise the ICH Q8 concepts already introduced in Section~1: its boundary is exactly the edge of failure, beyond which CQAs cannot be met.
In principle, a fully converged estimate of $\pasj$ would also describe the \textit{largest possible} design space, since every point inside satisfies every quality specification.
In practice, however, a regulator-approved design space is almost never drawn to coincide with our estimate of $\pasj$.
Real processes are subject to parameter variability and measurement noise, and any estimate of $\pasj$ obtained from finite data carries residual model uncertainty.
The design space is therefore deliberately chosen as a conservative subset strictly inside $\pasj$, with margins that reflect both process variability and the confidence level of the underlying characterization.
ICH Q8 explicitly notes that determining the edge of failure is not essential for establishing a design space.~\cite{ICH2009}
However, characterizing it efficiently is what makes better-justified design space choices possible, and this is exactly what JAREX is built to do.
We therefore adopt $\pasj$ as the mathematical target: it is the largest region consistent with all specifications, the upper bound against which any practical design space is measured.

\subsection{Joint Acceptable Region EXploration (JAREX)}

With the target $\pasj$ established, we now construct an acquisition function that directly learns its boundary, the edge of failure.
We build up from two well-understood single-objective components, upper confidence bound (UCB), and the randomized straddle algorithm, then extend to multiple objectives.

The classic UCB acquisition function~\cite{Auer2002,Srinivas2010} selects candidates by maximizing
\begin{align}
  \label{eq:ucb}
  \ucb(x) = \mu(x) + \beta^{1/2} \cdot \sigma(x),
\end{align}
where $\mu(x)$ and $\sigma(x)$ are the predicted mean and standard deviation from the surrogate, and $\beta^{1/2}$ tunes the balance between exploitation (high mean) and exploration (high uncertainty).
UCB targets extrema, not boundaries, but its exploitation-exploration trade-off structure is the scaffold for what follows.

The straddle algorithm of Bryan et al.~\cite{Bryan2005}, later formalized by Gotovos et al.~\cite{Gotovos2013}, reshapes UCB for level set estimation by optimizing \textit{against} the threshold instead of toward an extremum,
\begin{align}
  \label{eq:straddle-og}
  \str(x) = -|\mu(x) - h| + \beta^{1/2} \cdot \sigma(x).
\end{align}
The exploitation term $\mu(x)$ in UCB is replaced by $-|\mu(x) - h|$, which penalizes predictions \textit{far} from the threshold.
Maximizing $\str(x)$ therefore prefers candidates that are either close to the boundary (small $|\mu(x) - h|$) or highly uncertain (large $\sigma(x)$), exactly the two kinds of points that are most informative for learning the boundary.

The main practical difficulty with Eq.~\ref{eq:straddle-og} is choosing $\beta^{1/2}$ and the conventional value $\beta^{1/2} = 1.96$ limits efficient boundary exploration in practice.
To address this, Inatsu et al.~\cite{Inatsu2024} proposed a \textit{randomizing straddle variant} that samples $\beta$ at every iteration,
\begin{align}
  \label{eq:straddle-rand}
  \rstr(x) = \max\left[ \str(x), 0 \right],
\end{align}
with $\beta$ drawn fresh each step from a $\chi^2_2$ distribution,
\begin{align}
  \label{eq:chi2}
  \beta \sim \chi^2_2(\xi) = \tfrac{1}{2}\, e^{-\xi/2}, \quad \xi \ge 0,
\end{align}
which adaptively balances aggressive boundary refinement with occasional broader exploration.
More details are provided in Section~S1.2 of the Supplementary Information.

The randomized straddle algorithm $\rstr$ gives us a well-tuned way to learn a single boundary. MOPC, however, requires more than a per-objective treatment.
Running $\rstr_i$ independently per objective $i$ (a \textit{greedy} strategy) ignores the joint pass/fail condition and spends effort refining each individual boundary, including segments that lie inside another objective's fail region, rather than the \textit{joint} boundary that defines $\pasj$.
Joint Acceptable Region EXploration (JAREX) addresses this allocation problem by restricting the search to regions that are plausibly within $\pasj$ and then combining the per-objective boundary-learning scores there.

The first ingredient is an optimistic view of each individual pass region.
With surrogate models providing $\mu_i(\point)$ and $\sigma_i(\point)$, we cannot apply Eq.~\ref{eq:pass-region} directly: a point near the boundary with large $\sigma_i$ is exactly where sampling is most informative and must not be prematurely excluded.
Borrowing directly from UCB, we define the per-objective \textit{optimistic pass region} using the Upper Confidence Bound $\ucb_i(\point) = \mu_i(\point) + \beta_i^{1/2} \sigma_i(\point)$ (Eq.~\ref{eq:ucb}) as
\begin{align}
  \label{eq:optimistic-pass-region}
  \bopas{i} = \{\point \in \params : \ucb_i(\point) \ge h_i \},
\end{align}
and the \textit{joint optimistic pass region} as their intersection,
\begin{align}
  \label{eq:joint-optimistic-pass}
  \bopasj = \bigcap_{i=1}^{m} \bopas{i} = \{\point : \ucb_i(\point) \ge h_i, \forall i\}.
\end{align}
$\bopasj$ contains every candidate that is either confidently inside $\pasj$ or too uncertain to rule out.
It acts as a mask: only candidates $\hpoint \in \bopasj$ are eligible for selection.
The mask is needed because the straddle score introduced next is symmetric about the threshold.
It depends on $|\mu_i - h_i|$ and therefore takes the same value for a candidate that confidently passes objective $i$ as for one that confidently fails it.
Aggregating straddle scores alone thus cannot distinguish a point on the boundary of objective $j$ at which objective $i$ passes, which lies on the joint boundary, from one at which objective $i$ fails, which lies outside $\pasj$ and is uninformative for it.
The mask supplies this directional information by excluding candidates that confidently fail any objective.

Within this region, we compute the randomized-straddle score per objective,
\begin{align}
  \label{eq:rstr-i}
  \rstr_i(\hpoint) = \max\left[ -|\mu_i(\hpoint) - h_i| + \beta_i^{1/2} \cdot \sigma_i(\hpoint), 0 \right],
\end{align}
which quantifies how informative sampling $\hpoint$ would be for the $i$-th boundary.
We then combine the per-objective scores with a \textit{softmin}-weighted sum,
\begin{align}
  \label{eq:jarex}
  \jarex(\hpoint) &= \sum_{i=1}^{m} \text{softmin}_i\left[\rstr_i(\hpoint)\right] \cdot \rstr_i(\hpoint) \\
  \text{softmin}_i &= \frac{e^{-\rstr_i(\hpoint)/\tau}}{\sum_j e^{-\rstr_j(\hpoint)/\tau}}, \nonumber
\end{align}
with $\tau = 0.5$ by default.
The softmin-weighted sum interpolates between the mean ($\tau \to \infty$) and the min ($\tau \to 0$), and $\tau$ sets the score scale of this transition (scores are computed on standardized objectives).
Most of the joint boundary lies where a single objective is at its threshold while the others pass with margin, and the softmin does not exclude such points: when the scores are small compared with $\tau$, as near a well-resolved boundary, JAREX approaches the mean and ranks them by how informative they are for that objective.
When the scores are large, typically early in a campaign, it favors candidates informative for several objectives at once.
Whether a candidate can lie on the joint boundary at all is decided by the mask $\bopasj$, not by the aggregation: in closed-loop ablations on the kinetic benchmark, the mask carries most of JAREX's gain and the aggregation rules perform similarly once it is applied (Section~S5.1 of the Supplementary Information), so we regard the softmin as a reasonable default.
A detailed comparison with max, mean, fixed weighted-sum, and softmax combinations is given in Section~S1.3 of the Supplementary Information.
The next experiment is then selected via the familiar
\begin{align}
  \label{eq:best}
  \pointnext = \arg\max_{\hpoint} \left[\jarex(\hpoint)\right],
\end{align}
the surrogate is updated with the new observation, and the procedure repeats.

One implementation detail is worth addressing here, as it materially affects performance.
The joint optimistic pass region $\bopasj$ is an irregular, generally non-convex subset of $\params$ and cannot be written as a simple closed-form constraint.
A hard indicator mask creates a sharp cliff at its boundary with a flat zero plain outside, which blocks the gradient-based multi-start search, a strategy widely used by common optimizers to refine candidate points.~\cite{Balandat2020}
Section~S1.4 of the Supplementary Information gives the full discussion.
Instead, we use a differentiable soft mask.
For each $\point$, define the worst-case slack across objectives,
\begin{align}
  \label{eq:failure-distance}
  d(\point) = \min_i[\ucb_i(\point) - h_i],
\end{align}
so that $d \ge 0$ iff $\point \in \bopasj$.
The soft mask is then
\begin{align}
  \label{eq:soft-mask}
  M(\point) =
  \begin{cases}
    1 & d \ge 0 \\
    \operatorname{sech}(kd) & d < 0
  \end{cases},
\end{align}
where $k = 2$ by default and $\operatorname{sech}(x) = 2/(e^x + e^{-x})$.
$M$ is identically 1 throughout $\bopasj$ and decays smoothly with the slack $d$ outside it, avoiding the discontinuity of a hard indicator mask.
The softmin combination and the soft mask are the key components for extending classical randomized straddle to JAREX.
These two modifications break a specific step in the convergence proof of Ref.~\citenum{Inatsu2024}.
As a result, the single-objective guarantee does not carry through to the joint misclassification loss.
A much weaker statement can be recovered, but only by assuming that every objective is queried sufficiently often.
This assumption contradicts JAREX's explicit deprioritization of regions confidently outside $\bopasj$.
We therefore do not claim a formal bound here and defer a full analysis to future work.
Section~S1.5 of the Supplementary Information identifies exactly which step of the argument fails and why.
The empirical results in Section~3 demonstrate that this is an acceptable trade-off for the multi-objective setting.

Algorithm~\ref{alg:jarex} summarizes the full procedure as implemented in our open-source package \obsd,~\cite{obsidian} a library for algorithmic process design for pharmaceutical applications built on BoTorch~\cite{Balandat2020} and PyTorch.~\cite{Paszke2019}
The JAREX object inherits from BoTorch's \textit{MCAcquisitionFunction} class, so it can be readily reused in other packages of similar design.

Algorithm~\ref{alg:jarex} fits an independent GP to each objective, so JAREX couples the objectives only through the joint pass/fail condition, via the mask and the aggregation, and not through a correlated surrogate.
This is a deliberate default.
Every experiment measures all objectives, and for such fully observed designs with noise-free observations, a multi-output GP with a shared input kernel predicts each objective from that objective's observations alone, so correlations between objectives provide no transfer.~\cite{Bonilla2007}
Independent GPs additionally allow each objective its own length scales and noise level.
Because JAREX uses only the per-objective posterior marginals, a multi-output surrogate can be substituted without changing the acquisition function.

JAREX shares its target with earlier adaptive methods for regions that satisfy several specifications.
Garbuglia et al.~\cite{Garbuglia2021} and Qing et al.~\cite{Qing2022} identify feasible regions of engineering designs by Bayesian active learning, with one GP per specification and an information-gain acquisition function that, for independent GPs, is the sum of the gains for the individual specifications.
Qing et al.\ add a stopping criterion based on the classification accuracy estimated from the posterior.
The sum rewards uncertainty about any individual boundary, including boundary segments in regions that fail another specification, and is therefore structurally the unmasked mean rule of our ablation (Section~S5.1 of the Supplementary Information), with an entropy-based score in place of the straddle score.
Malkomes et al.~\cite{Malkomes2021} replace the Pareto frontier with user-defined thresholds and search for a diverse set of designs inside the resulting region of satisfaction through expected coverage improvement.
Unlike constraint active search, JAREX aims to classify the parameter space correctly rather than to collect satisfying designs, and therefore concentrates experiments on the joint boundary.
Its specific ingredients are per-objective randomized straddle scores, an optimistic joint-feasibility mask that keeps gradient-based candidate optimization available, and post-processing of the trained surrogates into the PARs and interaction maps expected in pharmaceutical process characterization.

\begin{algorithm}[h]
\caption{JAREX: Joint Acceptable Region EXploration}
\label{alg:jarex}
\begin{algorithmic}[1]
\Require Initial dataset $\mathcal{D}_0 = \{(\point_j, y_{i,j})\}$ for $m$ objectives
\Require Thresholds $\{h_i\}_{i=1}^m$, temperature $\tau$, decay $k$
\For{$t = 1, 2, \ldots, T$}
    \State \textbf{Train surrogate models:}
    \For{each objective $i = 1, \ldots, m$}
        \State Fit an independent GP model to obtain $\mu_i(\point)$ and $\sigma_i(\point)$
    \EndFor
    \State \textbf{Sample} $\beta_i \sim \chi^2_2$ for $i = 1, \ldots, m$ \Comment{Eq.~\ref{eq:chi2}}
    \State \textbf{Compute JAREX acquisition} for all candidates $\point$:
    \For{each objective $i = 1, \ldots, m$}
        \State $\rstr_i(\point) = \max[-|\mu_i(\point) - h_i| + \beta_i^{1/2}\sigma_i(\point), 0]$ \Comment{Eq.~\ref{eq:rstr-i}}
    \EndFor
    \State $\text{softmin}_i = \frac{\exp(-\rstr_i/\tau)}{\sum_j \exp(-\rstr_j/\tau)}$
    \State $\jarex(\point) = \sum_{i=1}^{m} \text{softmin}_i\left[\rstr_i(\point)\right] \cdot \rstr_i(\point)$ \Comment{Eq.~\ref{eq:jarex}}
    \State \textbf{Compute a soft mask} for all candidates $\point$:
    \State \quad $d(\point) = \min_i[\ucb_i(\point) - h_i]$ \Comment{Eq.~\ref{eq:failure-distance}}
    \State \quad $M(\point) = \begin{cases} 1 & d \ge 0 \\ \text{sech}(kd) & d < 0 \end{cases}$ \Comment{Eq.~\ref{eq:soft-mask}}
    \State \textbf{Select next point:}
    \State \quad $\pointnext = \arg\max_{\point} \left[ M(\point) \cdot \jarex(\point) \right]$ \Comment{Eq.~\ref{eq:best}}
    \State \textbf{Evaluate} $y_{i,t} = O_i(\pointnext)$ for all objectives
    \State \textbf{Update} $\mathcal{D}_t = \mathcal{D}_{t-1} \cup \{(\pointnext, \{y_{i,t}\})\}$
\EndFor
\end{algorithmic}
\end{algorithm}

\section{Results and Discussion}

In this section, we demonstrate the effectiveness of the randomized straddle and JAREX algorithms for process characterization.
We begin by establishing benchmark metrics and illustrating the boundary-focused selection strategy of randomized straddle.
We then present benchmark results for single-objective characterization using randomized straddle and multi-objective characterization using JAREX on a simulated chemical kinetics problem and on two pharmaceutical case studies.
Our results show that JAREX characterizes the joint pass region more accurately than the tested designed experiments and space-filling sampling in every case, and more accurately than a greedy multi-objective approach, by a wide margin on the kinetic model, where most of each objective's boundary lies outside the joint pass region, and by up to 0.06 in most pharmaceutical settings, while maintaining robust performance across different transformation strategies.
Finally, we demonstrate that surrogate models trained with JAREX can recover traditional process characterization outputs, including proven acceptable ranges and multi-factor interaction plots.

\subsection{Benchmark Details}

We evaluate method performance by comparing the identified pass regions with the true pass regions (known ground truth) using the Jaccard index,~\cite{Jaccard1901}
\begin{align}
  \label{eq:jaccard}
  J(\paspred, \pastrue) = \frac{|\paspred \cap \pastrue|}{|\paspred \cup \pastrue|},
\end{align}
where $\paspred$ and $\pastrue$ are the predicted and true pass regions, respectively.
The Jaccard index ranges from 0 (no overlap) to 1 (perfect match), providing a set-based metric focused on boundary classification rather than prediction error magnitude.
The Jaccard index is an appropriate metric for process characterization because correct boundary identification inherently requires accurate predictions near the threshold, where it matters most (SI Section S2 includes an extended discussion of this index).

For all benchmark studies presented here, we use response surface methodology (RSM) designs and space-filling (SF) sampling as baseline approaches.
The RSM designs comprise a full two-level factorial and a central composite design (CCD).
For the multi-objective benchmarks, we add the model-based design a practitioner would build when these become too large or too coarse: an I-optimal design for a full quadratic model, which minimizes the average prediction variance over the parameter space,~\cite{Goos2011} with the same budget as the adaptive methods (SI Section S5.12).
Our conclusions about designed experiments refer to these designs: the factorial and the CCD for every benchmark and, in addition, the I-optimal design for the multi-objective benchmarks.
Each test function is evaluated with 10 independent campaigns using fixed random seeds across methods.
All methods share the same initial observations generated via Latin hypercube sampling (LHS), with campaign budgets and implementation details provided in the Supplementary Information.
Gaussian process surrogate models with Mat\'ern kernels (BoTorch implementation) are used throughout.
For the single-objective and kinetic benchmarks, they use BoTorch's default hyperparameter priors, optimized via multi-start optimization (see SI Section S2 for optimization details).
For the two pharmaceutical case studies, with up to 13 inputs, they use the dimension-scaled priors of Hvarfner et al.~\cite{Hvarfner2024} with a Mat\'ern-5/2 kernel and BoTorch's default fit, for every method alike. For the chromatography case this choice was made before the campaigns by a model check on fixed space-filling designs, and the results with the default priors are reported in SI Section S5.6.
The crystallization case uses the same priors at every input count, including 3 inputs, where the model check favored the default priors.
Jaccard indices are calculated on Sobol-discretized parameter spaces, with evaluation point density adapted to surrogate uncertainty.

\begin{figure}[h]
  \centering
  \includegraphics[width=0.45\textwidth]{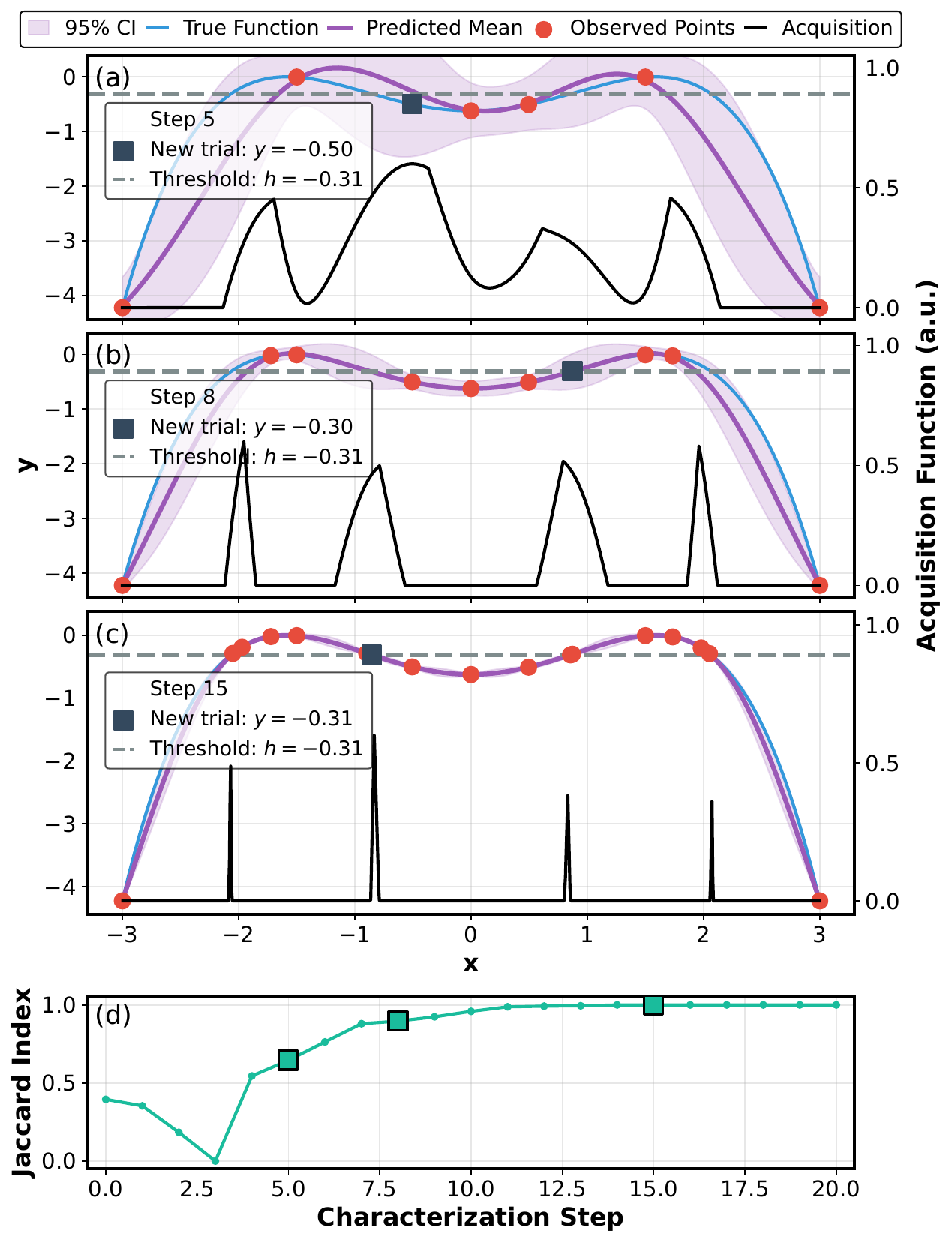}
  \caption{
    Next experiment selected by the randomized straddle algorithm at steps $5$ (panel a), $8$ (panel b), and $15$ (panel c) of a one-dimensional characterization campaign.
    Blue curve: true function; gray dashed line: threshold; purple curve and shaded area: GP mean and uncertainty; red dots: existing observations; black square: next selected experiment.
    The black curve shows the acquisition function value (arbitrary scale), with higher values indicating stronger sampling preference.
    Panel d: characterization history across iterations.
  }
  \label{fig:demo}
\end{figure}

\subsection{Selection Rules in Randomized Straddle}

To select samples efficiently in an iterative setting, the randomized straddle algorithm (Eq.~\ref{eq:straddle-rand}) prefers candidates that are either near the threshold or have high uncertainty.
Fig.~\ref{fig:demo} illustrates this behavior on a one-dimensional double-well potential test case (see SI Section S3.1.1 for function definition), showing the next experiments at different campaign steps (panels a-c) and the overall characterization history (panel d).
The algorithm selects candidates with high uncertainty early in the campaign (e.g., step 5) and focuses on the proximity of the threshold as uncertainty decreases (e.g., step 8).
By step 15, the algorithm continues to sample exactly at the threshold to refine the boundary, which causes the Jaccard index to plateau as further improvements become marginal.
Importantly, the algorithm ignores regions far from the boundary even when the surrogate mean deviates from the true function, as these regions do not affect pass/fail classification.
This boundary-focused sampling strategy exemplifies the fundamental difference between characterization (exploring pass/fail boundaries) and optimization (finding optimal points).

\subsection{Single-Objective Characterization}

\begin{figure}[h]
  \centering
  \includegraphics[width=0.45\textwidth]{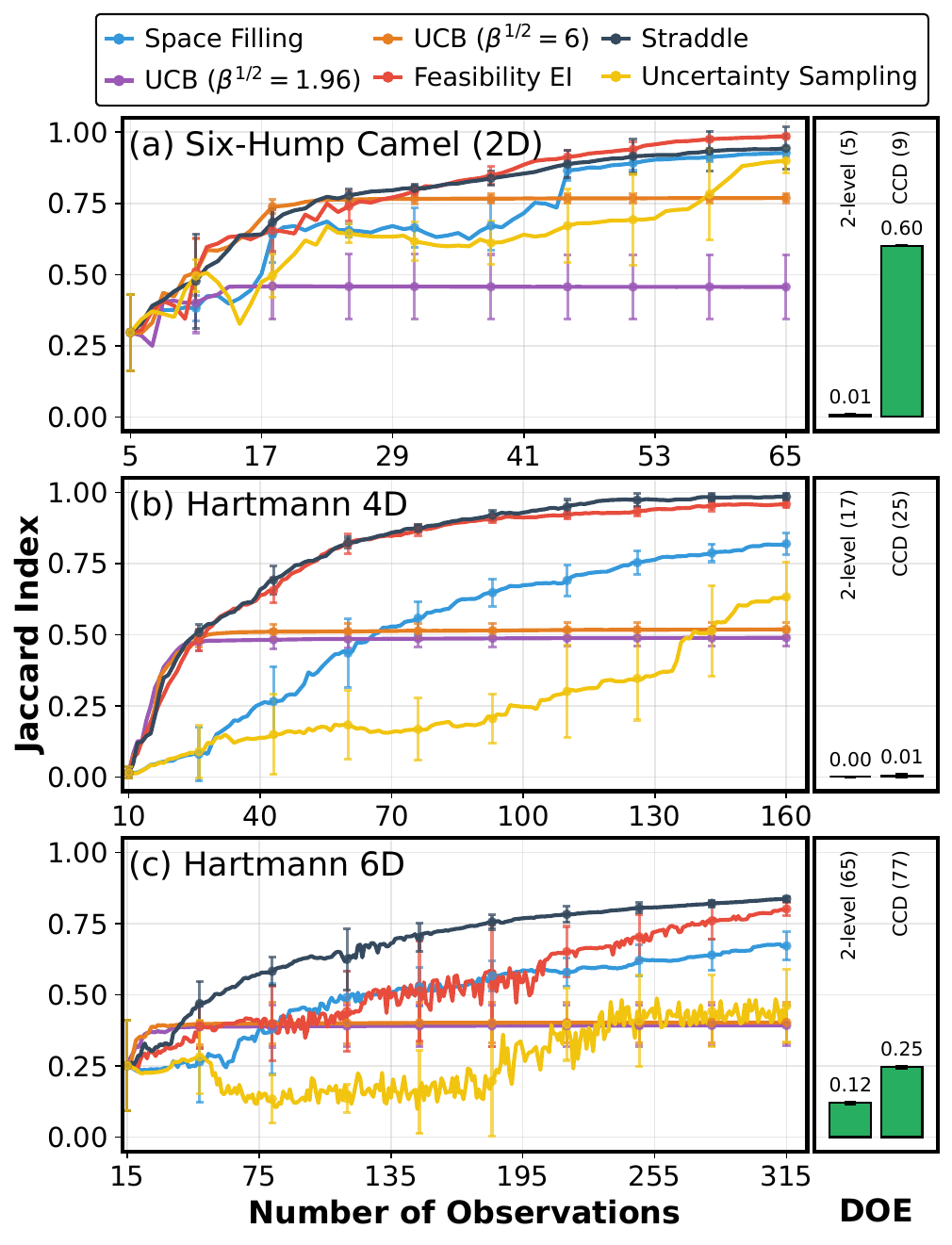}
  \caption{Benchmark results for single-objective characterization on noise-free analytical test functions: six-hump camel (2D), Hartmann 4D (4D), and Hartmann 6D (6D).
  Jaccard index convergence (mean and standard deviation averaged across 10 campaigns) for SF, UCB ($\beta^{1/2}=1.96$ and $\beta^{1/2}=6$), feasibility EI, randomized straddle, and uncertainty sampling.
  The side panel (DOE) reports the RSM designs, a full two-level factorial and a central composite design (CCD), as single-design Jaccard values, with run counts in parentheses.
  }
  \label{fig:so}
\end{figure}

Fig.~\ref{fig:so} shows the benchmark results of applying the above-mentioned randomized straddle algorithm to a single-objective case study.
We use noise-free response functions here (results with 5\% Gaussian noise are provided in the Supplementary Information) so that the comparison isolates each method's intrinsic sampling strategy at modest budgets, rather than its tolerance to observation noise.
This choice does not amplify any single method.

The results show three things.
First, UCB, which is designed for optimization rather than characterization, performs poorly beyond 2D, even with high exploration ($\beta^{1/2}=6$).
Second, SF is competitive in two dimensions but falls off sharply as the dimension increases, and the RSM designs are far less accurate on every function.
Third, among the characterization methods, feasibility EI is the most accurate on the two-dimensional function and randomized straddle on Hartmann 4D and 6D, and in six dimensions it clearly outperforms all other approaches.
Uncertainty sampling falls below SF on every function, because it samples where the surrogate is uncertain rather than where the threshold lies (final values in SI Table S17).

\subsection{Multi-Objective Characterization}

Next, we benchmark the developed acquisition function, JAREX, on a multi-objective task using a simulated reaction kinetic model.
Specifically, we compare how JAREX performs relative to RSM designs, SF, and a greedy multi-objective extension of randomized straddle (which is designed for single-objective problems).
The greedy approach works through the objectives blockwise, in a fixed order: for a contiguous block of iterations (40 per objective on this benchmark), randomized straddle selects every experiment based on one objective, and then moves on to the next objective, while all objectives are measured and used to train the surrogate models at every step.
We use this schedule to represent a staged objective-prioritization workflow, in which the experimental program focuses on one quality attribute for a block before moving to the next, rather than all possible objective-wise sampling policies.
The joint pass region is obtained by intersecting individual pass regions.
The per-objective acquisition does not distinguish boundary segments relevant to joint classification from those lying where another objective fails.

We developed a four-dimensional kinetic model simulating competing reaction pathways in a chemical process.
The reaction scheme consists of a main catalytic pathway producing the desired product P, alongside a competing reversible side reaction forming impurity X:
\begin{align}
    \ce{Y + Z &->[$k_{\text{main}}(T, \text{Cat})$] P} \label{eq:main_rxn} \\
    \ce{2Y + Z &<=>[$k_{\text{side,fwd}}(\text{Cat})$][$k_{\text{side,rev}}(T)$] X} \label{eq:side_rxn}
\end{align}
where $Z$ is the reactant, $Y$ is the co-reactant, $P$ is the desired product, $X$ is an undesired impurity, and $Cat$ represents the catalyst.

\begin{figure}[H]
  \centering
  \includegraphics[width=0.45\textwidth]{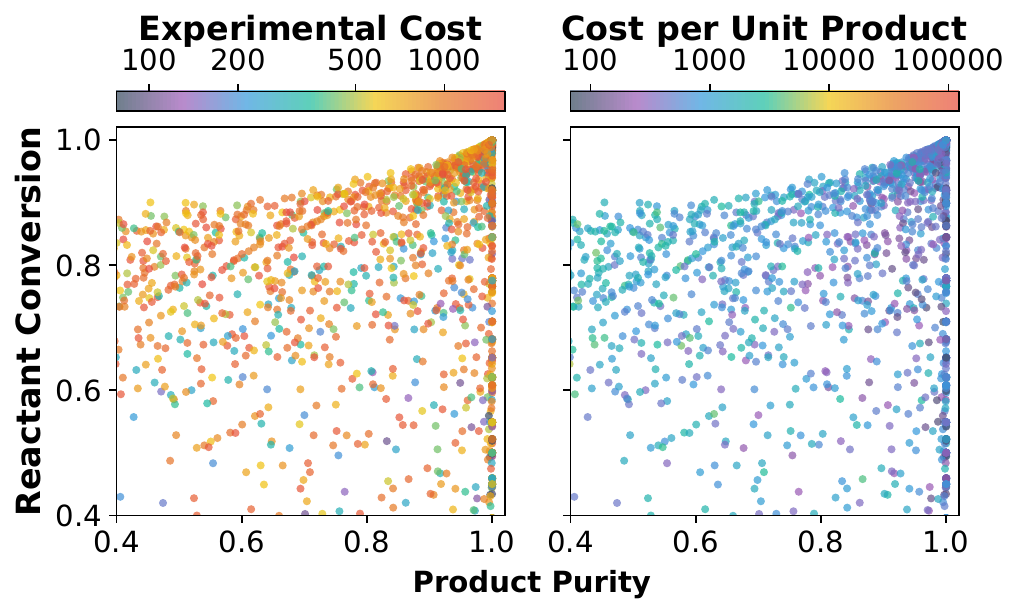}
  \caption{
    Trade-offs among kinetic model objectives.
    Reactant Z conversion vs. product purity colored by raw experimental cost in panel (a) and cost per unit product in panel (b).
    }
  \label{fig:model}
\end{figure}

The parameter space is defined by four input parameters: initial reactant concentration ([$Z_0$]), initial co-reactant concentration ([$Y_0$]), temperature ($T$), and catalyst loading ([$Cat_0$]).
Together, these parameters create coupled trade-offs among reactant Z conversion, product purity, and cost per unit product (Fig.~\ref{fig:model}).
Detailed reaction mechanisms and mechanistic interpretation are provided in Section~S4.1 of the Supplementary Information.

We evaluate three competing objectives: conversion (maximize), product purity (maximize), and cost per unit product (minimize).
Conversion is the fraction of initial reactant $Z_0$ consumed.
Purity is the molar ratio $[P]/([P]+[X])$ at the final time point.
Cost is defined per mole of desired product $P$, accounting for reactant $Z$, co-reactant $Y$, catalyst, and solvent consumption (temperature affects cost indirectly through kinetics and final yield).

These objectives exhibit inherent trade-offs, as shown in Fig.~\ref{fig:model}.
For instance, high conversion requires high [$Z_0$] and [$Y_0$], while high purity requires suppression of the side reaction, which often reduces conversion.
Conversely, the high-conversion, high-purity regime (upper right in Fig.~\ref{fig:model}) incurs the highest raw material cost yet yields the lowest cost per unit product due to superior process efficiency.
The objectives are physically coupled through the underlying kinetics with trade-offs and reinforcements that shift across the parameter space.
This non-trivial interplay makes the kinetic model a meaningful benchmark for multi-objective characterization methods, closer in structure to real process problems than analytical test functions.

Table~\ref{tbl:kinetic_model} lists the parameter space and objective thresholds. See Supplementary Information for complete kinetic equations, rate expressions, and integration details.

\begin{table}[h]
  \centering
  \small
  \caption{Parameter space and objective thresholds for the kinetic model.}
  \label{tbl:kinetic_model}
  \begin{tabular}{@{} m{1.2cm} m{2.1cm} !{\color{white}\vrule width 6pt} m{2.2cm} m{1.4cm} @{}}
    \cmidrule[\heavyrulewidth](r){1-2}\cmidrule[\heavyrulewidth](l){3-4}
    \multicolumn{2}{c}{\textbf{Parameter space}} & \multicolumn{2}{c}{\textbf{Objectives}} \\
    \cmidrule(r){1-2}\cmidrule(l){3-4}
    Variable & Range & Objective & Threshold \\
    \cmidrule[\lightrulewidth](r){1-2}\cmidrule[\lightrulewidth](l){3-4}
    $[Z_0]$ & [0.8, 3.0]\,M & Z conversion & $\ge 90\%$ \\
    $[Y_0]$ & [0.8, 6.0]\,M & Purity & $\ge 99.3\%$ \\
    $[Cat_0]$ & [0.005, 0.2]\,M & \multirow{2}{2.2cm}{Cost per unit product} & \multirow{2}{*}{$\le \$570$} \\
    $T$ & [273, 333]\,K & & \\
    \cmidrule[\heavyrulewidth](r){1-2}\cmidrule[\heavyrulewidth](l){3-4}
  \end{tabular}
\end{table}

\begin{figure}[h]
  \centering
  \includegraphics[width=0.45\textwidth]{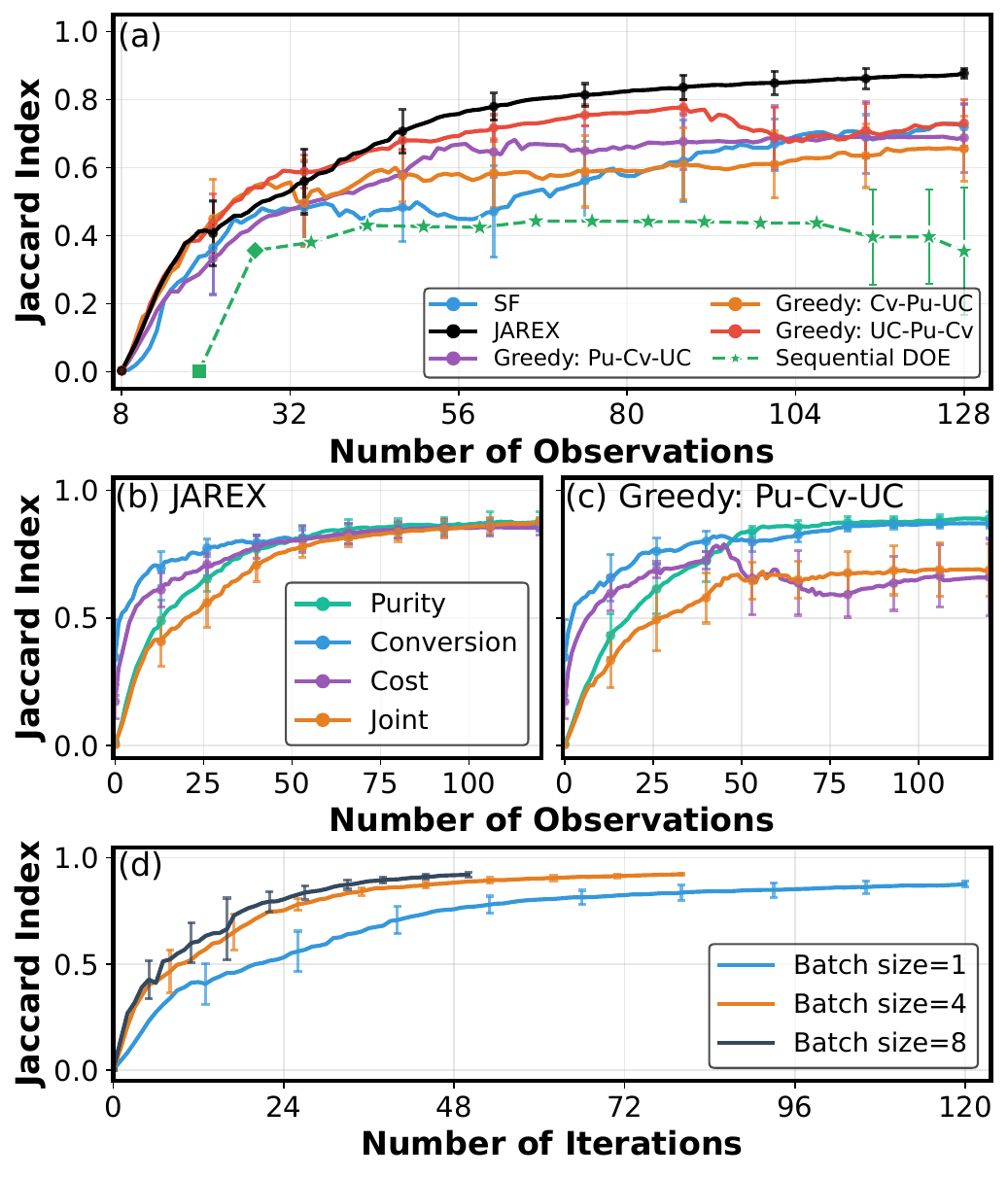}
  \caption{
    Multi-objective characterization benchmarks for reactant Z conversion, product purity, and cost per unit product with 5\% multiplicative Gaussian noise added to the simulated final concentrations.
    (a) Joint Jaccard index convergence for SF, greedy randomized straddle (multiple objective orderings), and JAREX, and for the sequential DOE (dashed green), scored with the same GP surrogate: its first two points are the full two-level factorial (square, 19 runs) and the CCD (diamond, 27 runs), followed by I-optimal blocks of eight runs (stars) up to 128.
    In two of the ten campaigns, the GP fit fails late in the sequence, which lowers the mean and widens the error bars at 115--128 runs.
    (b-c) Individual objective and joint Jaccard indices for JAREX and for greedy with the objective order purity, conversion, cost, respectively.
    (d) JAREX batching comparison: serial ($q=1$), moderate ($q=4$), and high ($q=8$) batch sizes.
    Curves show average Jaccard indices across 10 campaigns. Error bars indicate $\pm 1\sigma$.
  }
  \label{fig:mo_bench}
\end{figure}

Fig.~\ref{fig:mo_bench} presents benchmark results for the kinetic model.
All methods use a standard transformation of the response function (see Section~S4.2 for log-transformed results and transformation analysis).
Unlike the single-objective benchmarks, where noise is added to the response, measurement noise here is applied to the simulated final species concentrations before the objectives are computed.
The noise therefore propagates differently into each objective and is correlated between purity and cost through the product concentration (Section~S4.1).

Fig.~\ref{fig:mo_bench}a shows joint Jaccard convergence on this benchmark.
JAREX (black curve) reaches a higher final joint Jaccard index than the tested DOE, space-filling, and blockwise greedy comparators, rising rapidly within the first 30--40 iterations.
The curves cross early in the campaign, so this is not a claim of superiority at every intermediate budget.
This confirms that explicitly targeting the joint edge of failure (Eq.~\ref{eq:joint-optimistic-pass}) yields efficient characterization of the joint pass region even in this four-dimensional, three-objective, noisy setting.
The errors that remain after 120 iterations are mostly conservative and lie along the joint boundary, mainly where the steep purity response is not yet resolved (Section~S5.11 of the Supplementary Information).

The non-adaptive baseline falls well short of JAREX.
The RSM designs perform poorly despite being conventional go-to methods for process characterization, illustrating that grid-based designs scale poorly to realistic parameter spaces.
A sequential design built the way a practitioner would build it does not close the gap.
We augmented the CCD in blocks of eight I-optimal runs for a full quadratic model, each block chosen given all runs before it, up to 128 experiments (Fig.~\ref{fig:mo_bench}a; SI Section S5.12).
Scored with the same GP surrogate as the other methods or with full quadratic models, as in response surface methodology, it levels off far below JAREX (Fig.~\ref{fig:mo_bench}a).
Designs that are optimal for predicting a quadratic place their runs on three levels per input, which leaves few distinct points in the interior of the parameter space, where the boundaries lie.
SF improves on all of these designs through better coverage but still plateaus far below JAREX, because uninformed sampling cannot redirect effort toward the joint boundary as observations accrue.

On this benchmark, the tested blockwise strategy has two limitations.
First, convergence depends on the objective order, producing the different convergence histories shown for the three orderings.
The most useful objective to prioritize is not known in advance.
Our conclusions about order sensitivity and plateaus refer to this schedule.
We did not test interleaved schedules.
Second, its per-objective acquisition can spend experiments refining boundary segments that lie where another objective fails and therefore do not contribute to the joint pass region.
JAREX uses the joint-feasibility mask to deprioritize these segments and requires no fixed objective order, yielding steadier convergence and higher final joint accuracy than the tested blockwise schedules.
The size of this benefit depends on the response geometry and on whether the surrogates can identify regions that already fail another objective, as the case studies of Section~\ref{sec:pharma} show.

The individual-versus-joint performance trade-off is illustrated in Fig.~\ref{fig:mo_bench}b-c, where panel~b shows JAREX and panel~c shows the greedy results.
JAREX trades off individual objective performance to achieve superior joint characterization performance by deprioritizing high-uncertainty regions known to fail for at least one objective (Eq.~\ref{eq:joint-optimistic-pass}).
This strategy focuses the experimental budget on the joint pass region, which is what matters for process understanding and decision-making.
The greedy method, by contrast, achieves high Jaccard indices for purity and conversion but performs notably worse for cost.
Interestingly, the joint Jaccard index \textit{exceeds} the worst individual objective (cost). 
This occurs because the three objectives are physically coupled through product yield $P_f$.
The true joint pass region is 2.4 times larger than expected under independence assumptions.
The cost pass set includes a large region failing purity or conversion.
While these points inflate the cost surrogate’s classification errors, the joint classifier correctly rejects them, improving joint precision.
A detailed quantitative breakdown of these effects is provided in the Supplementary Information (Section~S4.3).

In practical experimental settings, real-time campaign duration is sometimes more constraining than total sample budget especially when development timelines are more precious than material cost.
Batching, where multiple experiments are run in parallel, is a critical strategy for reducing calendar time, especially when individual experiments require hours or days.
However, batching naturally poses a challenge for the efficiency of BO-based methods. 
Without the ability to update the surrogate between experiments within a batch, the algorithm must commit to multiple points simultaneously, potentially leading to redundant sampling or suboptimal exploration.
Fig.~\ref{fig:mo_bench}d evaluates JAREX under three batch scenarios: serial ($q=1$, 120 iterations/120 samples), moderate batching ($q=4$, 80 iterations/320 samples), and high batching ($q=8$, 50 iterations/400 samples).
Batching trades additional experiments for fewer experimental rounds.
Measured in rounds, $q=4$ and $q=8$ reach the final serial Jaccard index in less than half as many sequential rounds (46 and 35 instead of 120), and hence shorten the campaign when rounds take similar time.
Measured in experiments, the serial campaign remains the most sample-efficient, as expected, because every selection uses all previous observations: matching the final serial accuracy takes 1.5 and 2.25 times as many experiments with $q=4$ and $q=8$, and at an equal budget serial sampling is the most accurate (Fig.~S4 of the Supplementary Information).
JAREX therefore suits time-sensitive workflows in which the number of ``wait-and-decide'' cycles, rather than the total number of experiments, is the binding constraint.

\subsection{Pharmaceutical Case Studies}
\label{sec:pharma}

To test JAREX beyond the kinetic model, we used two published models in which both the number of inputs and the number of objectives can be varied (SI Section S5.6).
These are in silico case studies, not prospective experimental validation.
The first is a reversed-phase liquid chromatography (HPLC) method for five phenothiazines, simulated with the mechanistic retention model of Kamedulska et al.,~\cite{Kamedulska2022} which was fitted to measured retention times. 
The 5, 9, or 13 inputs control pH, temperature, flow rate, and the gradient program.
The 2, 4, or 7 objectives are critical-pair resolution, run time, and first-peak retention.
The second is a combined cooling and antisolvent crystallization of aspirin in a cascade of mixed-suspension, mixed-product-removal (MSMPR) crystallizers, adapted from the cascade model of Liu and Benyahia~\cite{Liu2022msmpr} with the kinetics measured by Lindenberg et al.~\cite{Lindenberg2009}
The 3--12 inputs control temperature, antisolvent flow, and residence time at each stage.
The 2, 3, or 4 objectives are yield, crystal size, productivity, and supersaturation in the first vessel.
Both specification sets are defined for the benchmark: the chromatography objectives cover selected peak pairs, not every pair, and much of the crystallization pass region lies beyond the range of the kinetic data (SI Section S5.6).
Both models include measurement noise, and every campaign starts from 8 points and runs 120 iterations (128 experiments).
The 12 combinations with 9 or more inputs were also run to 512 experiments.

\begin{figure*}[htp]
  \centering
  \includegraphics[width=\textwidth]{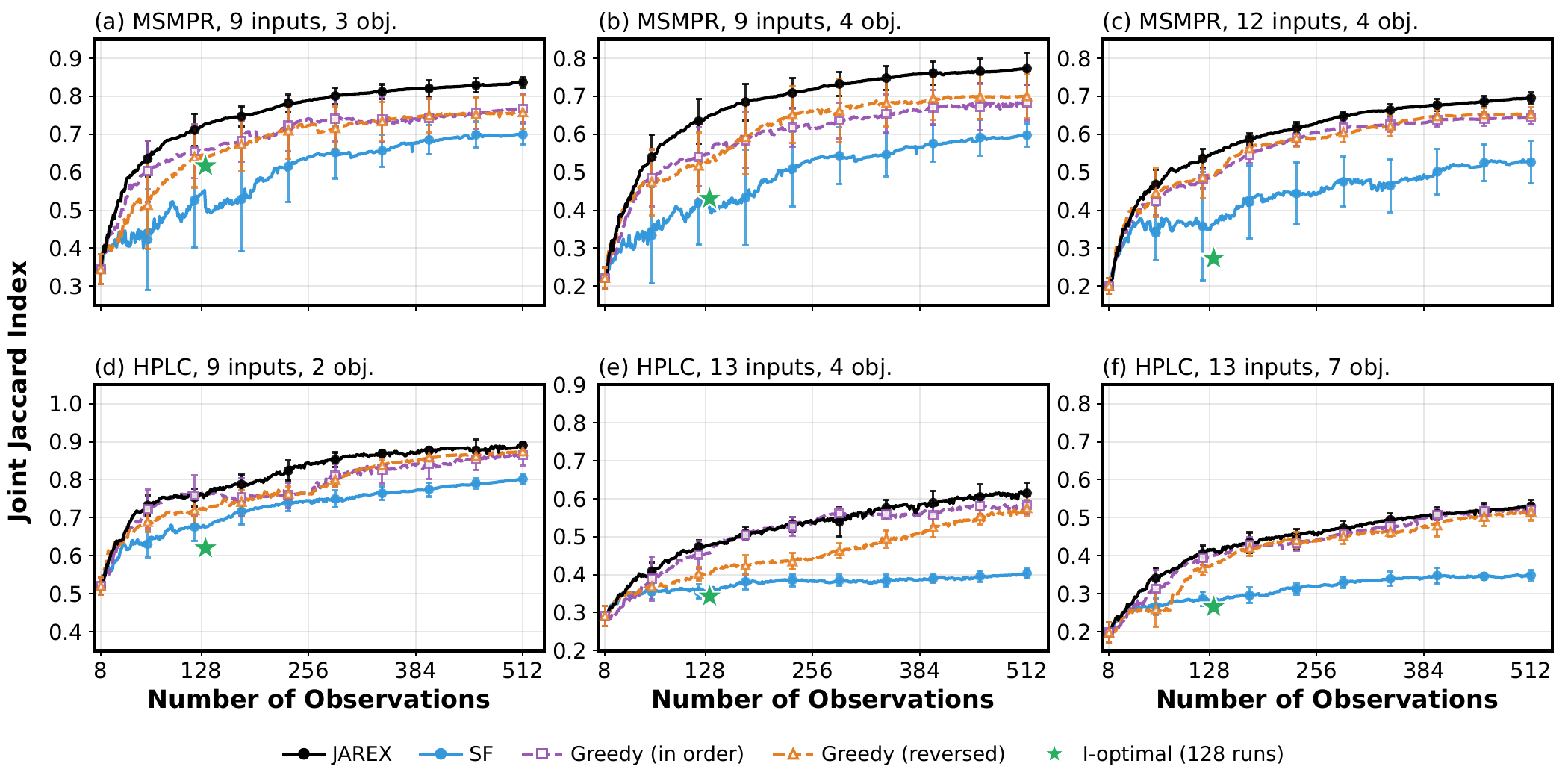}
  \caption{
    Joint Jaccard index on the pharmaceutical case studies over 512 experiments (mean of 10 campaigns with dimension-scaled GP priors; error bars $\pm 1\sigma$), for six combinations of inputs and objectives chosen after the runs.
    (a--c) MSMPR crystallization cascade; (d--f) HPLC method.
    JAREX and SF continue their 128-experiment campaigns.
    The greedy strategy, whose blocks depend on the total budget, was run to 512.
    All 21 combinations at 128 experiments, and the 12 with 9 or more inputs to 512, are in SI Section S5.6.
    Each panel spans 0.70 in the index, so differences have the same height in every panel.
    Green star: the 128-run I-optimal design, drawn just right of 128.
    The factorial and the CCD need more than 128 runs at these input counts.
  }
  \label{fig:pharma}
\end{figure*}

Fig.~\ref{fig:pharma} shows six of the 21 combinations of inputs and objectives, chosen after the runs, over 512 experiments.
All combinations are in SI Section S5.6.
JAREX is more accurate than SF and than every design in all 21 combinations, including the factorial and CCD designs that need up to 533 runs.
It is also more accurate than the greedy strategy, whose objective order must be fixed in advance: averaged over the two orders, JAREX is ahead in 12 combinations and behind in one.
Against whichever order turned out better, a choice possible only after the runs, it is ahead in seven and behind in two.
The two case studies differ in how large this gain is.
On the crystallization case, JAREX pulls clearly ahead of both greedy orders (panels a--c), and its lead grows with the budget, to about 0.07 at 512 experiments.
Even where the two were level at 128, JAREX is ahead at 512 (panel c).
On the chromatography case, JAREX stays ahead but by much less (panels d--f), with the curves close throughout.

The difference between the two cases follows the allocation argument of Section~2.2: the greedy strategy spends experiments on boundary segments that lie where another objective already fails, and JAREX can avoid them only where its surrogates recognize that failure.
We measured the share of each boundary that lies where another objective fails, the waste, on the noise-free truth (SI Section S5.6).
Most of each boundary is wasted in this sense on the kinetic model, where JAREX leads by 0.15, and on the crystallization combinations where its lead is largest.
The crystallization responses are smooth, and the surrogates learn where each specification fails.
On the chromatography case, the waste is as high with four or seven objectives, but each resolution response has a kink where two peaks swap their elution order, which leaves the GP surrogates uncertain exactly where other objectives fail.
The mask then can rarely exclude a candidate, and removing it barely changes the result (SI Section S5.1).
Refitting the same experiments with a smooth signed resolution in place of the kinked one raises the accuracy of every method and cuts false acceptance about threefold (SI Section S5.6), which supports this explanation.
A broad optimistic region, with about a fifth of the box passing, further limits what the mask can exclude.
On this case, therefore, joint-feasibility targeting adds little over refining each objective in turn, because the surrogates cannot resolve the responses where the objectives interact.
Two tests designed before the runs (SI Section S5.6) support the waste argument: on an integrated continuous aspirin process,~\cite{Nagy2021} the gain rose with the waste as predicted, and on the crystallization case, a measure combining each objective's waste with the difficulty of its boundary predicted the gain at 9 inputs, though not at 12 with 128 experiments.

These results use the dimension-scaled priors.
With BoTorch's default priors, nearly every method is less accurate on the chromatography case, and at 13 inputs even SF stays near its starting accuracy for most of the budget (SI Section S5.6), a failure of the default model rather than of any sampling strategy.

\subsection{Application to Standard Process Characterization Metrics}

Traditional process characterization studies are interpreted through metrics such as proven acceptable ranges (PARs) and multi-factor interactions.
Having demonstrated the effectiveness of JAREX for efficient characterization of the joint pass region, we next evaluated its utility for recovering example process characterization outputs.
Although randomized straddle and JAREX are not formulated to directly target these conventional metrics, they are designed to learn the underlying pass/fail boundary.
The resulting surrogate therefore provides a high-dimensional representation of process robustness, from which traditional characterization metrics can be recovered as lower-dimensional projections through post-processing.
Because this is a broader learning problem than dedicated univariate or bivariate characterization, accuracy for any single derived metric at a fixed experimental budget may be somewhat lower than that of approaches optimized specifically for that metric.

We used the first 40 experiments of one of the ten serial JAREX campaigns on the kinetic model (Section~3.4): 8 initial observations and 32 experiments designed by maximizing its acquisition function (Eq.~\ref{eq:jarex}), with the GP surrogate retrained after each new observation.
The campaign shown is at the median accuracy of the ten after 40 experiments.
The resulting trained surrogate is the input to all PAR analyses below.
The JAREX-derived PARs are obtained directly from the trained posterior, evaluated along univariate scans through the centroid of the identified joint pass region.

\begin{figure}[H]
  \centering
  \includegraphics[width=0.45\textwidth]{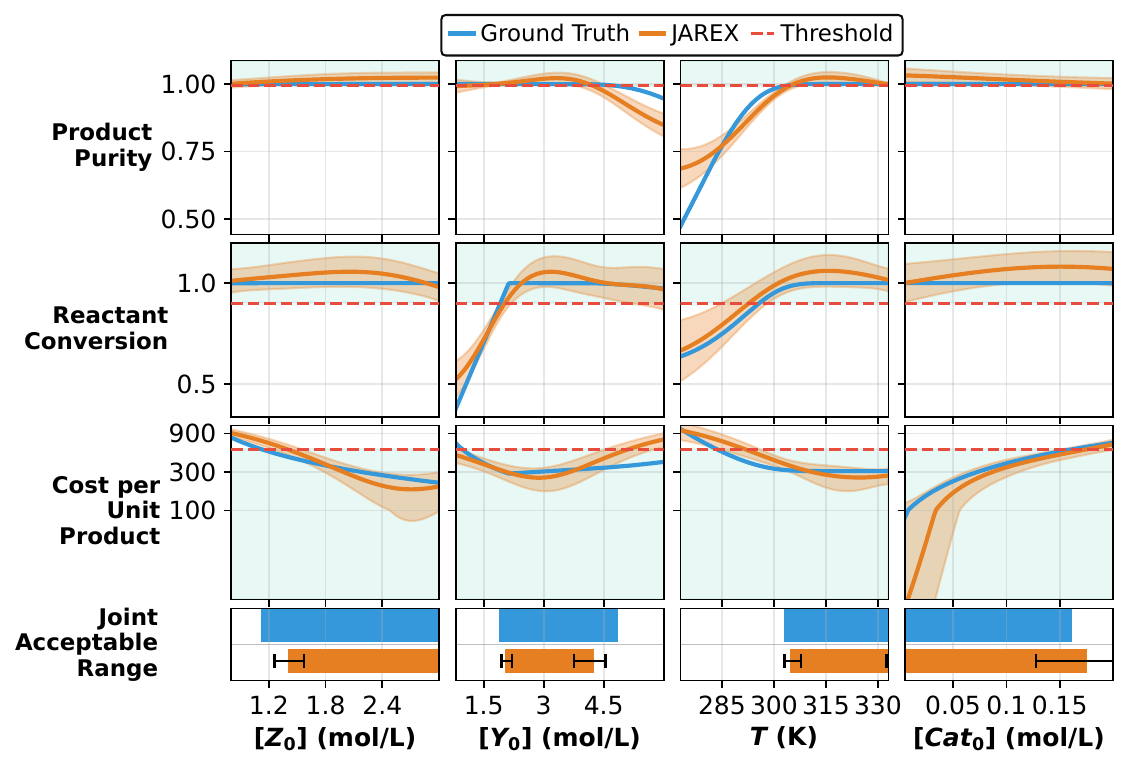}
  \caption{Comparison of proven acceptable ranges (PAR) for reactant Z conversion, product purity, and cost per unit product between JAREX after 40 experiments and the ground truth, a noise-free scan of each input through the same anchor.
  Top three rows: individual objective pass regions (green shaded areas).
  Bottom row: joint PAR shown as horizontal bars, with the ground truth in blue and the JAREX prediction at the mean level in orange.
  Each bracket shows how far a joint limit moves between every objective at its favorable and at its unfavorable 70\% prediction bound.
  The inner end is the conservative limit.
  JAREX predictions closely track ground truth within and near the joint pass region, while showing larger deviations outside, reflecting its strategy of focusing experimental budget on the joint pass region.
  }
  \label{fig:par}
\end{figure}

Fig.~\ref{fig:par} compares the JAREX-derived PARs with the true PARs, from a noise-free scan of each input through the same anchor, for all three objectives and for the joint region.
Each objective limits the joint ranges along different inputs: purity sets the upper limit of [$Y_0$] and the lower limit of $T$, conversion the lower limit of [$Y_0$], and cost the lower limit of [$Z_0$] and the upper limit of [$Cat_0$], and the same objectives set the same limits in all ten campaigns.
JAREX tracks the individual responses closely near their thresholds, and its joint ranges are mostly conservative, lying inside the true ranges at every limit except slightly beyond the upper limit of [$Cat_0$].
The other campaigns behave alike (Supplementary Information, Section~S4.4).
Outside the joint pass region, individual-objective uncertainty increases, which is expected because JAREX intentionally deprioritizes areas already likely to fail at least one objective.
This allocation concentrates experimental effort where it most affects joint characterization.
In this campaign, the joint range at the unfavorable 70\% bounds lies inside the true range at every limit, including the upper limit of [$Cat_0$].
Such a range is not guaranteed after 40 experiments: purity's threshold (0.993) lies just below its ceiling of 1, and in six of the ten campaigns the 70\% bounds of the purity surrogate still straddle the threshold at the anchor, so no 70\% joint range contains it.
After 80 experiments, one does in nine of the ten.

Two-dimensional interaction effects between pairs of process parameters are commonly used in process characterization to understand the interplay between parameters and identify potential synergies or trade-offs.
Traditionally, these interaction maps require dedicated two-factor campaigns at multiple fixed settings of the remaining parameters, which adds experimental cost and time to the process characterization study.
With a trained JAREX-based model, they can be obtained directly by evaluating the posterior on two-parameter grids while fixing the remaining parameters, with no additional experiments.

\begin{figure}[h]
  \centering
  \includegraphics[width=0.45\textwidth]{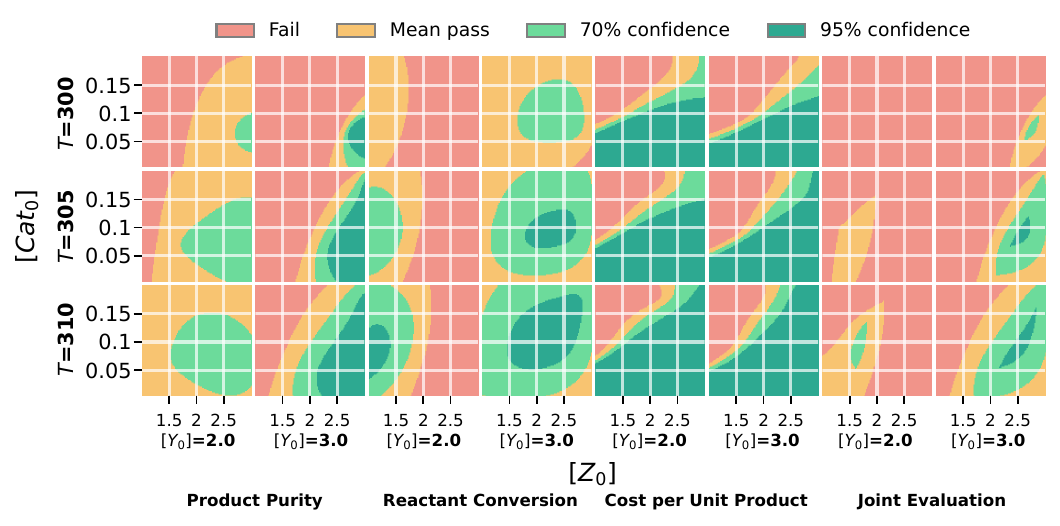}
  \caption{Two-dimensional interaction effects for reactant Z conversion, product purity, and cost per unit product between initial reactant concentration [$Z_0$] and catalyst concentration [$Cat_0$] at various fixed values of other parameters, obtained from the JAREX model of Fig.~\ref{fig:par}, trained with 40 samples.
  Confidence levels: mean pass (yellow), 70\% (light green), 95\% (dark green), fail (red).
  Last column: joint evaluation, intersecting the individual pass regions at each level.
  These are per-objective levels, not joint confidence levels.
  }
  \label{fig:interaction}
\end{figure}

Fig.~\ref{fig:interaction} shows the interaction effects between the initial reactant concentration [$Z_0$] and initial catalyst concentration [$Cat_0$] obtained from JAREX with other parameters fixed at various values.
From top to bottom, as the temperature rises from 300 to 310~K, the true joint pass region grows.
JAREX's joint maps reproduce this growth but are conservative at low temperature, where they show little joint pass region after 40 experiments.
With 80 experiments, these maps fill in and agree more closely with the truth (Supplementary Information, Section~S4.4).
This variation is primarily driven by the purity constraint, which fails at low temperature (Fig.~\ref{fig:par}) and responds sensitively to the other parameters (temperature $T$ and initial co-reactant concentration [$Y_0$]) held at different fixed values.
Such sensitivity highlights the importance of exploring interactions at multiple operating conditions rather than assuming a single interaction map is representative.

JAREX focuses sampling on the joint boundary, where experiments are most informative for defining the feasible operating region.
This targeted allocation of the experimental budget yields high-confidence characterization of the joint pass region while maintaining consistent boundary structure as the budget increases (Supplementary Information, Section~S4.4).
These analyses and visualizations are also available through the modules in the \obsd\ package.~\cite{obsidian}
Moreover, JAREX's iterative framework naturally supports targeted follow-up experiments, selecting new conditions that maximize the information gain and efficiently refine any remaining uncertainty where it matters most.

\section*{Conclusions}

In this paper, we introduce JAREX (Joint Acceptable Region EXploration), an acquisition function specifically designed for multi-objective process characterization using a Bayesian active-learning framework.
Methodologically, JAREX formulates process characterization as a joint boundary-learning problem.
Rather than optimizing a single response or characterizing each objective independently, it directly targets the joint pass region defined by simultaneous satisfaction of all quality specifications.
This aligns the acquisition rule with the actual objective of Quality-by-Design-driven characterization, namely, understanding the multidimensional edge of failure that governs acceptable process operation.
By combining an optimistic joint-feasibility mask with a randomized-straddle strategy, JAREX focuses on proposing experiments on the most informative parts of the joint boundary while accounting for the joint pass/fail condition across objectives.
Notably, we report JAREX as a modular open-source framework, implemented as part of a Python-based package (obsidian).

Moreover, our benchmark studies demonstrate that the JAREX approach improves learning efficiency for the process characterization task.
In the single-objective formulation, the underlying randomized-straddle strategy outperforms factorial DOE, space-filling methods, uncertainty sampling, and Upper Confidence Bound, and matches or exceeds feasibility Expected Improvement except on the two-dimensional function.
On a four-dimensional, three-objective kinetic model benchmark, JAREX reaches higher final joint accuracy than the tested blockwise extensions of randomized straddle, without requiring an objective order to be specified.
On two pharmaceutical case studies with up to 13 inputs and 7 objectives, it is more accurate than the tested designs and space-filling sampling in every setting and than the greedy strategy in most, with a lead that grows with the budget.
The gain is largest where much of each objective's boundary lies outside the joint pass region and the surrogates can recognize those parts.
Under batched experimentation (i.e., multiple experiments per iteration), JAREX more than halves the number of experimental rounds needed to match the accuracy of serial runs (i.e., one experiment per iteration), at the cost of more total experiments.

A practical strength of the framework presented here is its compatibility with the expected outputs from the traditional process characterization workflow.
Although JAREX is designed to learn the full multidimensional joint boundary, the trained surrogate can be leveraged to recover metrics such as proven acceptable ranges and critical multi-factor process interactions.
This makes the method directly relevant to existing pharmaceutical development workflows while providing a more data-efficient route to process understanding.

JAREX also does not require characterization to start from scratch: Algorithm~\ref{alg:jarex} accepts any initial dataset $\mathcal{D}_0$, so observations from earlier process-optimization campaigns can seed the surrogates directly.
Because JAREX and Bayesian-optimization acquisition functions can share the same surrogate models, the two could be combined within a single campaign, for example by alternating optimization and characterization batches or by using the JAREX surrogates as feasibility constraints in constrained Bayesian optimization, so that optimal conditions and their acceptable ranges are learned together, as Quality by Design intends.

Overall, JAREX and the accompanying open-source framework aim to bring to process characterization the same shift that Bayesian optimization enabled for process optimization: replacing largely static, uninformed experimentation with adaptive, model-driven decision-making.
In that sense, this work establishes a practical foundation for data-efficient, multivariate characterization of pharmaceutical processes.
Future efforts should focus on broader experimental validation across diverse unit operations, principled handling of correlated and heteroscedastic measurement noise, and integration with closed-loop robotic platforms to enable fully autonomous characterization workflows.

\section*{Author contributions}

XL derived and implemented the method and performed the benchmarking experiments.
AV and KS proposed and provided guidance on the project.
All authors contributed to manuscript revision.

\section*{Conflicts of interest}
The authors declare that they have no conflict of interest.

\section*{Data availability}

The single- and multi-objective characterization functions are available as part
of the open-source \href{https://github.com/MSDLLCPapers/obsidian}{\obsd} package.
The benchmarking scripts and all data supporting the findings of this study are openly available at
Zenodo, DOI:~\href{https://doi.org/10.5281/zenodo.21923037}{10.5281/zenodo.21923037}, which always resolves to the
latest version of the record.
The first version contains the benchmark campaigns of the original submission, the kinetic model scan, and the
process-characterization data, as JSON/CSV files with figure-reproduction scripts.
The latest version adds the campaigns of the revised benchmarks (the corrected kinetic benchmark and its
sensitivity studies, the single-objective benchmarks and baselines, and the chromatography, crystallization, and
integrated aspirin case studies, including the runs to 512 experiments), together with the
noise-free ground-truth sets and the design files used to score them.
The \obsd{} archive in the latest version is release 1.0.0 with the two corrections used for the revised benchmarks: the order of
averaging and clamping in the randomized straddle score, and the confidence levels after nonlinear output
transformations.

\section*{Acknowledgements}

XL acknowledges valuable discussions with Eugene Zakharov and Yingjie Chen at Merck \& Co., Inc., Rahway, NJ, USA.



\clearpage
\setcounter{section}{0}
\setcounter{figure}{0}
\setcounter{table}{0}
\setcounter{equation}{0}
\renewcommand{\thesection}{S\arabic{section}}
\renewcommand{\thefigure}{S\arabic{figure}}
\renewcommand{\thetable}{S\arabic{table}}
\renewcommand{\theequation}{S\arabic{equation}}
\begin{center}
  {\bfseries\LARGE Supplementary Information}\\[0.6em]
  {\large JAREX: An Acquisition Function for Multi-Objective Algorithmic Process Characterization}
\end{center}
\vspace{1em}

\section{Methodology Details}
\label{sec:si_methodology}

This section collects technical details that support the Methodology section in the main text but are not essential to the narrative.

\subsection{Feasibility Expected Improvement}
\label{sec:si_fei}

The main text mentions the Feasibility Expected Improvement (feasibility EI) acquisition function of Ierapetritou and coworkers~\cite{Metta2021} only conceptually.
Its explicit form is
\begin{equation}
  \label{eq:si_fei}
  \fei(x) = \sigma(x) \cdot \phi\!\left(\frac{\mu(x) - h}{\sigma(x)}\right),
\end{equation}
where $\mu(x)$ and $\sigma(x)$ are the predicted mean and standard deviation at $x$, $h$ is the threshold, and $\phi(\cdot)$ is the standard normal probability density function.
Conceptually, feasibility EI measures how likely a candidate is to lie near the threshold boundary: it is maximized when $\mu(x)$ is close to $h$ or when $\sigma(x)$ is large.
Sampling where feasibility EI is large therefore concentrates experiments both near the feasibility boundary and in regions of high uncertainty, efficiently classifying the parameter space into feasible ($\mu(x) > h$) and fail ($\mu(x) < h$) regions.
Although the original authors do not use the term ``characterization'', the goal of classifying the parameter space into pass/fail regions is the same.

\subsection{The \texorpdfstring{$\chi^2_2$}{chi-squared} Distribution}
\label{sec:si_chi2}

The central difficulty with the deterministic straddle acquisition function (Eq.~4 in the main text) is choosing a single value for $\beta^{1/2}$.
If $\beta^{1/2}$ is too small (too aggressive), the search concentrates too narrowly around the threshold and may miss other informative regions.
If it is too large (too conservative), it wastes samples in regions that are not critical for boundary identification.
The conventional value $\beta^{1/2} = 1.96$ turns out to be far too conservative in practice, which is what motivates randomizing $\beta$ rather than fixing it.

The randomized straddle acquisition function (Eq.~5 in the main text) instead draws $\beta$ from a $\chi^2_2$ distribution at every iteration.
The expectation of the multiplier $\beta^{1/2}$ is $\sqrt{2\pi}/2 \approx 1.25$, noticeably more aggressive than the traditional fixed choice $\beta^{1/2} = 1.96$.
Smaller $\beta^{1/2}$ values concentrate the search near the threshold for focused boundary refinement, while occasional larger values drawn from the long exponential tail drive broader exploration and prevent over-concentration on a single already-known boundary fragment.
This scheduling thus spends most iterations aggressively near the boundary while occasionally allowing broader exploration via the distribution's long tail.
Because $\chi^2_2$ has a simple closed-form CDF, $\beta$ can be sampled efficiently via inverse transform sampling at negligible cost per iteration.
Crucially, the schedule does not depend on the current sample count, so its adaptive character comes entirely from the distribution itself rather than from a hand-tuned annealing schedule.
Inatsu et al. provided the full theoretical analysis and empirical comparison with fixed-$\beta$ variants in Ref.~\citenum{Inatsu2024}.

\subsection{Aggregating Per-Objective Scores in JAREX}
\label{sec:si_softmin}

Once the per-objective randomized-straddle scores $\rstr_i(\hpoint) \ge 0$ are available, JAREX must reduce them to a single acquisition value at each $\hpoint$.
The main text uses a \textit{softmin}-weighted sum (Eq.~10 in the main text).
Several simpler alternatives are natural candidates:

\begin{itemize}
  \item \textbf{Maximum:} $\jarex^{\max}(\hpoint) = \max_i \rstr_i(\hpoint)$.
    This OR-style rule ignores objectives other than the locally most informative one and is essentially what the greedy baseline does.
  \item \textbf{Arithmetic mean:} $\jarex^{\mathrm{avg}}(\hpoint) = \frac{1}{m}\sum_i \rstr_i(\hpoint)$. Treats all objectives symmetrically and dilutes the joint signal with contributions from objectives that are confidently far from their thresholds.
  \item \textbf{Fixed weighted sum:} $\jarex^{w}(\hpoint) = \sum_i w_i \rstr_i(\hpoint)$ with user-chosen $w_i$. Requires prior knowledge about which objective deserves the most effort, which is precisely what characterization is trying to determine.
\end{itemize}

The softmin combination interpolates between these: as $\tau \to \infty$ it reduces to the arithmetic mean, and as $\tau \to 0$ it concentrates on $\min_i \rstr_i(\hpoint)$.
Most of the joint boundary consists of points where a single objective is at its threshold while the others pass with margin, so the behavior of the softmin on such points matters most.
Consider a candidate at which only objective $i$ is informative, with $\rstr_j \approx 0$ for all $j \neq i$.
Eq.~10 of the main text then reduces to
\begin{equation}
  \label{eq:si_single_active}
  \jarex(\hpoint) = \rstr_i \, \frac{e^{-\rstr_i/\tau}}{e^{-\rstr_i/\tau} + m - 1}.
\end{equation}
For $\rstr_i \ll \tau$ this is approximately $\rstr_i/m$, so candidates on single-objective segments of the joint boundary are retained and ranked by their informativeness for objective $i$, at about $1/m$ of the score they would receive if all $m$ objectives were equally informative.
Eq.~\ref{eq:si_single_active} increases with $\rstr_i$ up to $\rstr_i \approx 1.2\tau$ for $m = 3$ and decreases beyond it, so when scores are large the softmin deprioritizes candidates that are informative for only one objective.
For $m = 3$ and $\tau = 0.5$, a single-objective candidate scores $0.078$ at $\rstr_i = 0.5$ and $0.063$ at $\rstr_i = 1$, compared with $0.5$ and $1$ when all three objectives are equally informative.
Large scores occur where the surrogates are still uncertain, typically early in a campaign.
The softmin therefore first spends the budget on regions whose classification is ambiguous for several objectives and then, as the scores shrink below $\tau$, refines single-objective segments in a nearly mean-like regime.
The intersection logic that defines the joint boundary (Eq.~8 in the main text) is carried by the mask (Section~\ref{sec:si_softmask}), not by the aggregation.

The same expression indicates how JAREX scales with the number of objectives $m$.
In the mean-like regime, a candidate informative for $k$ of the $m$ objectives scores about $k\bar{\rstr}/m$, where $\bar{\rstr}$ is the typical score of those objectives, so adding objectives rescales all candidates without changing their ranking within each group, while increasingly favoring candidates near the intersections of several individual boundaries.
The mask becomes more restrictive as $m$ grows, since $\bopasj$ is an intersection of $m$ optimistic regions, and the computational cost grows linearly in $m$, with one surrogate and one straddle score per objective.
Flipping the sign of the exponent in Eq.~10 of the main text gives the corresponding softmax-weighted sum, $\sum_i \frac{e^{+\rstr_i/\tau}}{\sum_j e^{+\rstr_j/\tau}}\,\rstr_i(\hpoint)$, which is the smooth counterpart of the maximum aggregator above.
For $\rstr_i$ on the same scale as $\tau$, the exponential weighting collapses almost all mass onto whichever objective is locally largest, so the aggregator becomes effectively a hard maximum: small noise-driven fluctuations in $\rstr_i$ flip which objective dominates from one iteration to the next, leading to numerically unstable, abruptly-switching acquisition values.
In the closed-loop ablation on the kinetic benchmark (Section~\ref{sec:si_ablation}, Table~\ref{tbl:si_ablation}), unmasked softmax reaches a final joint Jaccard index of $0.702 \pm 0.077$, against $0.765 \pm 0.084$ for unmasked softmin.
With the mask, all four rules reach similar accuracy.
This comparison does not isolate numerical switching as the cause of the difference.
The default $\tau = 0.5$ is robust: for $\tau$ between 0.25 and 2, the final joint Jaccard index changes by at most 0.002, and no paired difference is distinguishable from zero.
JAREX is slightly more accurate at $\tau = 0.1$ (by 0.014; Section~\ref{sec:si_hyper}, Table~\ref{tbl:si_hyper}).

\subsection{Hard vs. Soft Mask and BoTorch Integration}
\label{sec:si_softmask}

The joint optimistic pass region $\bopasj$ (Eq.~8 in the main text) defines the set of candidates eligible for selection by JAREX.
A naive implementation uses $\bopasj$ as a hard indicator mask, multiplying the acquisition value by $1$ inside $\bopasj$ and $0$ outside.
This has two consequences that interact poorly with BoTorch's default candidate optimization.

First, when proposing new candidates, BoTorch~\cite{Balandat2020} launches multi-start gradient-based searches from the best of a large set of random samples to refine candidates to a numerical local optimum of the acquisition function.
A hard indicator mask is non-differentiable at the boundary of $\bopasj$ and identically zero outside it, so the gradient is undefined on the boundary and vanishes in the exterior.
Any starting point that falls outside $\bopasj$ therefore remains stuck at zero acquisition with no gradient signal to guide it back into the feasible region.
Second, $\bopasj$ is generally irregular and non-convex.
It cannot be expressed as a simple constraint such as a box or a polytope that BoTorch's constrained optimizers can handle natively.
These properties can leave infeasible starting points without a gradient for numerical refinement, even though searches initialized inside the region can still refine candidates.

The soft mask in Eq.~13 of the main text mitigates this difficulty.
It preserves $\bopasj$ exactly as the region where the mask equals $1$, but outside this region the mask decays smoothly with the slack via $\operatorname{sech}(k\, d(\point))$.
Where the underlying straddle aggregation is positive, the mask contributes a gradient toward the optimistic region.
It does not remove flat-zero regions created by the clamped straddle scores, so it does not guarantee a useful gradient from every starting point.
The smoothness parameter $k$ controls how steeply the mask decays.
We use $k = 2$ by default, the value used in all benchmarks.
Values from $k = 1$ to $k = 5$ give the same accuracy on the kinetic benchmark, whereas $k = 0.5$ and $k = 20$ lower the final joint Jaccard index by about 0.02 (Section~\ref{sec:si_hyper}, Table~\ref{tbl:si_hyper}).
The trade-off is a thin band immediately outside $\bopasj$ where the mask is non-zero but less than one, introducing a small amount of acquisition value in nominally infeasible territory.
In practice this band is narrow enough that selected candidates remain essentially within $\bopasj$, while the differentiability restores BoTorch's numerical candidate refinement.

The mask and the straddle scores use the same $\beta_i$.
In the exact form of Eq.~9 of the main text, every candidate outside $\bopas{i}$ therefore has $\rstr_i = 0$: if $\ucb_i(\hpoint) < h_i$, then $\mu_i < h_i$ and $h_i - \mu_i > \beta_i^{1/2}\sigma_i$, so the argument of the maximum is negative.
The converse does not hold, because $\rstr_i = 0$ also for candidates that confidently pass objective $i$.
The straddle scores alone therefore cannot tell a confident pass from a confident fail, and this directional information is what the mask adds.

In the implementation, the per-objective scores are estimated from posterior samples, as is standard for BoTorch's \textit{MCAcquisitionFunction}.
For posterior samples $f_i^{(s)}(\hpoint)$, $s = 1, \ldots, S$, let $\widehat\mu_i(\hpoint) = S^{-1}\sum_s f_i^{(s)}(\hpoint)$.
The implementation averages the sample contributions before clamping the per-objective score:
\begin{equation}
  \label{eq:si_mc_straddle}
  \rstr_i(\hpoint) \approx \max\left[\sqrt{\beta_i \pi/2}\, \frac{1}{S}\sum_{s=1}^{S} \left|f_i^{(s)}(\hpoint) - \widehat\mu_i(\hpoint)\right| - \left|\widehat\mu_i(\hpoint) - h_i\right|,\ 0\right].
\end{equation}
Because $\mathbb{E}\,|f_i - \mu_i| = \sigma_i\sqrt{2/\pi}$ for a Gaussian objective posterior, this estimator converges to Eq.~9 of the main text as the sample count increases.
Clamping each sample contribution before averaging would estimate a different acquisition, not an equivalent Monte Carlo implementation of Eq.~9.
The revised randomized-straddle and JAREX benchmarks use averaging followed by clamping, per objective and before aggregation.

\subsection{A Note on Mathematical Rigor}
\label{sec:si_rigor}

The single-objective randomized straddle algorithm of Inatsu et al.~\cite{Inatsu2024} builds on the level-set classification framework~\cite{Bryan2005,Gotovos2013} and comes with a clean theoretical guarantee.
Under standard Gaussian-process assumptions, its expected misclassification loss converges to zero at a sublinear rate.
This section explains exactly where that argument breaks down for JAREX and what could in principle be recovered under a restrictive additional assumption.
The argument is included because a reader interested in the theoretical status of the method deserves to see the concrete step that fails, not just a statement that ``the proof does not apply''.

Throughout this section we take the joint misclassification loss to be the sum of per-objective losses, $\ell^{\mathrm{joint}}_t(x)=\sum_{i=1}^{m}\ell_{t,i}(x)$.
Equivalently, joint misclassification at $x$ implies misclassification on at least one objective, so the joint loss is set-theoretically dominated by the sum of per-objective losses.

The Inatsu proof combines four ingredients.
\begin{enumerate}[label=(\roman*)]
  \item A Gaussian-process concentration bound that places $f(x)$ inside $[\mu_{t-1}(x)\pm\beta_\delta^{1/2}\sigma_{t-1}(x)]$ with high probability, where $\beta_\delta=2\log(1/\delta)$.
  \item A pointwise inequality $\ell_t(x)\le a_{t-1,\delta}(x)$ between the misclassification loss and the straddle score;
  \item The distributional identity that, when $\delta\sim U(0,1)$, the random variable $2\log(1/\delta)$ is exactly $\chi^2_2$-distributed, which lets one swap the expectation over $\delta$ for an expectation over $\beta_t\sim\chi^2_2$ and, combined with the selection rule $x_t=\arg\max_x a_{t-1}(x)$, upgrades the pointwise bound in (ii) to an expected bound evaluated at the selected point $x_t$;
  \item A Cauchy--Schwarz step combined with the maximum-information-gain bound $\sum_{t}\sigma_{t-1}^2(x_t)\le C\gamma_t$, which converts the per-iteration bound into a sublinear cumulative rate.
\end{enumerate}
The critical step for JAREX is (iii).
In the single-objective case the selection rule maximizes the same scalar quantity $a_{t-1}$ that upper-bounds the loss, so the inequality can be evaluated at the selected point and then summed over iterations.
In JAREX the selection rule maximizes $M(\point)\cdot\jarex(\point)$, a data-dependent softmin-weighted combination of $m$ per-objective straddle scores multiplied by a smoothly masked indicator of $\bopasj$.
Neither the softmin-weighted combination nor the masked acquisition equals the single-objective straddle score of any individual objective, and in general the selected point $\pointnext$ is not the argmax of $a_{i,t-1}$ for any fixed $i$.
As a result, the per-objective inequality in step (ii) cannot be pushed to the selected point through step (iii), and the Cauchy--Schwarz / information-gain argument in step (iv) can no longer be applied per objective at $\pointnext$ to yield a sublinear rate.
Per-objective ingredients (i) and (ii) themselves remain valid, and the joint misclassification loss is still bounded pointwise by the sum of per-objective losses by the set-inclusion argument above.
What is missing is the link between these pointwise bounds and the specific point JAREX chooses to sample.

A conditional, much weaker statement can be recovered only by restoring that link by force.
If we additionally assumed that every objective is queried sufficiently often, for example by appending a round-robin fallback that guarantees each objective receives a non-vanishing share of the iteration budget, then each per-objective surrogate would satisfy Inatsu's bound on its own subsequence of iterations.
The joint misclassification loss, bounded above by the sum of per-objective losses, would then inherit a sublinear rate via a union bound, with the constants degraded by at most a factor of $m$ and by the fraction of iterations assigned to each objective.
This is a genuine but essentially trivial guarantee: it only says that if we \emph{force} uniform coverage over objectives, no objective can be arbitrarily neglected and each individual boundary is eventually characterized.
It does not capture any of the joint-boundary behavior that motivates JAREX in the first place, and it contradicts the intended behavior of the algorithm, which is to deprioritize regions confidently outside $\bopasj$ and objectives whose boundary is locally uninformative.
Imposing uniform per-objective coverage would recover a formal bound only by disabling the feature of JAREX that actually delivers the empirical gains in Section~3 of the main text, so we do not pursue this route.

A tighter analysis that accommodates the softmin weighting and the soft mask, and that targets the joint misclassification loss directly rather than reducing it to per-objective losses, is what a full theoretical treatment of JAREX would require.
We view this as an interesting but substantial direction for future work: the pointwise loss control step (ii) remains a promising starting point, but a new argument is needed to handle a selection rule that depends jointly on all $m$ surrogates and on a data-dependent feasibility mask.

\section{Additional Benchmark Details}
\label{sec:si_implementation}

This section collects technical details on implementation choices referenced in the main text that apply to both single-objective and multi-objective characterization benchmarks.

The Jaccard index is a set-based similarity metric that scores only whether each parameter set is assigned the correct pass/fail label, making it particularly appropriate for process characterization, where the primary goal is to correctly identify the pass/fail boundary rather than to minimize prediction error uniformly across the parameter space.
Traditional metrics such as mean absolute error (MAE) or root mean squared error (RMSE) weight all predictions equally and are therefore dominated by contributions from regions far from the boundary, where accuracy is less critical.
A boundary-focused method may report a higher overall MAE while being substantially more accurate at the pass/fail threshold, which is the only region that affects characterization outcomes.
A high Jaccard score also implicitly rewards global coherence of the response surface, since it is unlikely that a model produces accurate boundary predictions while being grossly wrong elsewhere.

For all test cases, Gaussian process surrogate models were implemented using BoTorch~\cite{Balandat2020} with the default Matérn kernel.
Model hyperparameters were optimized using \texttt{scipy.optimize} with BFGS-family methods (L-BFGS-B for box-constrained optimization) with 10 multi-start restarts to ensure robust convergence.
Initial hyperparameter values for each restart were sampled randomly within physically reasonable ranges to avoid local minima.
Jaccard indices were calculated by discretizing the parameter space using quasi-random Sobol sequences, with the number of evaluation points determined adaptively based on the overall uncertainty of the surrogate model to ensure accurate boundary representation.
The adaptive strategy increases the point density when the GP posterior variance is high near decision boundaries.

\section{Single-Objective Characterization Details}

This section provides detailed descriptions of the benchmark test cases used to evaluate the performance of the randomized straddle algorithm for single-objective characterization.
We describe the analytical test functions used for benchmarking, starting with the one-dimensional double-well potential illustrated in Figure 1 of the main text.

\subsection{Single-Objective Analytical Functions}

\subsubsection{Double-Well Potential (1D)}

By design (Eq.~5 in the main text), the randomized straddle algorithm selects candidates that are either near the threshold or have high uncertainty.
This preference can easily be visualized using a classical one-dimensional double-well potential, which has two local minima separated by a barrier.
The double-well potential is defined as,
\begin{align}
  \label{eq:si_dwp}
  f(x) = ax^4 - bx^2 + c.
\end{align}
We chose $a = 1$, $b = \frac{1}{2}$, $c = \frac{b^2}{4a}$ for the test function.
$-f(x)$ is shown as the blue curve in Figure 1 of the main text.
We flipped the sign purely for visualization purposes.
The threshold $h = -\frac{c}{2}$ is shown as the horizontal gray dashed line.
2 initial observations are added at the beginning of the campaign.

To validate performance across higher-dimensional spaces, we use three additional standard analytical test functions: the six-hump camel function (2D), Hartmann 4D function (4D), and Hartmann 6D function (6D).
These functions are commonly used benchmarks in the optimization literature and provide increasing levels of complexity for testing boundary identification algorithms.
Each function is described below.

\subsubsection{Six-Hump Camel Function}

The six-hump camel function is a two-dimensional multimodal test function. The standard form is defined as,
\begin{equation}
    f(x_1, x_2) = \left(4 - 2.1x_1^2 + \frac{x_1^4}{3}\right)x_1^2 + x_1x_2 + (-4 + 4x_2^2)x_2^2,
\end{equation}
where $x_1 \in [-2, 2]$ and $x_2 \in [-1, 1]$. For our characterization benchmarking, we apply a transformation to convert this minimization problem to a maximization problem, $y(x_1, x_2) = 11.0316 - f(x_1, x_2)$. We use a threshold value of $h = 10.6316$ and classify the parameter space into regions where $y(x_1, x_2) \ge h$ (pass) and $y(x_1, x_2) < h$ (fail). Initial sampling consists of 5 observations generated using Latin hypercube sampling (LHS).~\cite{surjano_camel6}

\subsubsection{Hartmann 4D Function}

The Hartmann 4D function is a four-dimensional multimodal test function built from the sum
\begin{equation}
    S(\mathbf{x}) = \sum_{i=1}^{4} \alpha_i \exp\left(-\sum_{j=1}^{4} A_{ij}(x_j - P_{ij})^2\right),
\end{equation}
where $\mathbf{x} = (x_1, x_2, x_3, x_4) \in [0, 1]^4$, and the parameters are given by,
\begin{align*}
    \boldsymbol{\alpha} &= (1.0, 1.2, 3.0, 3.2)^T \\
    \mathbf{A} &= \begin{pmatrix}
        10 & 3 & 17 & 3.5 \\
        0.05 & 10 & 17 & 0.1 \\
        3 & 3.5 & 1.7 & 10 \\
        17 & 8 & 0.05 & 10
    \end{pmatrix} \\
    \mathbf{P} &= 10^{-4} \begin{pmatrix}
        1312 & 1696 & 5569 & 124 \\
        2329 & 4135 & 8307 & 3736 \\
        2348 & 1451 & 3522 & 2883 \\
        4047 & 8828 & 8732 & 5743
    \end{pmatrix}
\end{align*}
The benchmark response is $y(\mathbf{x}) = (S(\mathbf{x}) - 1.1)/0.839$, the negative of the rescaled Hartmann 4D function of Surjanovic and Bingham, so that the characterization seeks large values. It ranges from $-1.31$ to its global maximum of $3.135$ at $\mathbf{x}^* \approx (0.187, 0.194, 0.558, 0.265)$. We use a threshold of $h = 1.5$ and classify the parameter space into regions where $y \ge h$ (pass, 11\% of the parameter space) and $y < h$ (fail). Initial sampling consists of 10 observations generated using LHS.~\cite{surjano_hart4}

\subsubsection{Hartmann 6D Function}

The Hartmann 6D function is a six-dimensional multimodal test function built analogously to the 4D version from
\begin{equation}
    S(\mathbf{x}) = \sum_{i=1}^{4} \alpha_i \exp\left(-\sum_{j=1}^{6} A_{ij}(x_j - P_{ij})^2\right),
\end{equation}
where $\mathbf{x} = (x_1, x_2, x_3, x_4, x_5, x_6) \in [0, 1]^6$, with parameters,
\begin{align*}
    \boldsymbol{\alpha} &= (1.0, 1.2, 3.0, 3.2)^T \\
    \mathbf{A} &= \begin{pmatrix}
        10 & 3 & 17 & 3.5 & 1.7 & 8 \\
        0.05 & 10 & 17 & 0.1 & 8 & 14 \\
        3 & 3.5 & 1.7 & 10 & 17 & 8 \\
        17 & 8 & 0.05 & 10 & 0.1 & 14
    \end{pmatrix} \\
    \mathbf{P} &= 10^{-4} \begin{pmatrix}
        1312 & 1696 & 5569 & 124 & 8283 & 5886 \\
        2329 & 4135 & 8307 & 3736 & 1004 & 9991 \\
        2348 & 1451 & 3522 & 2883 & 3047 & 6650 \\
        4047 & 8828 & 8732 & 5743 & 1091 & 381
    \end{pmatrix}
\end{align*}
The benchmark response is $y(\mathbf{x}) = (2.58 + S(\mathbf{x}))/1.94$, again the negative of the rescaled function of Surjanovic and Bingham. It ranges from $1.33$ to its global maximum of $3.042$ at $\mathbf{x}^* \approx (0.202, 0.150, 0.477, 0.275, 0.312, 0.657)$. We use a threshold of $h = 1.5$ (pass, $y \ge h$: 25\% of the parameter space). Initial sampling consists of 15 observations generated using LHS.~\cite{surjano_hart6}

\subsection{Single-Objective Benchmarks with Noise}

\begin{figure}[H]
  \centering
  \includegraphics[width=0.6\textwidth]{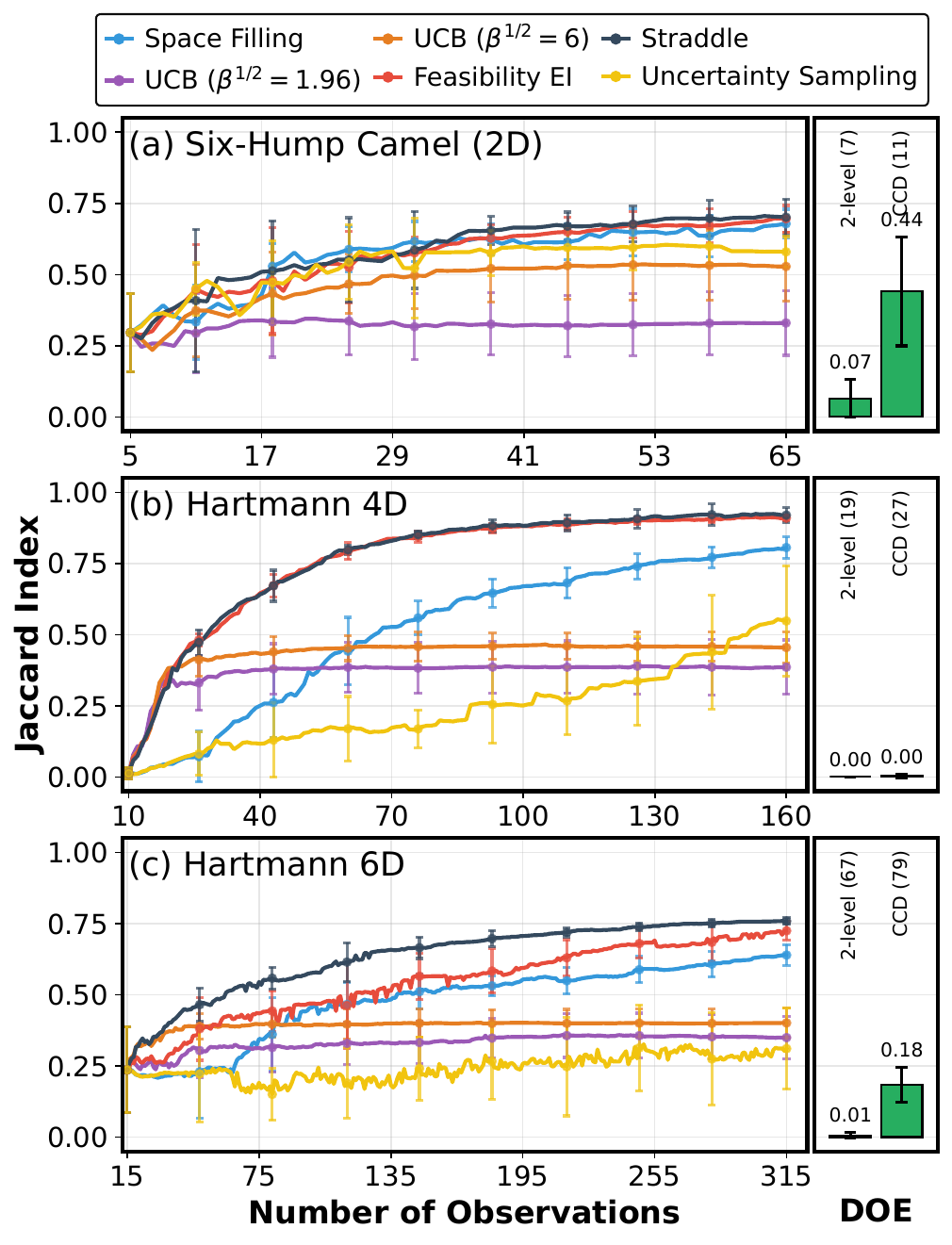}
  \caption{Benchmark results for single-objective characterization with 5\% Gaussian noise tested on various analytical functions: six-hump camel (2D), Hartmann 4D (4D), and Hartmann 6D (6D).
  A Gaussian noise $\mathcal{N}(0, \sigma^2)$ with $\sigma = 0.05 \times |y|$ is added to all function outputs.
  Each method is run with 10 independent campaigns with different random seeds, and the mean and standard deviation of the Jaccard index are plotted.
  All methods show some performance degradation compared to the noise-free case (Fig. 2 in main text), but the relative performance ranking remains consistent.
  The randomized straddle algorithm continues to demonstrate superior efficiency and stability, particularly for high-dimensional problems.
  While convergence to near-perfect characterization (approaching 100\% Jaccard index) is achievable with sufficiently large sample budgets, economical sample sizes for practical method development are used here.
  The side panel (DOE) in each transformation reports the RSM designs, a full two-level factorial and a central composite design (CCD), as single-design Jaccard values, with run counts in parentheses.
  }
  \label{fig:so_noise}
\end{figure}

To evaluate the robustness of the randomized straddle algorithm under realistic experimental conditions, we also benchmarked all methods with 5\% Gaussian noise added to the analytical function outputs.
The noise is modeled as $\mathcal{N}(0, \sigma^2)$ where $\sigma = 0.05 \times |y|$, representing a common level of experimental noise in real-world applications.
Fig.~\ref{fig:so_noise} shows the benchmark results with noise for all analytical functions.

\section{Multi-Objective Characterization Details}

\subsection{Kinetic Model}

The kinetic model consists of a main catalytic pathway with a competing reversible side reaction:
\begin{align}
    \ce{Y + Z &->[$k_{\text{main}}(T, \text{Cat})$] P} \\
    \ce{2Y + Z &<=>[$k_{\text{side,fwd}}(\text{Cat})$][$k_{\text{side,rev}}(T)$] X}
\end{align}
where Z is the reactant, Y is the co-reactant, P is the desired product, X is an undesired impurity, and Cat represents the catalyst. The main reaction exhibits both temperature and catalyst dependence (high activation energy), while the side forward reaction depends only on catalyst (no temperature dependence) and the side reverse reaction depends only on temperature (no catalyst dependence).

\subsubsection{Rate Expressions}

The reaction kinetics is described by the following system of ordinary differential equations (ODEs),
\begin{align}
    \frac{d[\text{Z}]}{dt} &= -r_{\text{main}} - r_{\text{sf}} + r_{\text{sr}} \\
    \frac{d[\text{Y}]}{dt} &= -r_{\text{main}} - 2 \cdot r_{\text{sf}} + 2 \cdot r_{\text{sr}} \\
    \frac{d[\text{P}]}{dt} &= r_{\text{main}} \\
    \frac{d[\text{X}]}{dt} &= r_{\text{sf}} - r_{\text{sr}} \\
    \frac{d[\text{Cat}]}{dt} &= -k_{\text{decay}} \cdot [\text{Cat}].
\end{align}
The stoichiometry reflects that the main reaction consumes both Y and Z, while the side reaction involves 2 moles of Y per mole of X.

The main reaction rate (Y + Z → P) follows a Langmuir-Hinshelwood dual-site mechanism with substrate inhibition:
\begin{equation}
    r_{\text{main}} = k_{\text{main}}(T, [\text{Cat}]) \cdot \frac{[\text{Y}] \cdot [\text{Z}]}{(1 + [\text{Y}]/K_{Y,\text{main}} + [\text{Z}]/K_{Z,\text{main}})^2},
\end{equation}
where the rate constant exhibits Arrhenius temperature dependence and catalyst substrate inhibition:
\begin{equation}
    k_{\text{main}}(T, [\text{Cat}]) = A_{\text{main}} \exp\left(-\frac{E_{a,\text{main}}}{RT}\right) \cdot \frac{[\text{Cat}]}{K_{m,\text{main}} + [\text{Cat}] + \frac{[\text{Cat}]^2}{K_{i,\text{main}}}},
\end{equation}
with $A_{\text{main}} = 6 \times 10^6$ L/(mol·min) as the pre-exponential factor, $E_{a,\text{main}} = 40$ kJ/mol as the activation energy (high, providing temperature-dependent purity control), $K_{m,\text{main}} = 0.01$ mol/L as the Michaelis constant, $K_{i,\text{main}} = 0.15$ mol/L as the substrate inhibition constant, $K_{Y,\text{main}} = 0.5$ mol/L and $K_{Z,\text{main}} = 1.2$ mol/L as the competitive adsorption constants for Y and Z, respectively. The squared denominator reflects a dual-site mechanism where both Y and Z compete for catalyst sites. $R = 8.314$ J/(mol·K) is the gas constant.

The forward side reaction rate (2Y + Z → X) follows a Langmuir-Hinshelwood mechanism with fractional order in Y and catalyst substrate inhibition (no temperature dependence):
\begin{equation}
    r_{\text{sf}} = k_{\text{sf}}([\text{Cat}]) \cdot \frac{[\text{Y}]^{n_{Y,\text{sf}}} \cdot [\text{Z}]}{(1 + [\text{Y}]/K_{Y,\text{sf}} + [\text{Z}]/K_{Z,\text{sf}})^2},
\end{equation}
where the rate constant depends only on catalyst loading (no temperature dependence):
\begin{equation}
    k_{\text{sf}}([\text{Cat}]) = k_{\text{sf}}^{\text{max}} \cdot \frac{[\text{Cat}]}{K_{m,\text{sf}} + [\text{Cat}] + \frac{[\text{Cat}]^2}{K_{i,\text{sf}}}},
\end{equation}
with $k_{\text{sf}}^{\text{max}} = 4.5$ L$^2$/(mol$^2$·min) as the maximum rate at saturating catalyst, $K_{m,\text{sf}} = 0.015$ mol/L as the Michaelis constant, $K_{i,\text{sf}} = 0.35$ mol/L as the substrate inhibition constant, $K_{Y,\text{sf}} = 4.0$ mol/L and $K_{Z,\text{sf}} = 1.5$ mol/L as the competitive adsorption constants, and $n_{Y,\text{sf}} = 2.0$ as the fractional order in Y. The squared denominator reflects dual-site competition. The absence of temperature dependence makes this pathway relatively more favorable at low temperatures.

The reverse side reaction rate (X → 2Y + Z) depends only on temperature (no catalyst dependence) with product inhibition:
\begin{equation}
    r_{\text{sr}} = k_{\text{sr}}(T) \cdot \frac{[\text{X}]}{1 + [\text{Y}]/K_{Y,\text{rev}} + [\text{Z}]/K_{Z,\text{rev}}},
\end{equation}
where the rate constant follows Arrhenius temperature dependence:
\begin{equation}
    k_{\text{sr}}(T) = A_{\text{sr}} \cdot \exp\left(-\frac{E_{a,\text{sr}}}{R T}\right),
\end{equation}
with $A_{\text{sr}} = 5 \times 10^6$ min$^{-1}$ as the pre-exponential factor, $E_{a,\text{sr}} = 32$ kJ/mol as the activation energy, $K_{Y,\text{rev}} = 1.0$ mol/L and $K_{Z,\text{rev}} = 0.8$ mol/L as product inhibition constants. The relatively low activation energy (compared to the main reaction) and absence of catalyst dependence makes this pathway relatively more favorable at higher temperatures, enabling impurity recycling.

The following table summarizes all kinetic parameters used in the model.
\begin{table}[H]
\centering
\begin{tabular}{llp{5.5cm}}
\toprule
\textbf{Parameter} & \textbf{Value} & \textbf{Description} \\
\midrule
\multicolumn{3}{l}{\textit{Main reaction: Y + Z $\rightarrow$ P}} \\
$A_{\text{main}}$ & $6 \times 10^6$ L/(mol·min) & Pre-exponential factor \\
$E_{a,\text{main}}$ & 40 kJ/mol & Activation energy \\
$K_{m,\text{main}}$ & 0.01 mol/L & Michaelis constant \\
$K_{i,\text{main}}$ & 0.15 mol/L & Substrate inhibition constant \\
$K_{Y,\text{main}}$ & 0.5 mol/L & Y competitive adsorption constant \\
$K_{Z,\text{main}}$ & 1.2 mol/L & Z competitive adsorption constant \\
\midrule
\multicolumn{3}{l}{\textit{Side forward reaction: 2Y + Z $\rightarrow$ X}} \\
$k_{\text{sf}}^{\text{max}}$ & 4.5 L$^2$/(mol$^2$·min) & Maximum rate at saturating catalyst \\
$K_{m,\text{sf}}$ & 0.015 mol/L & Michaelis constant \\
$K_{i,\text{sf}}$ & 0.35 mol/L & Substrate inhibition constant \\
$K_{Y,\text{sf}}$ & 4.0 mol/L & Y competitive adsorption constant \\
$K_{Z,\text{sf}}$ & 1.5 mol/L & Z competitive adsorption constant \\
$n_{Y,\text{sf}}$ & 2.0 & Fractional order in Y \\
\midrule
\multicolumn{3}{l}{\textit{Side reverse reaction: X $\rightarrow$ 2Y + Z}} \\
$A_{\text{sr}}$ & $5 \times 10^6$ min$^{-1}$ & Pre-exponential factor \\
$E_{a,\text{sr}}$ & 32 kJ/mol & Activation energy \\
$K_{Y,\text{rev}}$ & 1.0 mol/L & Y product inhibition constant \\
$K_{Z,\text{rev}}$ & 0.8 mol/L & Z product inhibition constant \\
\midrule
\multicolumn{3}{l}{\textit{Catalyst decay}} \\
$k_{\text{decay}}$ & $2 \times 10^{-4}$ min$^{-1}$ & First-order decay constant \\
\bottomrule
\end{tabular}
\end{table}

\subsubsection{Objectives}

The three objectives are calculated from the final concentrations as follows:

\begin{align}
    \text{Z conversion rate} &= \frac{[\text{Z}]_0 - [\text{Z}]_{\text{final}}}{[\text{Z}]_0} \quad \text{[maximize]} \\
    \text{Product P purity} &= \frac{[\text{P}]_{\text{final}}}{[\text{P}]_{\text{final}} + [\text{X}]_{\text{final}}} \quad \text{[maximize]} \\
    \text{Unit product cost} &= \frac{u_Z [\text{Z}]_0 + u_Y [\text{Y}]_0 + u_{\text{Cat}} [\text{Cat}]_0 + u_{\text{sol}}}{[\text{P}]_{\text{final}}} \quad \text{[minimize]}
\end{align}

The unit cost represents the total cost per mole of product P formed (\$/mol), accounting for:
\begin{itemize}
    \item Material costs: reactant Z ($u_Z = 10$ \$/mol), co-reactant Y ($u_Y = 45$ \$/mol), and catalyst ($u_{\text{Cat}} = 6,500$ \$/mol)
    \item Solvent cost: fixed cost per batch ($u_{\text{sol}} = 1$ \$/batch)
\end{itemize}
All costs are normalized by the amount of product P formed.
Note that temperature T does not directly enter the cost function but affects the kinetics and thus the final product yield.

For multi-objective characterization benchmarking, we define the following target specifications:
\begin{itemize}
    \item \textbf{Purity}: Product purity $\geq 99.3\%$
    \item \textbf{Conversion}: Reactant Z conversion $\geq 90\%$
    \item \textbf{Cost}: Unit product cost $\leq 570$ \$/mol P
\end{itemize}
These thresholds define a pass/fail classification of the four-dimensional parameter space.
The goal of characterization is to identify the boundary between the pass region (where all three constraints are satisfied) and the fail region (where at least one constraint is violated).

\subsubsection{Integration Details}

The system of ordinary differential equations is integrated numerically using the BDF (Backward Differentiation Formula) method from \texttt{scipy.integrate.solve\_ivp}.
The integration time span is $t \in [0, t_{\text{final}}]$ minutes, where $t_{\text{final}}$ is typically 120 min, with early termination if 99.9\% of substrate Z is consumed.
Initial conditions are specified by the decision variables:
\begin{itemize}
    \item Reactant $Z$ initial concentration $[\text{Z}]_0 \in [0.8, 3.0]$ mol/L
    \item Co-reactant $Y$ initial concentration $[\text{Y}]_0 \in [0.8, 6.0]$ mol/L
    \item Reaction temperature $T \in [273, 333]$ K
    \item Catalyst loading $[\text{Cat}]_0 \in [0.005, 0.2]$ mol/L
\end{itemize}
with fixed initial values $[\text{P}]_0 = 0$ mol/L and $[\text{X}]_0 = 0$ mol/L.
The solver uses a relative tolerance of $10^{-6}$ and an absolute tolerance of $[\text{Z}]_0 \times 10^{-7}$ to ensure accurate integration.
The output consists of species concentrations at the final time point.

\subsubsection{Measurement Noise}

Measurement noise in the multi-objective benchmark is applied to the simulated final concentrations rather than to the objectives.
Each final concentration $c \in \{[\text{Z}], [\text{Y}], [\text{P}], [\text{X}], [\text{Cat}]\}$ is multiplied by $1 + 0.05\,\epsilon_c$, with $\epsilon_c \sim \mathcal{N}(0, 1)$ drawn independently for each species and each experiment, and the objectives are then computed from the perturbed concentrations.
This differs from the single-objective benchmarks, where noise is added directly to the response with standard deviation $0.05\,|y|$.
Propagating concentration noise through the objective definitions makes the noise heteroscedastic and different for each objective.
To first order near the thresholds, the noise standard deviation is about 5\% of the value for cost (about \$28 at the \$570 threshold), because cost scales as $1/[\text{P}]_{\text{final}}$.
It is about 0.5 percentage points for conversion, because $[\text{Z}]_{\text{final}} \approx 0.1\,[\text{Z}]_0$ at 90\% conversion, and about 0.05 percentage points for purity, because $[\text{X}]_{\text{final}}/[\text{P}]_{\text{final}} \approx 0.007$ at 99.3\% purity.
Purity and cost share the noise in $[\text{P}]_{\text{final}}$ and are therefore correlated.
The ground-truth pass regions used for the Jaccard index are computed from the noise-free model.

\subsection{Transformation Strategies}
\label{sec:si_log_transform}

The kinetic model objectives (purity, conversion, cost) exhibit highly skewed distributions, making standardization alone less effective for Gaussian process modeling.
This section provides detailed analysis of transformation strategies and their impact on method performance.

Two transformation strategies were evaluated:
\begin{itemize}
    \item \textbf{Standard transformation}: Simple standardization (zero mean, unit variance) applied to each objective independently.
    \item \textbf{Log transformation}: Logit transformation for purity and conversion (bounded objectives) combined with log transformation for cost (positive objective), followed by standardization.
\end{itemize}

The log-based transformations help stabilize the surrogate model by mapping the highly skewed objective distributions to more Gaussian-like distributions, which better match the GP modeling assumptions.

\begin{figure}[H]
  \centering
  \includegraphics[width=0.6\textwidth]{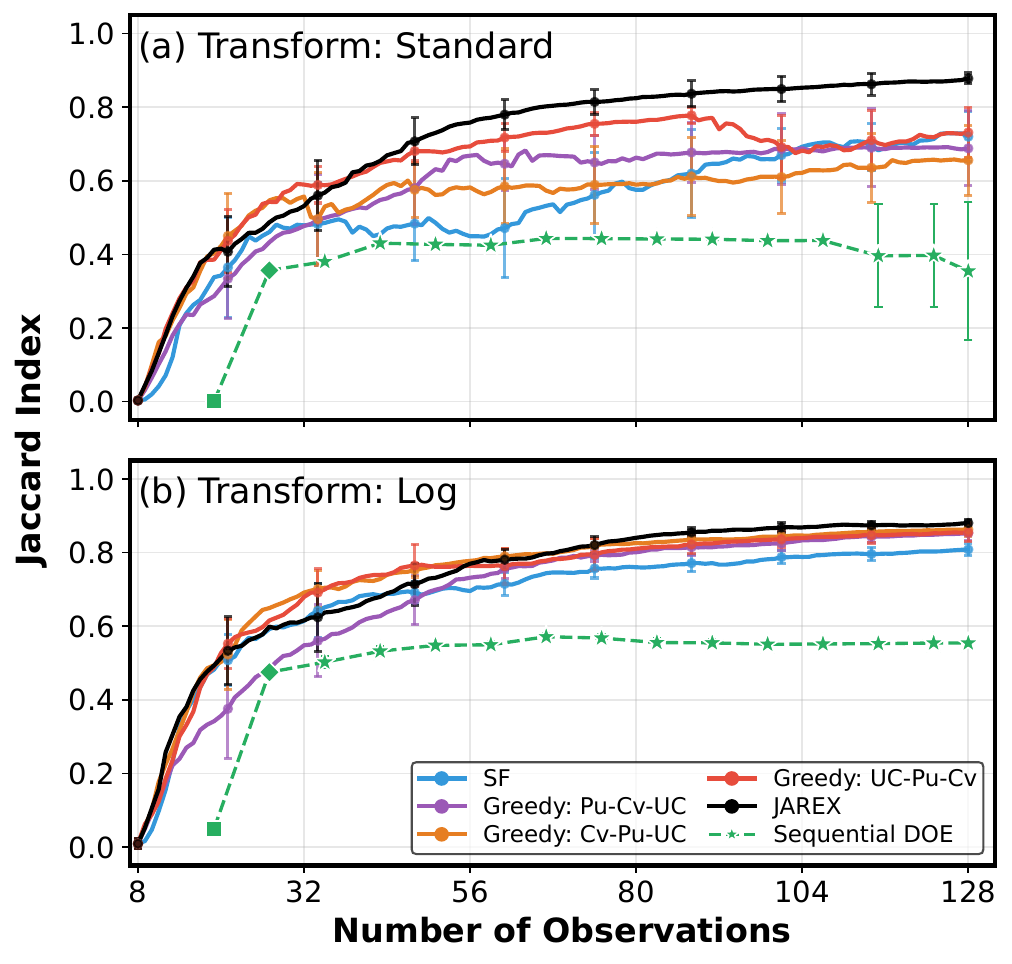}
  \caption{Full benchmark results for sequential multi-objective characterization showing both standard (panel a) and log (panel b) transformations.
  JAREX (black) maintains robust performance across both transformation strategies, while other methods show stronger dependence on transformation choice.
  Dashed green: the sequential design of Section~\ref{sec:si_baselines}, scored with the same GP surrogate, whose first two points are the full two-level factorial (square, 19 runs) and the central composite design (CCD; diamond, 27 runs), followed by I-optimal blocks of eight runs (stars) up to 128.
  }
  \label{fig:si_serial}
\end{figure}

\begin{figure}[H]
  \centering
  \includegraphics[width=0.6\textwidth]{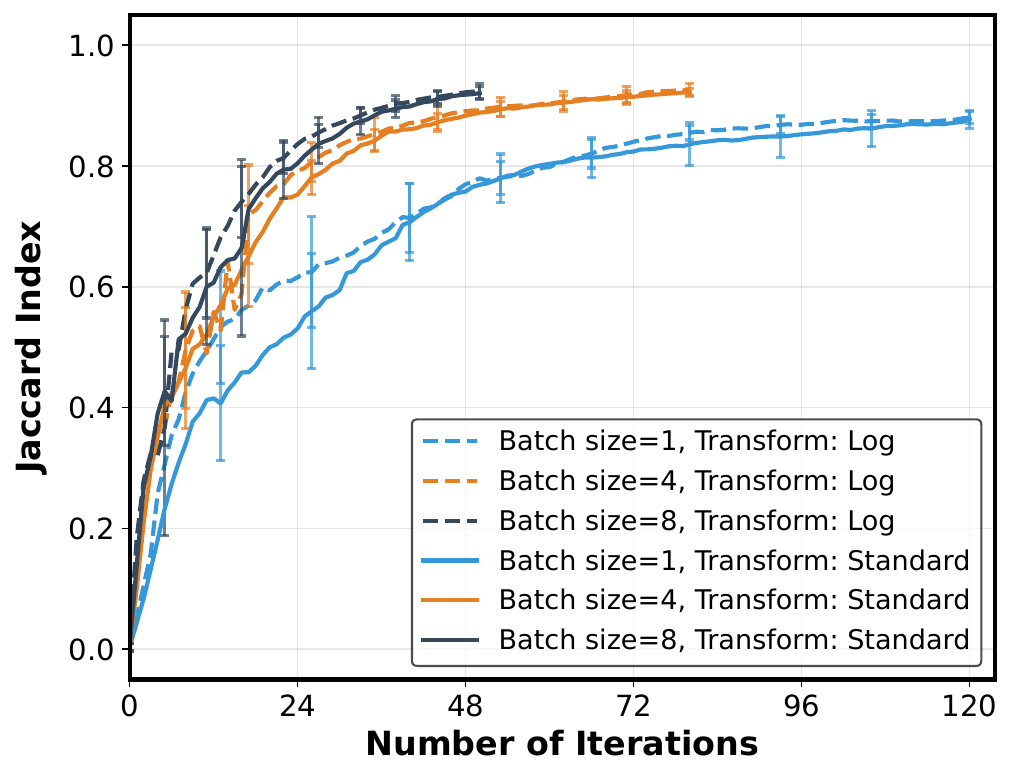}
  \caption{Full batched characterization results comparing serial ($q=1$), moderate ($q=4$), and high ($q=8$) batch sizes with the standard (solid lines) and log (dashed lines) transformations.
  Batching provides consistent benefits across transformation strategies.
  }
  \label{fig:si_batch}
\end{figure}

Full benchmark results with both transformation strategies are shown in Fig.~\ref{fig:si_serial} for serial characterization and Fig.~\ref{fig:si_batch} for batched characterization.
The log transformation significantly improves the performance of the RSM designs, SF, and greedy randomized straddle methods.
However, JAREX demonstrates remarkable numerical stability, performing equally well with either transformation (final joint Jaccard index 0.878 with the standard and 0.880 with the log transformation).
This robustness is a desirable property for real-world applications where the optimal transformation strategy may not be known beforehand.

\begin{figure}[H]
  \centering
  \includegraphics[width=0.6\textwidth]{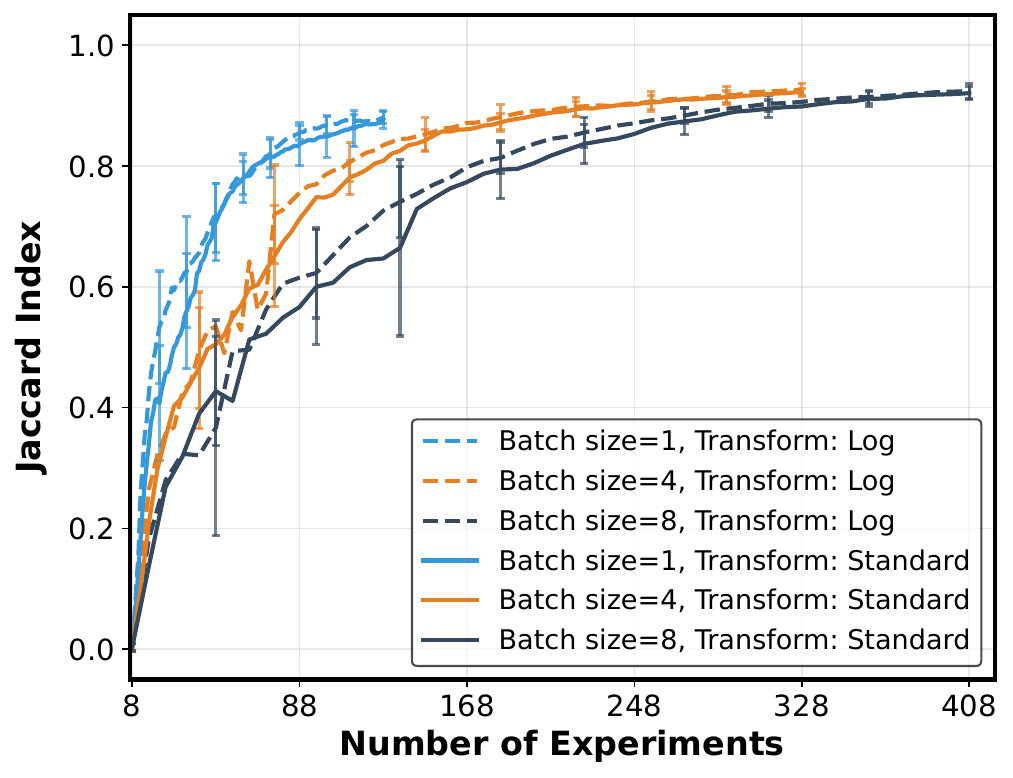}
  \caption{The campaigns of Fig.~\ref{fig:si_batch} plotted against the total number of experiments, including the eight initial experiments.
  Per experiment, the serial campaign is the most sample-efficient: at 128 experiments, the joint Jaccard index is 0.878 for $q=1$, 0.809 for $q=4$, and 0.647 for $q=8$ (standard transformation).
  The batched campaigns reach the final serial accuracy after 192 ($q=4$) and 288 ($q=8$) experiments, in 46 and 35 rounds instead of 120.
  }
  \label{fig:si_batch_experiments}
\end{figure}

\subsection{Quantitative Analysis of Multi-Objective Trade-offs}
\label{sec:si_jaccard}

As discussed in the main text, the joint Jaccard index can exceed the worst individual Jaccard index, which appears counter-intuitive.
This section provides a quantitative breakdown of the underlying mechanisms.

The Jaccard index is defined as $J = \mathrm{TP}/(\mathrm{TP} + \mathrm{FP} + \mathrm{FN})$.
Because the denominator counts errors over the specific region being evaluated, the joint Jaccard index is \emph{not} the product of the individual ones,
\begin{equation}
  J(\hat{A}_1 \cap \hat{A}_2 \cap \hat{A}_3,\; A_1 \cap A_2 \cap A_3) \;\neq\; \prod_{i} J(\hat{A}_i, A_i),
\end{equation}
even when the objectives are fully independent.
For the corrected serial JAREX campaigns at 128 experiments (standard transformation, means over 10 seeds), the individual indices are $J_\text{purity} = 0.879$, $J_\text{conv} = 0.865$, and $J_\text{cost} = 0.855$, whereas $J_\text{joint} = 0.878$.
The product of these mean individual indices is about 0.650, not the mean joint index, and the joint index exceeds the cost index.
These quantities describe different classification targets, so there is no requirement that the joint index be lower than every individual index.

\begin{figure}[H]
  \centering
  \includegraphics[width=0.6\textwidth]{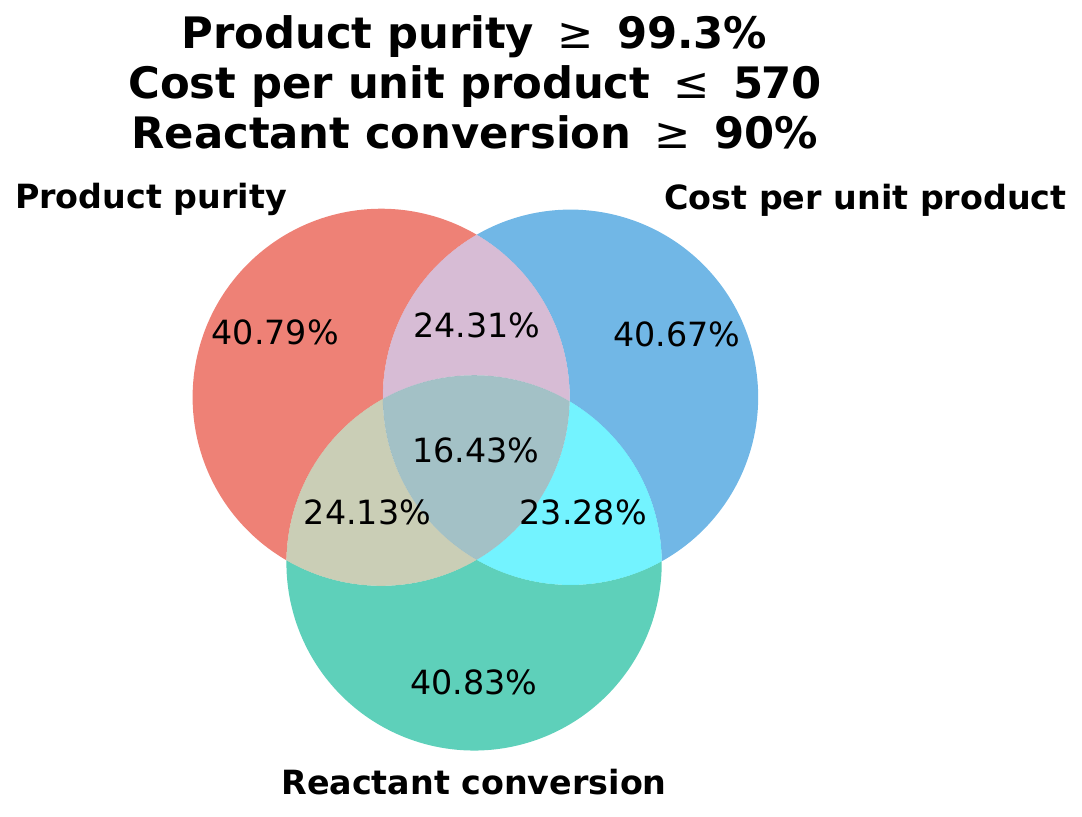}
  \caption{Venn diagram of the true pass sets for the three objectives in the kinetic model on a grid of 10{,}000 points.
  The marginal pass fractions are 40.79\% (purity), 40.67\% (cost), and 40.83\% (conversion).
  The triple intersection is 16.43\%, about 2.43 times the 6.77\% expected under independence.
  This grid illustrates the pass-set geometry.
  Campaign classification metrics are evaluated separately on the fixed 32,768-point noise-free Sobol set.}
  \label{fig:si_venn}
\end{figure}

Fig.~\ref{fig:si_venn} illustrates the geometry on a grid of $N = 10{,}000$ points, distinct from the 32,768-point Sobol set used to score the corrected campaigns.
On this grid, purity, conversion, and cost pass at 40.79\%, 40.83\%, and 40.67\% of the points, respectively.
The joint pass region covers 16.43\%, about 2.43 times the 6.77\% expected under independence ($0.4079 \times 0.4083 \times 0.4067 = 0.0677$).
This positive association is consistent with the shared dependence of the objectives on product yield $P_f$: high conversion and purity increase $P_f$, while cost per unit product decreases with it.
Using counts from this same grid,
\begin{equation}
  P(\text{cost passes} \mid \text{purity} \cap \text{conv pass}) = \frac{1643}{2413} = 68.1\%.
\end{equation}
The cost pass set also contains 2,424 points outside the joint pass region (24.24\% of the grid).

An individual cost-classification error outside the true joint region need not be a joint-classification error: if another objective correctly rejects that point, the joint classifier also rejects it.
Intersecting the predicted pass sets can therefore filter out errors on irrelevant parts of an individual boundary without improving that individual classifier.
It can also introduce false rejections if any individual classifier rejects a truly joint-feasible point.
The balance of these effects, rather than a product rule for individual Jaccard indices, determines the joint index.
The pass-set overlap is a property of the response geometry.
It does not imply that the independent GP surrogates model statistical cross-objective correlations.

\subsection{Two-Dimensional Interaction Plots}

\begin{figure}[H]
  \centering
  \includegraphics[width=0.8\textwidth]{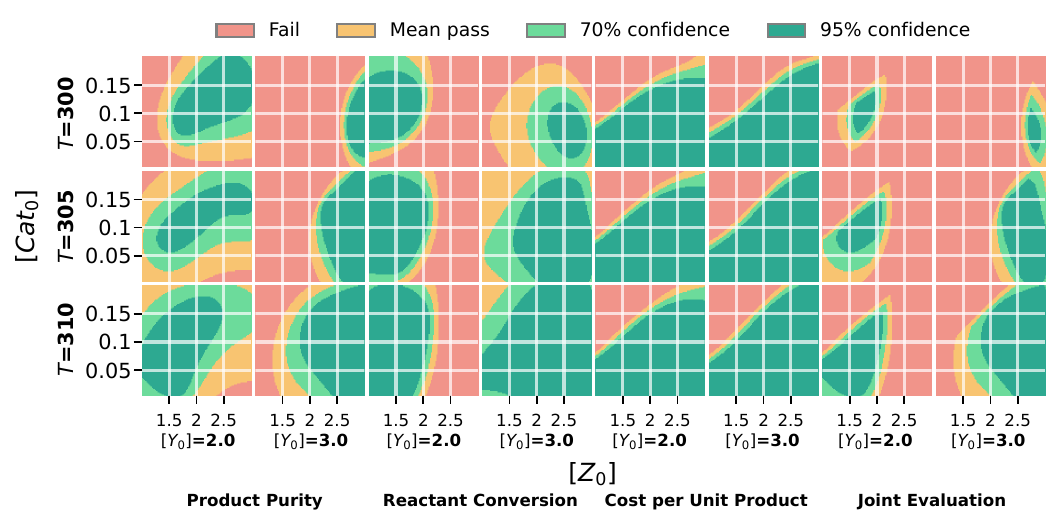}
  \caption{Two-dimensional interaction effects between the initial reactant concentration [$Z_0$] and initial catalyst concentration [$Cat_0$] obtained from JAREX with other parameters fixed at various values.
  80 samples of the campaign of main-text Figs.~6 and 7 are used to train the models.
  Each panel shows the predicted pass region at three confidence levels derived from the Gaussian process posterior: mean pass (predicted mean exceeds threshold) in yellow, 70\% confidence in light green, and 95\% confidence in dark green, with the remaining area (red) classified as fail based on the predicted mean values.
  The last column shows the joint evaluation, where confidence levels are determined by the intersection of individual objective pass regions at each confidence level.
  Compared with Fig.~7 in the main text (40 samples), the 95\% confidence regions are substantially wider, indicating improved model certainty with the larger training set.
  }
  \label{fig:interaction_80}
\end{figure}

Fig.~\ref{fig:interaction_80} shows the same two-dimensional interaction effects as Fig.~7 in the main text, but with the JAREX model of the same campaign trained on 80 samples (8 initial and 72 iterative) instead of 40.
Doubling the sample budget sharpens the boundaries and widens the high-confidence regions: the 95\% confidence pass regions (dark green) grow substantially, and the yellow band (mean pass only) that borders the fail region narrows in most panels.
The joint maps agree better with the noise-free model: their Jaccard index over the six maps rises from 0.57 to 0.81.
The largest change is at 300~K, where the 40-sample model is conservative: it predicts a joint pass region on 0\% and 12\% of the two maps, against 33\% and 22\% in the truth, and the 80-sample model on 14\% of each.
At 310~K, the predicted shares (22\% and 57\% with 40 samples, 35\% and 66\% with 80) approach the true 37\% and 67\%.
The wider high-confidence regions of the 80-sample model would provide stronger statistical support for defining proven acceptable ranges in a regulatory context.

\paragraph*{Variation across campaigns.}
Main-text Figs.~6 and 7 show one of the ten serial JAREX campaigns of the kinetic benchmark (standard transformation), chosen at the median accuracy after 40 experiments (joint Jaccard index 0.648; median 0.652, range 0.462--0.753) and among the four in which a 70\% joint range contains the anchor of the PAR analysis.
Over the ten campaigns, the joint maps of Fig.~7 agree with the noise-free model with a Jaccard index of 0.25--0.72 after 40 experiments (median 0.48; 0.57 for the campaign shown) and 0.60--0.81 after 80 (median 0.76; 0.81), and every campaign improves with the larger budget.
Every campaign is conservative at 300~K after 40 experiments, predicting a joint pass region on 0--15\% and 0--13\% of the two maps (true 33\% and 22\%).
For the PARs, the anchor is the centroid of each campaign's predicted joint pass region, and the true PARs come from the same anchor.
The same objectives set the same five joint limits in every campaign.
At the mean level, the joint range extends beyond the true one by more than 1\% of the input's range at none of the five limits in three campaigns, at one in three, and at two in three.
The upper limit of [$Cat_0$], set by cost, is the one most often extended (five campaigns).
In the remaining campaign, the anchor lies outside the campaign's own predicted joint range, because the purity surrogate's mean falls just below the threshold there.
At the unfavorable 70\% bounds, no joint range contains the anchor in six of the ten campaigns after 40 experiments: purity's threshold (0.993) lies just below its ceiling of 1, and in each of the six the 70\% bounds of the purity surrogate straddle the threshold at the anchor, while those of conversion and cost pass.
After 80 experiments, a 70\% joint range contains the anchor in nine of the ten campaigns.

\section{Additional Benchmarks and Analyses}
\label{sec:si_r1}

Unless stated otherwise, every multi-objective result in this section comes from the corrected benchmark: each observation receives an independent, reproducible measurement-noise draw, and every campaign is scored against the same fixed noise-free ground truth of 32,768 scrambled Sobol points.
Each configuration was run with 10 campaigns that share their initial designs and noise draws with the corresponding campaigns of every other configuration, so differences are paired by seed.
Confidence intervals for paired differences are percentile bootstrap intervals over seeds (10,000 resamples).
Greedy randomized straddle works through the objectives blockwise: it spends a contiguous block of iterations on each objective, in the stated order, before moving to the next (on the kinetic benchmark 40 per objective; elsewhere the budget split as evenly as possible, e.g.\ 18 + 6 $\times$ 17 for seven objectives).
This schedule represents staged objective prioritization, not all possible objective-wise sampling policies.
All objectives are measured and all surrogates updated at every experiment.
The blocks change the selection criterion, not the set of measured objectives.
The benchmarks count experiments and assign no additional cost to changing the prioritized objective.
The results, and conclusions about order sensitivity and plateaus, refer to this blockwise schedule.
We did not test an interleaved schedule.

\subsection{Ablations of the Aggregation Rule and the Optimistic Mask}
\label{sec:si_ablation}

\paragraph*{Composition of the joint boundary.}
We located 1,960 points on the true, noise-free joint boundary of the kinetic benchmark by bisection along the edges of a $10^4$-point grid that cross it, and counted at each point the objectives lying within one, two, or three local measurement-noise standard deviations of their thresholds.
Within one standard deviation, 95.7\% of the boundary points are defined by a single objective, 4.3\% by two, and none by all three.
Within three standard deviations, the shares are 87.4\%, 12.3\%, and 0.3\%.
Purity binds at 46\% of the boundary points, conversion at 31\%, and cost at 23\%.
The joint boundary is therefore predominantly single-constraint.

\paragraph*{Closed-loop ablation.}
Table~\ref{tbl:si_ablation} compares four aggregation rules (softmin, mean, max, and softmax), each with and without the mask ($k=0$ makes the mask identically 1), and two arms that keep the mask but replace the straddle aggregation with uncertainty sampling or random sampling inside the optimistic region.
The latter are simplified adaptive sampling policies that retain the learned per-objective surrogates and updated optimistic joint-feasibility region.
We use them as internal ablations to isolate the incremental contribution of boundary-directed acquisition, not as comparisons between experimental workflows.
The decision rule was stated before the runs: softmin would be retained unless an alternative's paired gain in the final joint Jaccard index and in the area under the joint-Jaccard curve both had confidence intervals above zero on both benchmarks.
Retaining the default under this rule does not establish its superiority on individual benchmarks.
No alternative meets this criterion on the kinetic benchmark.
With the mask, the four aggregation rules reach similar accuracy.
Removing the mask lowers the final joint Jaccard index of every rule by 0.11--0.22.
Random sampling inside the learned optimistic region reaches a final joint Jaccard index of 0.856, against 0.878 for JAREX, indicating that the mask carries much of the benefit on this benchmark.
The straddle aggregation adds 0.022 in the final index, but its AUC advantage is not statistically resolved (Table~\ref{tbl:si_ablation}).
It also allocates a larger share of experiments near the boundary (Table~\ref{tbl:si_selection}).
On the HPLC case study (Section~\ref{sec:si_benchmark_b}; 5 inputs, 7 objectives, dimension-scaled priors, 10 paired seeds), no alternative meets the criterion either: every rule, with and without the mask, reaches a final joint Jaccard index of 0.508--0.534, softmin the highest with and without the mask (0.534), and removing the mask lowers no rule's final index by more than 0.019.
In the simplified adaptive arms, however, random sampling inside the learned optimistic region is more accurate than JAREX ($0.564 \pm 0.027$ against $0.534$; paired difference $+0.030$, 95\% CI $+0.012$ to $+0.046$), and uncertainty sampling inside it less accurate ($0.464 \pm 0.022$).
On this case study, the diagnostics suggest that the straddle scores of the kinked resolution surrogates direct experiments to individual boundaries that the joint classification does not need.
Spreading the experiments over the optimistic region does better (Section~\ref{sec:si_benchmark_b}).
The learned joint-feasibility region is therefore the core of JAREX's accuracy, and how much boundary-directed acquisition adds on top of it depends on the response regime.

\begin{table}[H]
\centering
\caption{Aggregation $\times$ mask ablation on the kinetic benchmark (standard transformation, 120 experiments after the initial design, 10 paired seeds). Final and AUC: final joint Jaccard index and its mean over all steps, mean $\pm$ standard deviation over seeds. $\Delta$: paired difference to softmin with the mask, with 95\% bootstrap confidence interval.}
\label{tbl:si_ablation}
\small
\begin{tabular}{lcccc}
\toprule
Arm & Final & AUC & $\Delta$ final & $\Delta$ AUC \\
\midrule
Softmin + mask (JAREX) & $0.878 \pm 0.015$ & $0.703 \pm 0.039$ & -- & -- \\
Mean + mask & $0.877 \pm 0.015$ & $0.717 \pm 0.022$ & $-0.001$ [$-0.014$, $+0.014$] & $+0.015$ [$-0.010$, $+0.040$] \\
Max + mask & $0.877 \pm 0.011$ & $0.719 \pm 0.031$ & $-0.001$ [$-0.007$, $+0.007$] & $+0.016$ [$-0.007$, $+0.039$] \\
Softmax + mask & $0.874 \pm 0.012$ & $0.710 \pm 0.031$ & $-0.004$ [$-0.016$, $+0.009$] & $+0.007$ [$-0.009$, $+0.023$] \\
Softmin, no mask & $0.765 \pm 0.084$ & $0.658 \pm 0.047$ & $-0.113$ [$-0.161$, $-0.075$] & $-0.045$ [$-0.074$, $-0.017$] \\
Mean, no mask & $0.701 \pm 0.039$ & $0.607 \pm 0.021$ & $-0.177$ [$-0.202$, $-0.150$] & $-0.096$ [$-0.119$, $-0.070$] \\
Max, no mask & $0.660 \pm 0.057$ & $0.543 \pm 0.037$ & $-0.218$ [$-0.258$, $-0.184$] & $-0.160$ [$-0.196$, $-0.125$] \\
Softmax, no mask & $0.702 \pm 0.077$ & $0.574 \pm 0.071$ & $-0.175$ [$-0.224$, $-0.126$] & $-0.129$ [$-0.181$, $-0.077$] \\
Mask + uncertainty sampling & $0.803 \pm 0.036$ & $0.601 \pm 0.079$ & $-0.075$ [$-0.097$, $-0.054$] & $-0.101$ [$-0.146$, $-0.055$] \\
Mask + random sampling & $0.856 \pm 0.021$ & $0.697 \pm 0.024$ & $-0.022$ [$-0.040$, $-0.004$] & $-0.006$ [$-0.031$, $+0.023$] \\
\bottomrule
\end{tabular}
\end{table}

\paragraph*{Where the selected experiments lie.}
We evaluated the noise-free model at every experiment selected after the initial design and classified it as inside the true joint pass region, near its boundary (within one or three local noise standard deviations), or outside it by more than three standard deviations (Table~\ref{tbl:si_selection}).
Softmin with the mask places the largest share of its experiments within one noise standard deviation of the boundary (13.8\%).
Within three, the mean rule places a similar share (30.4\% against 30.2\%).
Without the mask, every rule samples almost as untargeted as space-filling sampling, and uncertainty sampling inside the mask concentrates on high-variance corners of the region rather than on the boundary.

\begin{table}[H]
\centering
\caption{Location of the experiments selected after the initial design, relative to the true joint pass region (mean over 10 campaigns; distances in local measurement-noise standard deviations).}
\label{tbl:si_selection}
\small
\begin{tabular}{lcccc}
\toprule
Method & Inside & Within 1 sd & Within 3 sd & Outside, $>3$ sd \\
\midrule
JAREX (softmin + mask) & 39.7\% & 13.8\% & 30.2\% & 44.8\% \\
Mean, max, softmax + mask & 32--37\% & 8--12\% & 23--30\% & 47--54\% \\
Any rule, no mask & 14--23\% & 1--4\% & 6--13\% & 71--83\% \\
Mask + uncertainty sampling & 28.2\% & 1.5\% & 8.7\% & 67.4\% \\
Mask + random sampling & 78.5\% & 7.1\% & 18.6\% & 14.2\% \\
Space-filling & 16.1\% & 1.3\% & 5.5\% & 81.2\% \\
Greedy randomized straddle & 17--20\% & 1--3\% & 7--8\% & 76--80\% \\
\bottomrule
\end{tabular}
\end{table}

\paragraph*{Score regime and counterfactual selection.}
To check whether the softmin suppresses single-objective boundary segments, we restored the fitted surrogates of 10 JAREX campaigns at 12 steps and evaluated the closed-form per-objective straddle scores at the 1,960 boundary points, with $\beta$ drawn from $\chi^2_2$ as in JAREX.
The binding objective's score is below $\tau$ at 75\% of the points after 10 iterations, at 94\% after 20, and at 98\% or more from iteration 30 onward (median $0.05\tau$--$0.06\tau$ late in the campaign), so JAREX operates mostly in the mean-like regime of Section~\ref{sec:si_softmin}.
There, softmin scores boundary points defined by one objective at 0.31--0.35 of the max rule's score and points defined by two objectives at 0.33--0.36, a nearly uniform damping.
On the same surrogates and a fixed pool of 4,096 candidates, the ten best candidates of softmin with the mask lie within one noise standard deviation of the true boundary in 8--11\% of cases (26--36\% within three), compared with 5--8\% (15--26\%) for the mean, 3--7\% (10--19\%) for max, 3--6\% (10--18\%) for softmax, and a base rate of 1.9\% (6.5\%).
Of softmin's near-boundary choices, 96\% lie on single-objective segments, close to the pool's 98.7\%.
Without the mask, essentially none of any rule's ten best candidates lie near the boundary after iteration 40.

\subsection{Sensitivity to the Hyperparameters \texorpdfstring{$\tau$}{tau} and \texorpdfstring{$k$}{k}}
\label{sec:si_hyper}

Table~\ref{tbl:si_hyper} varies the softmin temperature $\tau$ and the mask decay rate $k$ one at a time around the defaults ($\tau = 0.5$, $k = 2$).
Performance is flat in $\tau$ from 0.25 to 2: no paired difference in the final joint Jaccard index or its area under the curve has a confidence interval that excludes zero.
At $\tau = 0.1$, JAREX is slightly more accurate than with the default (by 0.014 in the final index and 0.037 in its area under the curve, both with confidence intervals above zero).
We keep $\tau = 0.5$, which was fixed before these runs.
For $k$, values from 1 to 5 perform equally well, whereas $k = 0.5$ and $k = 20$ lower the final index slightly (by 0.018 and 0.019, with confidence intervals below zero), as the trade-off of Section~\ref{sec:si_softmask} suggests: a slow decay leaves acquisition value outside the optimistic region, and a fast one weakens the gradient that guides the candidate search back into it.

\begin{table}[H]
\centering
\caption{Sensitivity to $\tau$ and $k$ on the kinetic benchmark (standard transformation, 120 experiments after the initial design, 10 paired seeds). $\Delta$: paired difference to the defaults, with 95\% bootstrap confidence interval.}
\label{tbl:si_hyper}
\small
\begin{tabular}{lccc}
\toprule
Setting & Final joint Jaccard & $\Delta$ final & $\Delta$ AUC \\
\midrule
Default ($\tau = 0.5$, $k = 2$) & $0.878 \pm 0.015$ & -- & -- \\
$\tau = 0.1$ & $0.891 \pm 0.010$ & $+0.014$ [$+0.003$, $+0.024$] & $+0.037$ [$+0.003$, $+0.069$] \\
$\tau = 0.25$ & $0.876 \pm 0.019$ & $-0.002$ [$-0.015$, $+0.012$] & $+0.011$ [$-0.008$, $+0.028$] \\
$\tau = 1$ & $0.878 \pm 0.009$ & $0.000$ [$-0.010$, $+0.013$] & $+0.008$ [$-0.007$, $+0.022$] \\
$\tau = 2$ & $0.880 \pm 0.014$ & $+0.002$ [$-0.006$, $+0.011$] & $+0.006$ [$-0.011$, $+0.025$] \\
$k = 0.5$ & $0.859 \pm 0.013$ & $-0.018$ [$-0.027$, $-0.009$] & $-0.012$ [$-0.025$, $0.000$] \\
$k = 1$ & $0.876 \pm 0.009$ & $-0.002$ [$-0.012$, $+0.009$] & $-0.003$ [$-0.013$, $+0.008$] \\
$k = 5$ & $0.871 \pm 0.017$ & $-0.006$ [$-0.018$, $+0.006$] & $-0.005$ [$-0.024$, $+0.015$] \\
$k = 20$ & $0.859 \pm 0.021$ & $-0.019$ [$-0.028$, $-0.009$] & $-0.011$ [$-0.026$, $+0.005$] \\
\bottomrule
\end{tabular}
\end{table}

\subsection{Threshold Choice}
\label{sec:si_thresholds}

Table~\ref{tbl:si_thresholds} repeats the kinetic benchmark with three alternative threshold sets chosen to change the size of the joint pass region and which objective binds.
For all three, 87--88\% of the joint boundary is single-constraint within three noise standard deviations.
JAREX has the highest joint Jaccard index and the lowest false-positive rate for every set.

\begin{table}[H]
\centering
\caption{Final joint Jaccard index for alternative threshold sets (standard transformation, 128 experiments, 10 paired seeds; mean $\pm$ standard deviation). Joint pass fractions on the noise-free $10^4$-point grid. Greedy: randomized straddle with objective order conversion, purity, cost.}
\label{tbl:si_thresholds}
\small
\begin{tabular}{lccccc}
\toprule
Set & Thresholds (purity, conversion, cost) & Joint pass & JAREX & Space-filling & Greedy \\
\midrule
Paper & 0.993, 0.90, 570 & 16.4\% & $0.878 \pm 0.015$ & $0.720 \pm 0.067$ & $0.655 \pm 0.095$ \\
A (loose) & 0.990, 0.85, 650 & 20.1\% & $0.884 \pm 0.015$ & $0.773 \pm 0.057$ & $0.737 \pm 0.083$ \\
B (tight) & 0.995, 0.95, 500 & 13.0\% & $0.844 \pm 0.010$ & $0.637 \pm 0.073$ & $0.697 \pm 0.090$ \\
C (cost 500) & 0.993, 0.90, 500 & 14.4\% & $0.873 \pm 0.015$ & $0.688 \pm 0.077$ & $0.640 \pm 0.079$ \\
\bottomrule
\end{tabular}
\end{table}

For the single-objective benchmarks, we set the threshold of each function so that its pass region covers 10\%, 30\%, or 50\% of the parameter space (quantiles of the noise-free function on $2^{16}$ scrambled Sobol points) and repeated the benchmarks with 5\% noise (Table~\ref{tbl:si_thresholds_so}).
The paper's thresholds pass 32\%, 11\%, and 25\% for six-hump camel, Hartmann 4D, and Hartmann 6D.
Randomized straddle has the highest mean final Jaccard index in seven of the nine settings.
Feasibility expected improvement (FEI) is ahead on six-hump camel at 10\% pass, and space-filling sampling on Hartmann 6D at 50\%, neither with a confidence interval that excludes zero.
Against space-filling sampling, the paired confidence intervals lie below zero except on six-hump camel at 10\% and 50\% pass and on Hartmann 6D at 50\%.
Against FEI, they lie below zero on Hartmann 6D at every pass fraction, on Hartmann 4D at 30\% and 50\%, and on six-hump camel at 50\%, reach zero on Hartmann 4D at 10\%, and include zero on six-hump camel at 10\% and 30\%.
Its advantage over space-filling sampling shrinks as the pass region grows, most clearly on Hartmann 6D, where the paired difference shrinks from $-0.157$ at 10\% pass to $-0.079$ at 30\%.
At 50\%, space-filling sampling is level with or slightly ahead of randomized straddle ($+0.017$, 95\% CI $-0.004$ to $+0.035$).
What governs this is how much of the domain lies close to the threshold relative to the measurement noise.
The pass fraction is only a proxy for it.
On Hartmann 6D, the share of the domain within one noise standard deviation of the threshold grows from 7\% at 10\% pass to 27\% at 30\% and 68\% at 50\%.
At 50\%, most of the domain is therefore ambiguous, and spreading experiments over it, as space-filling sampling does by design, pays off more than repeatedly selecting the single most informative point, as randomized straddle does.
On Hartmann 4D, the share instead falls from 2.3\% to 0.5\%, because the 50\% threshold lies near zero, where the relative noise vanishes, and randomized straddle stays ahead.
For six-hump camel, it lies between 23\% and 33\% at every pass fraction.
Because the noise is proportional to the response, changing the threshold also changes the noise at the boundary (for Hartmann 4D, from a standard deviation of 0.078 at 10\% pass to 0.008 at 50\%), so these variations change the pass fraction and the boundary noise together.
Inatsu et al.~\cite{Inatsu2024} report a consistent pattern on noise-free five-dimensional problems: random sampling reached a higher F-score than randomized straddle for the first 250--300 of 500 iterations on the Rosenbrock and Styblinski--Tang functions, whose superlevel sets cover 40\% and 50\% of the domain by our calculation, whereas randomized straddle was at least as accurate throughout on the sphere function (30\%).
In the multi-objective benchmark, only 1.9\% of the parameter space lies within one local noise standard deviation of the joint boundary (Section~\ref{sec:si_ablation}), the regime in which targeting pays off most.

\begin{table}[H]
\centering
\caption{Final Jaccard index on the single-objective benchmarks with thresholds set by pass fraction (5\% noise, 10 paired seeds; mean $\pm$ standard deviation). FEI: feasibility expected improvement. $\Delta$: paired difference in the final Jaccard index to randomized straddle, with 95\% bootstrap confidence interval.}
\label{tbl:si_thresholds_so}
\scriptsize
\setlength{\tabcolsep}{2.5pt}
\begin{tabular}{llccccc}
\toprule
Function & Pass & Rand.\ straddle & FEI & Space-filling & $\Delta$ FEI & $\Delta$ space-filling \\
\midrule
Six-hump camel & 0.1 & $0.525 \pm 0.236$ & $0.602 \pm 0.194$ & $0.485 \pm 0.134$ & $+0.078$ [$-0.051$, $+0.224$] & $-0.040$ [$-0.218$, $+0.165$] \\
 & 0.3 & $0.700 \pm 0.044$ & $0.681 \pm 0.044$ & $0.669 \pm 0.049$ & $-0.020$ [$-0.048$, $+0.011$] & $-0.031$ [$-0.057$, $-0.007$] \\
 & 0.5 & $0.858 \pm 0.052$ & $0.819 \pm 0.067$ & $0.818 \pm 0.059$ & $-0.039$ [$-0.080$, $-0.002$] & $-0.040$ [$-0.097$, $+0.017$] \\
Hartmann 4D & 0.1 & $0.920 \pm 0.036$ & $0.905 \pm 0.024$ & $0.798 \pm 0.028$ & $-0.015$ [$-0.031$, $0.000$] & $-0.122$ [$-0.154$, $-0.092$] \\
 & 0.3 & $0.965 \pm 0.008$ & $0.956 \pm 0.008$ & $0.911 \pm 0.009$ & $-0.008$ [$-0.015$, $-0.002$] & $-0.054$ [$-0.059$, $-0.049$] \\
 & 0.5 & $0.959 \pm 0.007$ & $0.930 \pm 0.019$ & $0.918 \pm 0.005$ & $-0.028$ [$-0.037$, $-0.020$] & $-0.041$ [$-0.045$, $-0.036$] \\
Hartmann 6D & 0.1 & $0.653 \pm 0.149$ & $0.500 \pm 0.181$ & $0.495 \pm 0.113$ & $-0.153$ [$-0.283$, $-0.040$] & $-0.157$ [$-0.280$, $-0.018$] \\
 & 0.3 & $0.741 \pm 0.026$ & $0.658 \pm 0.074$ & $0.662 \pm 0.027$ & $-0.083$ [$-0.139$, $-0.039$] & $-0.079$ [$-0.100$, $-0.056$] \\
 & 0.5 & $0.695 \pm 0.039$ & $0.638 \pm 0.126$ & $0.711 \pm 0.015$ & $-0.057$ [$-0.129$, $-0.007$] & $+0.017$ [$-0.004$, $+0.035$] \\
\bottomrule
\end{tabular}
\end{table}

\subsection{Noise Level and Batch-Correlated Noise}
\label{sec:si_noise}

Table~\ref{tbl:si_noise} repeats the kinetic benchmark at 0\%, 10\%, and 20\% relative measurement noise, in addition to the 5\% of the main text.
JAREX degrades steadily as the noise grows (from 0.891 at 0\% to 0.797 at 20\%), space-filling sampling and the greedy strategy less regularly, and JAREX remains the most accurate at every level.
To test the effect of noise shared within a batch, we also correlated the noise of the experiments evaluated together (the initial design and, for $q = 4$, each batch) with correlation $\rho$.
This changes the final joint Jaccard index of every method by less than 0.01.

\begin{table}[H]
\centering
\caption{Final joint Jaccard index at different noise levels and with batch-correlated noise (standard transformation, 128 experiments for $q=1$ and 328 for $q=4$, 10 paired seeds; mean $\pm$ standard deviation). Greedy: objective order conversion, purity, cost.}
\label{tbl:si_noise}
\small
\begin{tabular}{lccc}
\toprule
Noise & JAREX & Space-filling & Greedy \\
\midrule
0\% & $0.891 \pm 0.011$ & $0.715 \pm 0.072$ & $0.720 \pm 0.068$ \\
5\% (main text) & $0.878 \pm 0.015$ & $0.720 \pm 0.067$ & $0.655 \pm 0.095$ \\
10\% & $0.852 \pm 0.017$ & $0.711 \pm 0.059$ & $0.668 \pm 0.072$ \\
20\% & $0.797 \pm 0.028$ & $0.678 \pm 0.045$ & $0.644 \pm 0.076$ \\
\midrule
5\%, $\rho = 0.5$ & $0.882 \pm 0.010$ & $0.715 \pm 0.064$ & -- \\
5\%, $\rho = 1$ & $0.884 \pm 0.012$ & $0.711 \pm 0.067$ & -- \\
5\%, $q = 4$, $\rho = 0$ & $0.923 \pm 0.006$ & -- & -- \\
5\%, $q = 4$, $\rho = 0.5$ & $0.918 \pm 0.007$ & -- & -- \\
5\%, $q = 4$, $\rho = 1$ & $0.920 \pm 0.008$ & -- & -- \\
\bottomrule
\end{tabular}
\end{table}

\subsection{Dimensional Scaling}
\label{sec:si_scaling}

We extended the kinetic benchmark in two ways.
In the all-active variant, the reaction time, fixed at 120~min in the main text, becomes a fifth input in $[30, 240]$~min.
Every input then affects the responses: the joint pass region covers 18.5\% of the parameter space overall, but 6\% for reaction times of 30--60~min and 26\% for 180--240~min.
In the irrelevant-input variants, 2, 4, 6, or 8 inputs that do not affect the model are added ($d = 6$, 8, 10, and 12).
These test robustness to non-critical parameters, as in screening, not a higher intrinsic dimension.
Each campaign starts from 8 Latin hypercube points and runs for $30d$ iterations (120 for the original model), with JAREX, greedy randomized straddle (blockwise, in the order conversion, purity, cost, with $10d$ iterations per objective), and space-filling sampling (10 paired seeds, standard transformation, ground truth on $2^{15}$ scrambled Sobol points per variant).

Table~\ref{tbl:si_scaling} reports the joint Jaccard index at an equal budget of 128 experiments and at the end of each campaign.
With the budget scaled with $d$, JAREX reaches 0.86--0.87 with 6 to 12 inputs, close to its 0.878 on the original model, and is the most accurate method for every variant, with every paired confidence interval below zero.
The number of experiments it needs to reach a joint Jaccard index of 0.8 grows roughly linearly with the number of inputs, from 70 for $d = 4$ to 114, 164, 209, and 248 for $d = 6$, 8, 10, and 12.
At the equal budget of 128 experiments, the accuracy of JAREX falls as inputs are added, and its lead over the greedy strategy shrinks and then reverses: the greedy strategy, which has by then spent all 120 iterations on the conversion boundary, is ahead by 0.049 at $d = 10$ (95\% CI 0.002 to 0.100) and by 0.150 at $d = 12$ (95\% CI 0.081 to 0.221).
The all-active variant is the hardest: after 150 iterations, JAREX reaches $0.772 \pm 0.036$, and only 1 of 10 campaigns reaches 0.8, but it remains ahead of the other methods (by 0.04 over the greedy strategy and 0.35 over space-filling sampling).

\begin{table}[H]
\centering
\caption{Joint Jaccard index for the dimensional-scaling variants of the kinetic benchmark (standard transformation, 10 paired seeds; mean $\pm$ standard deviation). Final: after $8 + 30d$ experiments. $\Delta$: paired difference in the final index to JAREX, with 95\% bootstrap confidence interval. Last column: mean number of experiments at which the index first reaches 0.8, among the campaigns that reach it (in parentheses).}
\label{tbl:si_scaling}
\footnotesize
\setlength{\tabcolsep}{4pt}
\begin{tabular}{llcccc}
\toprule
Inputs $d$ & Method & 128 experiments & Final & $\Delta$ final & Experiments to 0.8 \\
\midrule
4 (paper) & JAREX & $0.878 \pm 0.015$ & $0.878 \pm 0.015$ & -- & 70 (10/10) \\
 & Greedy & $0.655 \pm 0.095$ & $0.655 \pm 0.095$ & $-0.222$ [$-0.281$, $-0.167$] & 104 (1/10) \\
 & Space-filling & $0.720 \pm 0.067$ & $0.720 \pm 0.067$ & $-0.158$ [$-0.199$, $-0.117$] & 114 (2/10) \\
5, reaction time & JAREX & $0.722 \pm 0.054$ & $0.772 \pm 0.036$ & -- & 139 (1/10) \\
 & Greedy & $0.709 \pm 0.040$ & $0.732 \pm 0.039$ & $-0.041$ [$-0.078$, $-0.001$] & -- \\
 & Space-filling & $0.398 \pm 0.123$ & $0.427 \pm 0.100$ & $-0.346$ [$-0.416$, $-0.285$] & -- \\
6 (2 inert) & JAREX & $0.821 \pm 0.031$ & $0.873 \pm 0.015$ & -- & 114 (10/10) \\
 & Greedy & $0.745 \pm 0.053$ & $0.766 \pm 0.069$ & $-0.107$ [$-0.148$, $-0.071$] & 160 (5/10) \\
 & Space-filling & $0.573 \pm 0.099$ & $0.654 \pm 0.070$ & $-0.219$ [$-0.264$, $-0.179$] & -- \\
8 (4 inert) & JAREX & $0.729 \pm 0.040$ & $0.863 \pm 0.021$ & -- & 164 (10/10) \\
 & Greedy & $0.717 \pm 0.059$ & $0.798 \pm 0.065$ & $-0.065$ [$-0.101$, $-0.037$] & 202 (8/10) \\
 & Space-filling & $0.552 \pm 0.077$ & $0.700 \pm 0.021$ & $-0.164$ [$-0.174$, $-0.151$] & -- \\
10 (6 inert) & JAREX & $0.657 \pm 0.061$ & $0.869 \pm 0.014$ & -- & 209 (10/10) \\
 & Greedy & $0.706 \pm 0.062$ & $0.794 \pm 0.078$ & $-0.075$ [$-0.126$, $-0.042$] & 227 (8/10) \\
 & Space-filling & $0.352 \pm 0.135$ & $0.694 \pm 0.040$ & $-0.174$ [$-0.193$, $-0.157$] & -- \\
12 (8 inert) & JAREX & $0.570 \pm 0.104$ & $0.868 \pm 0.018$ & -- & 248 (10/10) \\
 & Greedy & $0.721 \pm 0.055$ & $0.829 \pm 0.037$ & $-0.039$ [$-0.068$, $-0.015$] & 231 (9/10) \\
 & Space-filling & $0.230 \pm 0.150$ & $0.580 \pm 0.136$ & $-0.288$ [$-0.375$, $-0.210$] & -- \\
\bottomrule
\end{tabular}
\end{table}

To measure the computational cost without interference from other jobs, we timed single JAREX iterations (surrogate fitting, candidate optimization, and scoring) on one saved campaign of each variant, truncated to a fixed number of observations, within one job using two threads of a 96-core cloud node (median of three repetitions).
With 128 observations, an iteration takes 17~s for $d = 4$ and 25~s for $d = 12$ (18, 17, 20, and 20~s for $d = 5$, 6, 8, and 10).
Fitting the three GPs takes 13--17~s regardless of $d$, and candidate optimization 1--10~s.
The time grows mainly with the number of observations: 21--25~s with 188 observations ($d = 6$ to 12) and 42--48~s with 308 ($d = 10$ and 12), dominated by fitting.
Scoring against the ground truth (2--5~s) belongs to the benchmark, not to the method.
In the production runs, which shared their nodes with up to 47 other tasks, iterations took roughly twice as long.
In either case, the computation per iteration is small compared with the duration of a typical experiment.

\subsection{Pharmaceutical Case Studies}
\label{sec:si_benchmark_b}

\paragraph*{Chromatography model.}
The HPLC case study simulates a reversed-phase gradient separation of five phenothiazines (chlorpromazine, perphenazine, prochlorperazine, promethazine, and thioridazine) with acetonitrile, using the mechanistic retention model of Kamedulska et al.~\cite{Kamedulska2022} with its published posterior-mean analyte parameters, ported from the authors' code.
The model gives the retention factor of each ionization form of an analyte as a function of the organic fraction, pH, and temperature, mixes the forms through the analyte's dissociation constants, and obtains the retention time by integrating the gradient program.
For the benchmark, we added piecewise-linear gradient programs (an initial hold and up to three intermediate nodes), the flow rate, peak widths from a fixed plate number, and a continuous pH (the buffer coefficients interpolated between the measured buffers).
These additions are ours, not part of the published retention model: the resolution of the benchmark is an approximation built on the published retention model, and its operating box extends the measured conditions (linear gradients of 30, 90, and 270~min at a fixed flow of 0.5~mL/min) to gradient times of 10--60~min, variable flow, and segmented programs.
Table~\ref{tbl:si_hplc_inputs} lists the inputs.
The objectives are the resolution $R_s \ge 1.5$ of each of the five peak pairs that fail this specification on more than 5\% of the parameter space with five or more inputs, the run time (retention time of the last peak) below a limit set per instance so that about 20\% of the parameter space passes jointly (15.9, 17.6, and 17.7~min for 5, 9, and 13 inputs), and a retention factor of at least 1 for the first peak.
With two objectives, they are the first critical pair and the run time.
With four, they are the first three critical pairs and the run time.
The objectives are therefore a defined set of chromatographic specifications, not a guarantee that every peak is resolved.
On the evaluated Sobol sets, one of the five omitted pairs, prochlorperazine and promethazine, fails the resolution specification on only 3.8--3.9\% of the box, below the 5\% rule, but on 8.9--9.7\% of the conditions that pass all seven objectives (5, 9, and 13 inputs), at 3.7--3.9\% of them with $R_s < 0.5$.
Of the other four omitted pairs, only chlorpromazine and thioridazine fail, at 15 of the 32,768 points with 13 inputs (3 of them in the joint pass region), and the other three never fail on these sets.
On these sets, requiring resolution of all ten pairs would reduce the joint pass region from 20\% to 18\% of the box.
All methods are evaluated against the same predefined specifications, so the reported comparison concerns this selected-specification task.
Their relative performance under an all-pairs resolution requirement has not been evaluated.
Passing the seven objectives does not guarantee that every peak pair meets the resolution specification.
Measurement noise is a retention-time standard deviation of 0.02~min and a relative peak-width standard deviation of 5\% per analyte.
This is simulated measurement repeatability, not the accuracy of the retention model, whose errors against the measured retention times are several tenths of a minute.

\begin{table}[H]
\centering
\caption{Inputs of the HPLC case study. Inputs are nested: an instance with fewer inputs keeps the others at the nominal method.}
\label{tbl:si_hplc_inputs}
\small
\begin{tabular}{llll}
\toprule
Inputs & Parameter & Range & Nominal \\
\midrule
5 & pH & 2.5--10.5 & 6.5 \\
 & Gradient time & 10--60~min & 35~min \\
 & Temperature & 25--35~$^\circ$C & 30~$^\circ$C \\
 & Initial and final organic fraction & 0.20--0.45 and 0.60--0.80 & 0.325, 0.70 \\
9 & Flow rate & 0.3--0.7~mL/min & 0.5~mL/min \\
 & Initial hold & 0--5~min & 0 \\
 & Intermediate gradient node (time, composition) & 0.2--0.8, 0--1 (fractions) & 0.5, 0.5 \\
13 & Second and third gradient nodes & as the first & 0.5, 0.5 \\
\bottomrule
\end{tabular}
\end{table}

\paragraph*{Crystallization model.}
In the crystallization case study, aspirin crystallizes from an ethanol--water solution in a cascade of mixed-suspension, mixed-product-removal (MSMPR) crystallizers, by cooling and by the addition of water as the antisolvent to each stage.
We simulate its steady state with an adaptation of the cascade model of Liu and Benyahia,~\cite{Liu2022msmpr} using the growth and nucleation kinetics measured by Lindenberg et al.~\cite{Lindenberg2009} for this system.
At steady state, the population balance of each stage reduces to algebraic equations for the moments of the size distribution (size-independent growth, no breakage or agglomeration), which are solved stage by stage with the solute mass balance.
The feed flow, 24.97~mL/min, is back-calculated from the published steady-state data.~\cite{Liu2024msmpr}
We assume a 75/25 wt\% ethanol/water feed saturated at 40~$^\circ$C, adopting the initial-solution conditions reported for the authors' related fed-batch case.~\cite{Liu2022techno}
The vessel temperatures are inputs, which assumes perfect temperature control, because the energy balance of the published model needs heat-transfer areas that are not published.
With the stage temperatures implied by the published steady state of the three-stage cascade,~\cite{Liu2024msmpr} our implementation agrees with it (dissolved aspirin within 0.0002~g/g, mean crystal size within 2~$\mu$m, and yield 0.749 against 0.75).
This is a consistency check of the reconstruction, not an independent validation of the energy-balanced model.
Each stage has three inputs: the temperature, the antisolvent flow (0--10~g/min), and the residence time (2--20~min).
At a fixed feed, the residence time also sets the vessel's working volume.
The first stage's temperature is 25--40~$^\circ$C, and each later stage cools by a fraction (0--1) of the remaining span to 25~$^\circ$C, so temperatures never rise, because the model has no dissolution.
With 3, 6, or 9 inputs, the first one, two, or all three stages of a three-stage cascade are adjustable, and the others stay at nominal settings based on the published optimization.~\cite{Liu2022msmpr}
The 12-input instance is a four-stage cascade.
Unlike that optimization, which fixed the total antisolvent addition and residence time, the benchmark varies them, so that yield and productivity compete.
The objectives are a yield (the share of the fed aspirin that crystallizes) of at least 0.55, a number-mean crystal size at the outlet of at least 200~$\mu$m (for filtration), a productivity of at least 300~g of crystals per liter of crystallizer volume per hour, and a relative supersaturation in the first vessel of at most 1.25 (against encrustation and nucleation bursts).
With two objectives, they are the yield and the size, and with three, also the productivity.
These thresholds are round numbers set before any campaign so that every specification binds and about 20\% of the 9-input box passes jointly (19\%).
With the same thresholds, 29\% passes with 3 and 6 inputs.
The four-stage cascade crystallizes more, so its yield and productivity thresholds were reset by the same rule before any campaign (0.65 and 250~g/L/h; 18\% joint pass).
The surrogates model the logarithms of the size and the productivity.
Where no crystals leave the cascade, the size is set to 1~$\mu$m.
Measurement noise is a relative standard deviation of 2\% on the yield (the productivity follows from the measured yield), 5\% on the mean size, and 2\% on the supersaturation.
The kinetics were measured in batch experiments with up to 55 wt\% water, and the nominal settings already reach 60\%.
On the evaluated Sobol sets, at least one stage exceeds 55 wt\% water at 51\%, 51\%, 66\%, and 96\% of the four-objective joint-pass points for the 3-, 6-, 9-, and 12-input instances, respectively.
These points reach up to 71 wt\% water, within the range of the solubility data (up to 75\%) but beyond that of the kinetic experiments.
The crystallization case is therefore a literature-derived computational test of the methods, and its pass regions, particularly with four stages, are model predictions rather than operating ranges supported by data.

\paragraph*{Benchmark design and surrogate.}
The objective subsets, thresholds, and parameter ranges were fixed before any campaign.
Every campaign starts from 8 Latin hypercube points and runs 120 iterations (10 paired seeds).
Greedy randomized straddle splits the iterations evenly over the objectives and was run with the objectives in their listed order and in reverse.
The ground truth is the noise-free model on $2^{15}$ scrambled Sobol points per instance.
The surrogates use the dimension-scaled priors of Hvarfner et al.~\cite{Hvarfner2024} (lengthscales $\mathrm{LogNormal}(\sqrt{2} + \tfrac{1}{2}\log d, \sqrt{3})$, noise $\mathrm{LogNormal}(-4, 1)$, no outputscale) with a Mat\'ern-5/2 kernel and BoTorch's default fit (up to five attempts, keeping the first successful one).
This choice was made before the campaigns, for every method alike, by fitting the candidate surrogates to fixed space-filling designs of each instance and comparing their joint Jaccard index on the truth set.
The Mat\'ern-5/2 kernel was preferred over BoTorch's default radial-basis kernel for these priors because it was more accurate on every instance checked and gave smoother learning curves.
The comparison used the truth set that also scores the campaigns, so it is not an independent test set.
Using the same surrogate configuration for every method controls that choice across methods, but does not rule out effects on their relative rankings.
The reported accuracies and comparisons are conditional on this selection procedure and may be optimistic.
An independently held-out evaluation set would be needed to remove this source of selection bias.
For the crystallization case, the model check favored the dimension-scaled priors at 6, 9, and 12 inputs and the default priors at 3 inputs.
We used the dimension-scaled priors at every input count, fixed before its campaigns, so that its 12 combinations differ only in the problem.
For comparison, Fig.~\ref{fig:si_pharma_hplc_default} repeats the chromatography campaigns with the default priors.
The crystallization campaigns were run with the dimension-scaled priors only.

\begin{figure}[H]
  \centering
  \includegraphics[width=\textwidth]{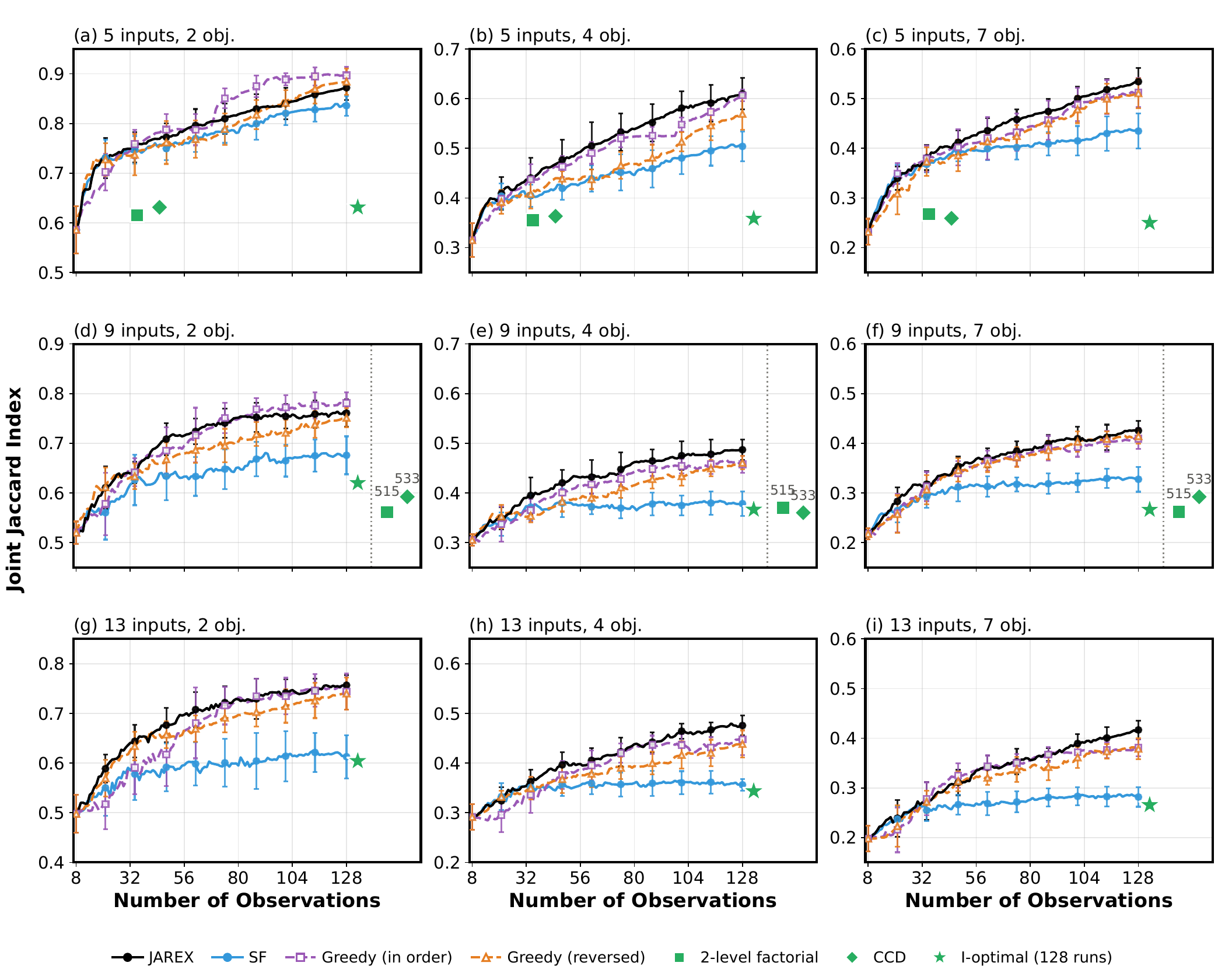}
  \caption{HPLC case study with the dimension-scaled priors: joint Jaccard index for every combination of inputs (rows) and objectives (columns); mean of 10 campaigns, error bars $\pm 1\sigma$. Every panel spans 0.45 in the joint Jaccard index, placed around its own curves, so differences have the same height in every panel.
  Read absolute values from each panel's axis. Green markers (DOE), scored with the same surrogate: two-level factorial (square) and CCD (diamond) at their run counts, and the 128-run I-optimal design (star, drawn just right of 128).
  Designs with more than 128 runs are drawn right of the dotted line and labeled with their run counts.}
  \label{fig:si_pharma_hplc}
\end{figure}

\begin{figure}[H]
  \centering
  \includegraphics[width=\textwidth]{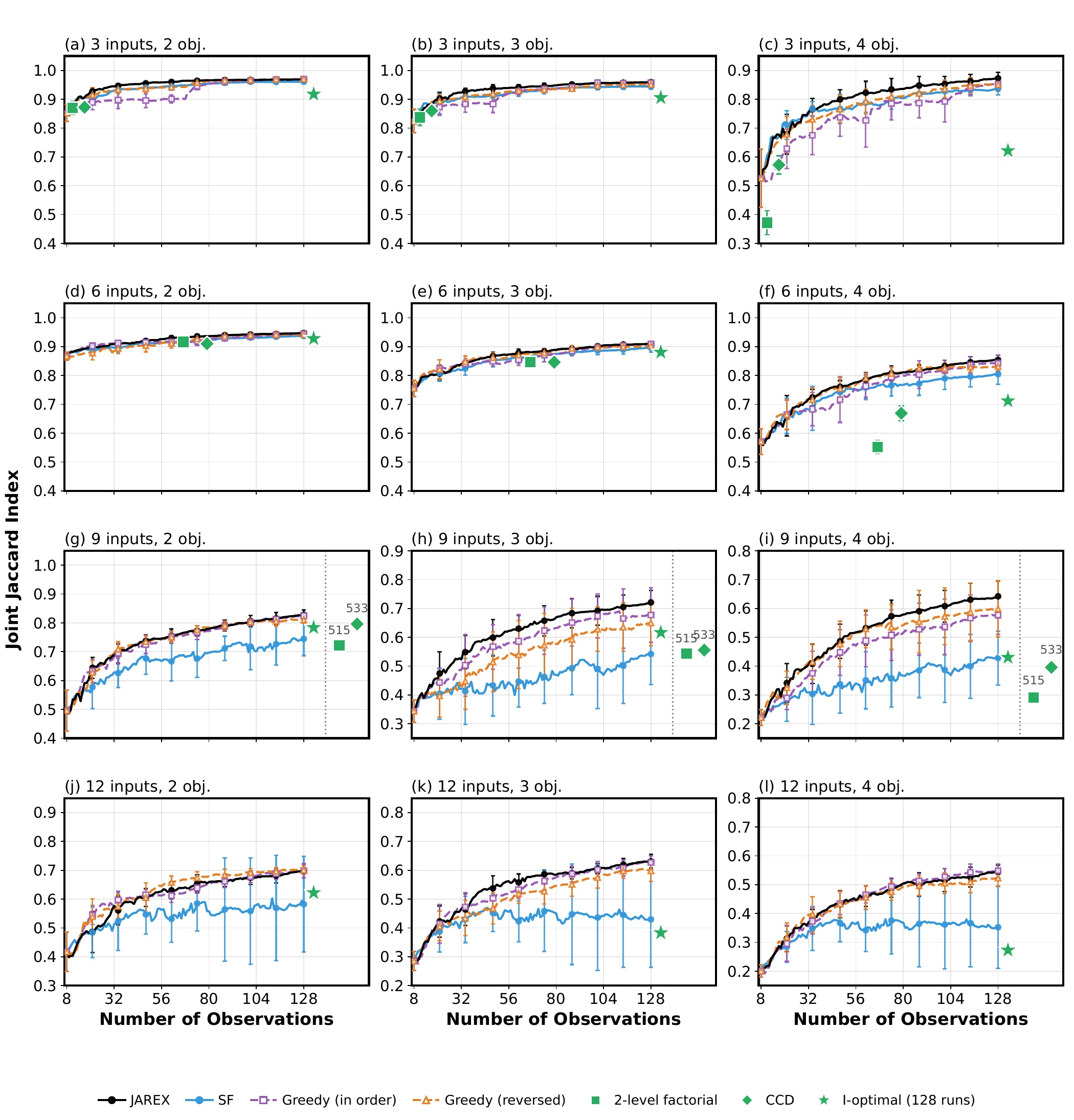}
  \caption{Crystallization case study (MSMPR cascade, the benchmark's own thresholds) with the dimension-scaled priors, as in Fig.~\ref{fig:si_pharma_hplc}. Rows: 3, 6, and 9 inputs (one, two, or three adjustable stages of a three-stage cascade) and 12 inputs (four-stage cascade, with its own yield and productivity thresholds); columns: two objectives (yield and size), three (plus productivity), and four (plus the first-vessel supersaturation). Every panel spans 0.65. The CCD was not run with 12 inputs.}
  \label{fig:si_pharma_msmpr}
\end{figure}

\begin{figure}[H]
  \centering
  \includegraphics[width=\textwidth]{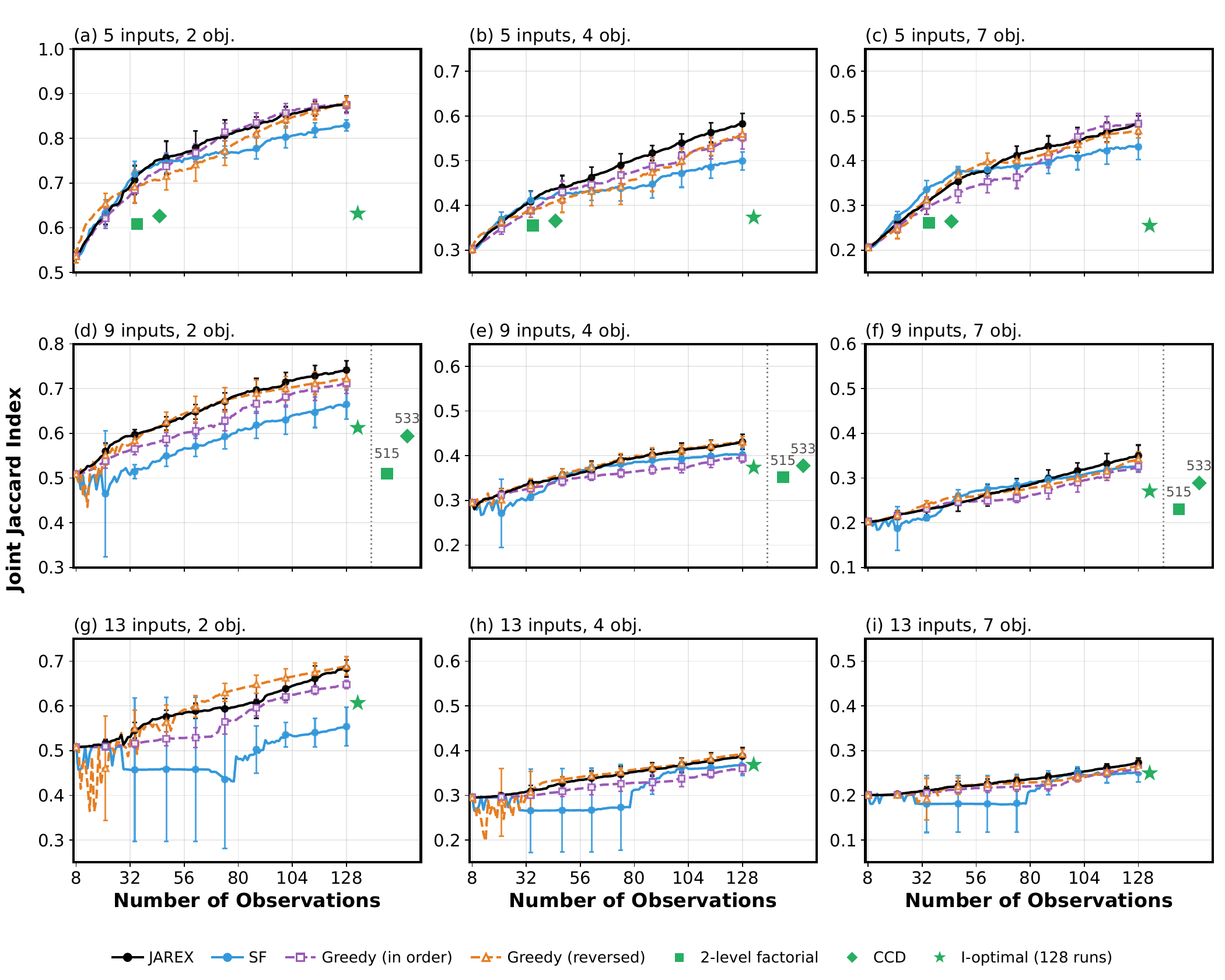}
  \caption{HPLC case study with BoTorch's default priors, as in Fig.~\ref{fig:si_pharma_hplc}.
  Every panel spans 0.50.}
  \label{fig:si_pharma_hplc_default}
\end{figure}

\paragraph*{Results.}
Table~\ref{tbl:si_pharma} and Figs.~\ref{fig:si_pharma_hplc} and~\ref{fig:si_pharma_msmpr} report every combination.
JAREX is more accurate than every design and than space-filling sampling in every combination.
At the end of the budget, JAREX is more accurate than the greedy strategy with its objective order fixed in advance (the average of the two orders run) in 12 of the 21 combinations and behind it in one (two chromatography objectives at 5 inputs, $-0.019$ [$-0.033$, $-0.006$]).
Its mean is above both orders in 15.
Against whichever order turned out better, a choice that can only be made after the runs, JAREX is ahead in four chromatography combinations (seven objectives at 5 and 13 inputs and four objectives at 9 and 13 inputs, by 0.022 to 0.036) and in three crystallization combinations (four objectives at 3 and 9 inputs, by 0.020 and 0.043, and two objectives at 6 inputs, by 0.003), and behind it in the two chromatography combinations with two objectives at 5 and 9 inputs (by 0.025 and 0.020).
In the other 12, the paired confidence intervals include zero, so these data show no statistically resolved difference there.
The largest of these unresolved differences is JAREX's lead with three crystallization objectives at 9 inputs ($+0.044$ [$-0.021$, $+0.112$]), which is resolved against the average of the two orders ($+0.057$ [$+0.012$, $+0.105$]).
On the crystallization case, the designs are below space-filling sampling at 3 and 6 inputs.
At 9 inputs, the CCD with 533 runs and the 128-run I-optimal design are above space-filling sampling's 128 runs with two and three objectives, and below JAREX in every combination.
With the default priors (Fig.~\ref{fig:si_pharma_hplc_default}; chromatography case), nearly every method is less accurate: the final joint Jaccard index changes by $-0.005$ to $+0.144$ for JAREX, by $-0.024$ to $+0.059$ for space-filling sampling, and by $+0.005$ to $+0.122$ for greedy randomized straddle when the dimension-scaled priors replace them (33 of the 36 method and combination pairs improve).
At 13 inputs, the default model stays within 0.02 of its initial accuracy for 83--99 experiments even with space-filling data, against 13--15 experiments with the dimension-scaled priors, so this failure does not depend on how the experiments are chosen.
Under the default priors, JAREX remains ahead of space-filling sampling in all nine combinations (by 0.020 to 0.130).
Against the better greedy order, it is ahead in two combinations (four objectives at 5 inputs and two at 9, by 0.022 and 0.019), and not resolved in the other seven.

\begin{table}[H]
\centering
\caption{Final joint Jaccard index on the pharmaceutical case studies with the dimension-scaled priors (128 experiments; mean of 10 seeds). Waste: share of each objective's boundary on which another objective fails, averaged over the objectives (see below). Greedy: blockwise randomized straddle, one contiguous block of iterations per objective, in their listed order (fwd) or reversed (rev). CCD: central composite design, with its run count; I-opt: the 128-run I-optimal design scored with the GP surrogate (GP) or with full quadratic models (quad; not computed for the crystallization case). $\Delta$: paired difference of JAREX to space-filling sampling and to the better greedy order (selected by its mean), with 95\% bootstrap confidence intervals.}
\label{tbl:si_pharma}
\scriptsize
\setlength{\tabcolsep}{2.2pt}
\resizebox{\textwidth}{!}{%
\begin{tabular}{lcccccccccccc}
\toprule
Case & Inputs & Obj. & Waste & JAREX & Greedy fwd & Greedy rev & SF & CCD (runs) & I-opt GP & I-opt quad & $\Delta$ SF & $\Delta$ greedy \\
\midrule
HPLC & 5 & 2 & 0.12 & 0.872 & 0.897 & 0.884 & 0.836 & 0.631 (45) & 0.631 & 0.679 & $+0.036$ [$+0.015$, $+0.055$] & $-0.025$ [$-0.043$, $-0.007$] \\
HPLC & 5 & 4 & 0.45 & 0.610 & 0.606 & 0.570 & 0.504 & 0.363 (45) & 0.358 & 0.380 & $+0.106$ [$+0.080$, $+0.131$] & $+0.003$ [$-0.019$, $+0.021$] \\
HPLC & 5 & 7 & 0.76 & 0.534 & 0.512 & 0.511 & 0.435 & 0.259 (45) & 0.250 & 0.279 & $+0.100$ [$+0.072$, $+0.131$] & $+0.022$ [$+0.009$, $+0.038$] \\
HPLC & 9 & 2 & 0.19 & 0.761 & 0.781 & 0.752 & 0.676 & 0.592 (533) & 0.620 & 0.613 & $+0.085$ [$+0.060$, $+0.111$] & $-0.020$ [$-0.039$, $-0.0005$] \\
HPLC & 9 & 4 & 0.59 & 0.487 & 0.458 & 0.461 & 0.378 & 0.360 (533) & 0.367 & 0.360 & $+0.109$ [$+0.087$, $+0.130$] & $+0.026$ [$+0.007$, $+0.042$] \\
HPLC & 9 & 7 & 0.85 & 0.425 & 0.409 & 0.414 & 0.328 & 0.292 (533) & 0.266 & 0.298 & $+0.098$ [$+0.076$, $+0.121$] & $+0.011$ [$-0.004$, $+0.027$] \\
HPLC & 13 & 2 & 0.21 & 0.757 & 0.744 & 0.740 & 0.613 & -- & 0.604 & 0.614 & $+0.144$ [$+0.111$, $+0.171$] & $+0.013$ [$-0.007$, $+0.031$] \\
HPLC & 13 & 4 & 0.62 & 0.476 & 0.448 & 0.438 & 0.356 & -- & 0.343 & 0.354 & $+0.119$ [$+0.107$, $+0.133$] & $+0.028$ [$+0.007$, $+0.049$] \\
HPLC & 13 & 7 & 0.86 & 0.417 & 0.381 & 0.379 & 0.282 & -- & 0.266 & 0.249 & $+0.135$ [$+0.116$, $+0.154$] & $+0.036$ [$+0.022$, $+0.049$] \\
\midrule
Crystallization & 3 & 2 & 0.00 & 0.969 & 0.970 & 0.970 & 0.961 & 0.872 (17) & 0.917 & -- & $+0.008$ [$+0.005$, $+0.011$] & $-0.001$ [$-0.003$, $+0.0005$] \\
Crystallization & 3 & 3 & 0.47 & 0.959 & 0.957 & 0.951 & 0.944 & 0.860 (17) & 0.905 & -- & $+0.015$ [$+0.009$, $+0.021$] & $+0.002$ [$-0.003$, $+0.008$] \\
Crystallization & 3 & 4 & 0.38 & 0.873 & 0.852 & 0.851 & 0.837 & 0.572 (17) & 0.621 & -- & $+0.036$ [$+0.014$, $+0.056$] & $+0.020$ [$+0.005$, $+0.033$] \\
Crystallization & 6 & 2 & 0.00 & 0.947 & 0.944 & 0.942 & 0.937 & 0.910 (79) & 0.928 & -- & $+0.010$ [$+0.007$, $+0.014$] & $+0.003$ [$+0.0008$, $+0.006$] \\
Crystallization & 6 & 3 & 0.42 & 0.909 & 0.909 & 0.903 & 0.896 & 0.846 (79) & 0.881 & -- & $+0.013$ [$+0.006$, $+0.022$] & $0.000$ [$-0.005$, $+0.005$] \\
Crystallization & 6 & 4 & 0.52 & 0.854 & 0.843 & 0.831 & 0.805 & 0.669 (79) & 0.712 & -- & $+0.049$ [$+0.028$, $+0.069$] & $+0.011$ [$-0.004$, $+0.027$] \\
Crystallization & 9 & 2 & 0.26 & 0.828 & 0.825 & 0.811 & 0.745 & 0.796 (533) & 0.784 & -- & $+0.084$ [$+0.053$, $+0.118$] & $+0.003$ [$-0.010$, $+0.015$] \\
Crystallization & 9 & 3 & 0.58 & 0.721 & 0.677 & 0.650 & 0.542 & 0.556 (533) & 0.616 & -- & $+0.179$ [$+0.121$, $+0.246$] & $+0.044$ [$-0.021$, $+0.112$] \\
Crystallization & 9 & 4 & 0.68 & 0.642 & 0.577 & 0.599 & 0.428 & 0.396 (533) & 0.430 & -- & $+0.214$ [$+0.162$, $+0.265$] & $+0.043$ [$+0.005$, $+0.099$] \\
Crystallization & 12 & 2 & 0.23 & 0.698 & 0.699 & 0.700 & 0.582 & -- & 0.622 & -- & $+0.116$ [$+0.048$, $+0.220$] & $-0.003$ [$-0.017$, $+0.013$] \\
Crystallization & 12 & 3 & 0.60 & 0.631 & 0.627 & 0.598 & 0.430 & -- & 0.383 & -- & $+0.201$ [$+0.112$, $+0.304$] & $+0.004$ [$-0.019$, $+0.025$] \\
Crystallization & 12 & 4 & 0.73 & 0.544 & 0.548 & 0.523 & 0.352 & -- & 0.273 & -- & $+0.191$ [$+0.112$, $+0.285$] & $-0.005$ [$-0.023$, $+0.016$] \\
\bottomrule
\end{tabular}}
\end{table}

\begin{figure}[H]
  \centering
  \includegraphics[width=\textwidth]{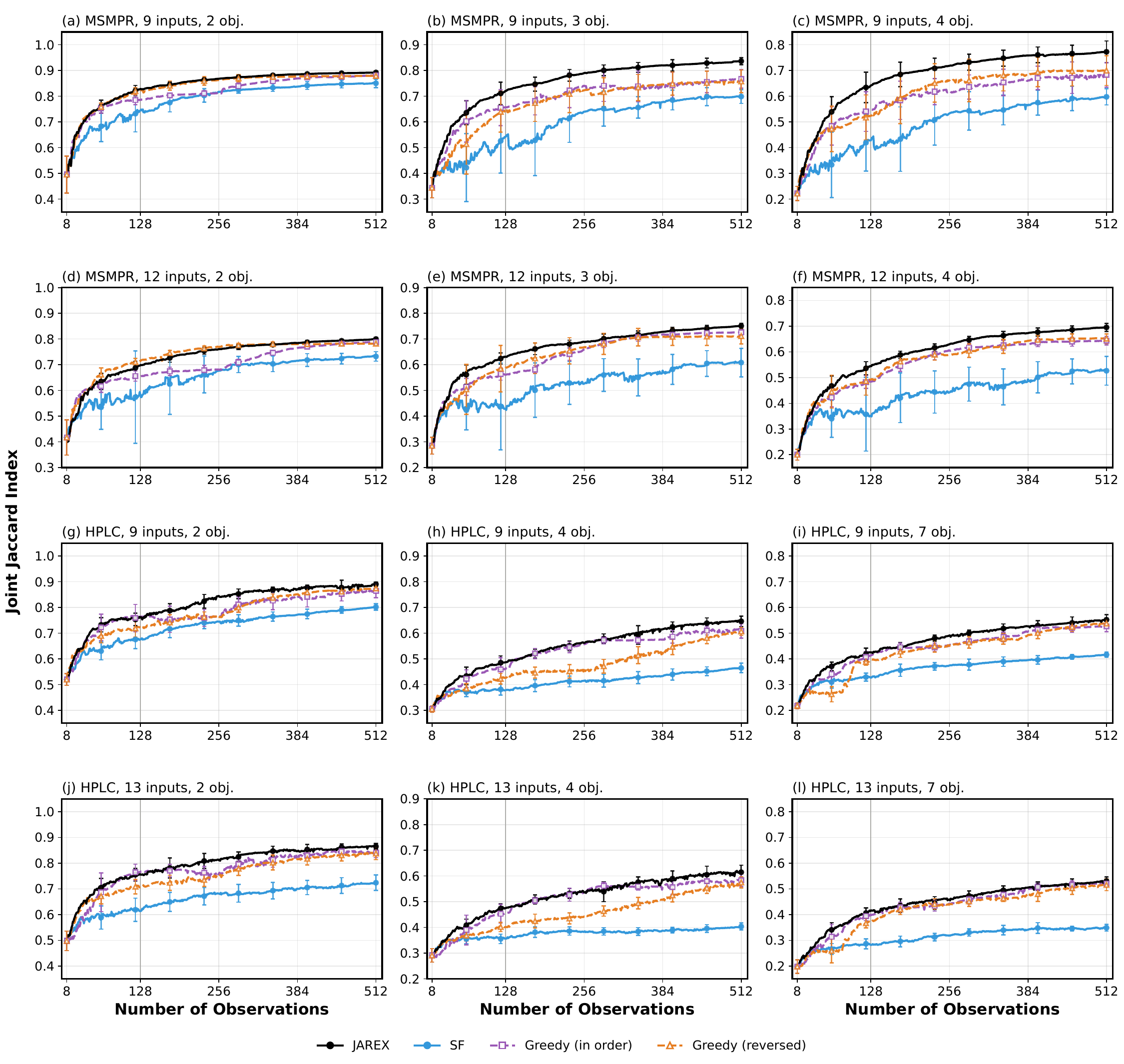}
  \caption{Joint Jaccard index on the 12 case-study combinations with 9 or more inputs, to 512 experiments (dimension-scaled priors; mean of 10 campaigns, error bars $\pm 1\sigma$). JAREX and space-filling sampling continue their campaigns to 128 experiments.
  The greedy strategy was run to 512 (and separately to 256; Table~\ref{tbl:si_pharma_long}). The gray line marks 128 experiments, the budget of the other figures. Every panel spans 0.70 in the joint Jaccard index.}
  \label{fig:si_pharma_long}
\end{figure}

\paragraph*{Longer budgets.}
The 12 combinations with 9 or more inputs were also run to 512 experiments, with the same seeds and settings (Fig.~\ref{fig:si_pharma_long} and Table~\ref{tbl:si_pharma_long}).
JAREX and space-filling sampling do not depend on the total budget, so their campaigns to 512 continue those to 128, which they reproduce exactly.
The greedy strategy splits its budget into equal blocks per objective, so it was run separately to 256 and to 512.
Every method keeps improving beyond 128 experiments.
JAREX is ahead of the greedy strategy with its order fixed in advance (the average of the two orders) in 7 of the 12 combinations at 128 experiments, 10 at 256, and all 12 at 512.
Against the better order, chosen after the runs, it is ahead in 4, 6, and 11 (behind in one at 128, none after), by 0.010 to 0.073 at 512.
Its lead widens most where the two strategies were level at 128: with 12 crystallization inputs, JAREX's difference to the better order is $-0.003$, $+0.004$, and $-0.005$ at 128 experiments and $+0.011$, $+0.025$, and $+0.043$ at 512 with two, three, and four objectives, and with two chromatography objectives at 9 inputs it changes from $-0.020$ to $+0.016$.
With seven chromatography objectives at 13 inputs, JAREX's lead of 0.036 at 128 experiments is not resolved at 256 ($+0.007$ [$-0.015$, $+0.029$]) and is $+0.010$ [$+0.002$, $+0.021$] at 512.
With seven objectives at 9 inputs, its lead over the better order at 512 ($+0.015$) is the only one not resolved.
JAREX remains ahead of space-filling sampling at every budget, by 0.042 to 0.211 at 512.

\begin{table}[H]
\centering
\caption{JAREX's paired difference in the final joint Jaccard index to the greedy strategy at 128, 256, and 512 experiments: to the average of its two objective orders (``avg''; the order fixed in advance) and to the better order by mean (``better''; chosen after the runs), with 95\% bootstrap confidence intervals over 10 seeds. At each budget, the greedy strategy was run to that budget.}
\label{tbl:si_pharma_long}
\scriptsize
\setlength{\tabcolsep}{2.0pt}
\resizebox{\textwidth}{!}{%
\begin{tabular}{lcccccccc}
\toprule
 & & & \multicolumn{2}{c}{128 experiments} & \multicolumn{2}{c}{256 experiments} & \multicolumn{2}{c}{512 experiments} \\
\cmidrule(lr){4-5}\cmidrule(lr){6-7}\cmidrule(lr){8-9}
Case & Inputs & Obj. & avg & better & avg & better & avg & better \\
\midrule
Crystallization & 9 & 2 & $+0.010$ [$-0.001$, $+0.020$] & $+0.003$ [$-0.010$, $+0.015$] & $+0.012$ [$+0.006$, $+0.019$] & $+0.008$ [$+0.003$, $+0.013$] & $+0.012$ [$+0.009$, $+0.015$] & $+0.011$ [$+0.008$, $+0.015$] \\
Crystallization & 9 & 3 & $+0.057$ [$+0.012$, $+0.105$] & $+0.044$ [$-0.020$, $+0.111$] & $+0.067$ [$+0.039$, $+0.092$] & $+0.061$ [$+0.017$, $+0.097$] & $+0.074$ [$+0.057$, $+0.090$] & $+0.069$ [$+0.048$, $+0.091$] \\
Crystallization & 9 & 4 & $+0.054$ [$+0.022$, $+0.102$] & $+0.043$ [$+0.005$, $+0.100$] & $+0.081$ [$+0.046$, $+0.121$] & $+0.078$ [$+0.040$, $+0.118$] & $+0.081$ [$+0.052$, $+0.108$] & $+0.073$ [$+0.041$, $+0.109$] \\
Crystallization & 12 & 2 & $-0.002$ [$-0.017$, $+0.015$] & $-0.003$ [$-0.017$, $+0.013$] & $+0.013$ [$+0.008$, $+0.019$] & $+0.006$ [$0.000$, $+0.013$] & $+0.014$ [$+0.009$, $+0.020$] & $+0.011$ [$+0.006$, $+0.017$] \\
Crystallization & 12 & 3 & $+0.018$ [$+0.005$, $+0.031$] & $+0.004$ [$-0.019$, $+0.025$] & $+0.020$ [$-0.001$, $+0.049$] & $+0.005$ [$-0.003$, $+0.014$] & $+0.032$ [$+0.022$, $+0.045$] & $+0.025$ [$+0.017$, $+0.034$] \\
Crystallization & 12 & 4 & $+0.008$ [$-0.012$, $+0.025$] & $-0.005$ [$-0.024$, $+0.016$] & $+0.021$ [$+0.012$, $+0.029$] & $+0.021$ [$+0.011$, $+0.030$] & $+0.048$ [$+0.035$, $+0.062$] & $+0.043$ [$+0.030$, $+0.060$] \\
HPLC & 9 & 2 & $-0.005$ [$-0.024$, $+0.012$] & $-0.020$ [$-0.040$, $-0.001$] & $+0.028$ [$+0.008$, $+0.047$] & $+0.020$ [$-0.009$, $+0.045$] & $+0.021$ [$+0.011$, $+0.033$] & $+0.016$ [$+0.010$, $+0.022$] \\
HPLC & 9 & 4 & $+0.028$ [$+0.013$, $+0.041$] & $+0.026$ [$+0.008$, $+0.042$] & $+0.038$ [$+0.022$, $+0.051$] & $+0.026$ [$+0.005$, $+0.045$] & $+0.039$ [$+0.027$, $+0.051$] & $+0.038$ [$+0.023$, $+0.052$] \\
HPLC & 9 & 7 & $+0.014$ [$-0.003$, $+0.030$] & $+0.011$ [$-0.004$, $+0.027$] & $+0.026$ [$+0.018$, $+0.034$] & $+0.023$ [$+0.014$, $+0.033$] & $+0.020$ [$+0.009$, $+0.032$] & $+0.015$ [$-0.004$, $+0.031$] \\
HPLC & 13 & 2 & $+0.015$ [$+0.003$, $+0.026$] & $+0.013$ [$-0.007$, $+0.030$] & $+0.012$ [$+0.001$, $+0.024$] & $+0.006$ [$-0.011$, $+0.023$] & $+0.026$ [$+0.017$, $+0.034$] & $+0.025$ [$+0.012$, $+0.036$] \\
HPLC & 13 & 4 & $+0.033$ [$+0.014$, $+0.051$] & $+0.028$ [$+0.008$, $+0.048$] & $+0.031$ [$+0.006$, $+0.054$] & $+0.021$ [$-0.003$, $+0.042$] & $+0.035$ [$+0.017$, $+0.055$] & $+0.031$ [$+0.011$, $+0.053$] \\
HPLC & 13 & 7 & $+0.037$ [$+0.022$, $+0.052$] & $+0.036$ [$+0.022$, $+0.049$] & $+0.011$ [$-0.008$, $+0.029$] & $+0.007$ [$-0.015$, $+0.029$] & $+0.012$ [$+0.005$, $+0.021$] & $+0.010$ [$+0.002$, $+0.021$] \\
\bottomrule
\end{tabular}}
\end{table}

\begin{figure}[H]
  \centering
  \includegraphics[width=\textwidth]{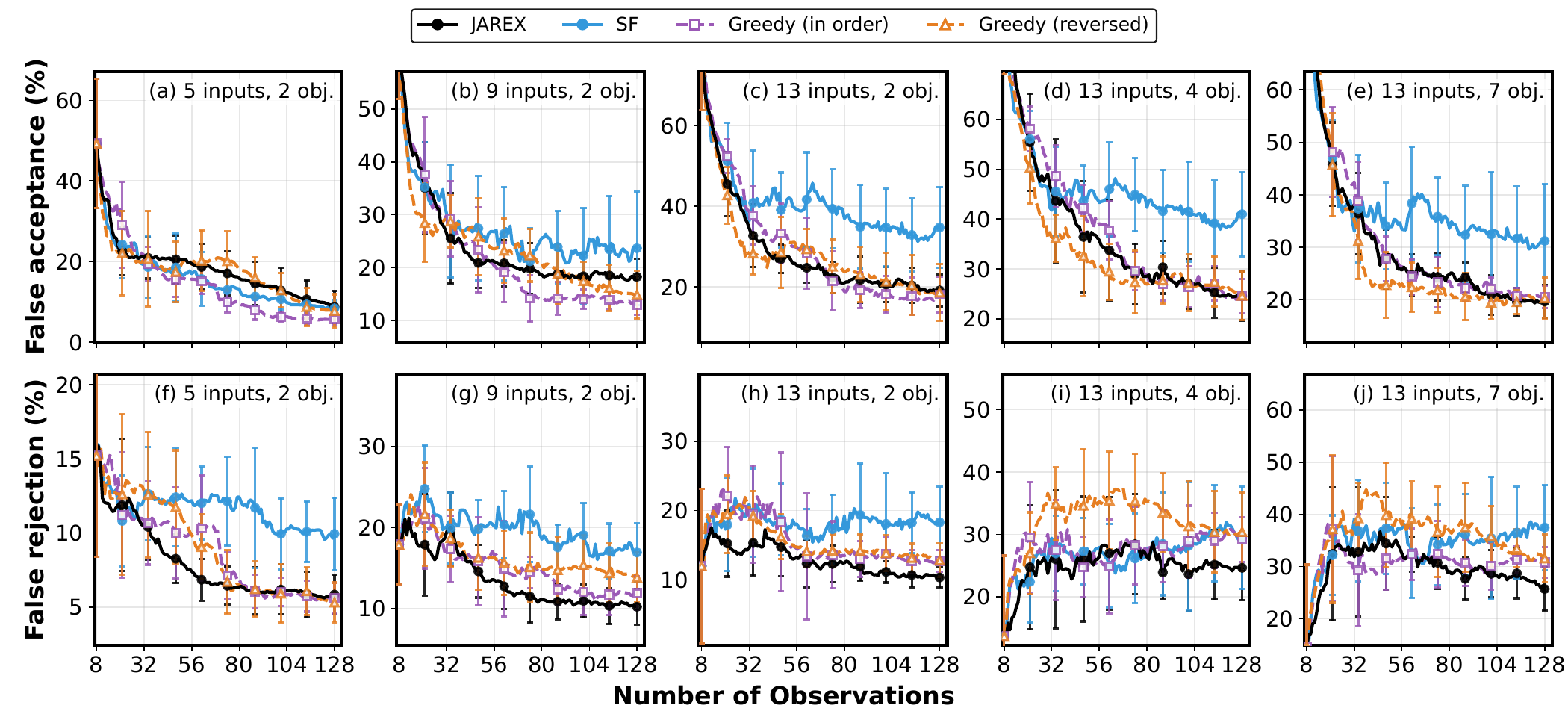}
  \caption{False acceptance, FP/(FP + TN), and false rejection, FN/(FN + TP), of the joint pass region over the campaign on the HPLC case study (dimension-scaled priors; mean of 10 campaigns, error bars $\pm 1\sigma$; each panel has its own vertical range).}
  \label{fig:si_pharma_errors_hplc}
\end{figure}

\paragraph*{False acceptance and false rejection during the campaign.}
Fig.~\ref{fig:si_pharma_errors_hplc} decomposes the errors of the joint classification into false acceptance (the share of failing conditions predicted to pass) and false rejection (the share of passing conditions predicted to fail), from the confusion matrix recorded at every step, for the chromatography case.
On the HPLC case study, JAREX and the greedy strategy accept failing conditions at similar rates (within 2.5 percentage points at 128 experiments, except with two objectives, where JAREX accepts more: for example, 18.3\% against 13.0--14.9\% with 9 inputs), which with 9 and 13 inputs are a sixth to a half lower than those of space-filling sampling (for example, 19.1\% against 34.7\% at 128 experiments with 13 inputs and two objectives).
With 9 and 13 inputs, JAREX also rejects 2 to 7 percentage points fewer passing conditions than the greedy strategy.

\begin{figure}[H]
  \centering
  \includegraphics[width=\textwidth]{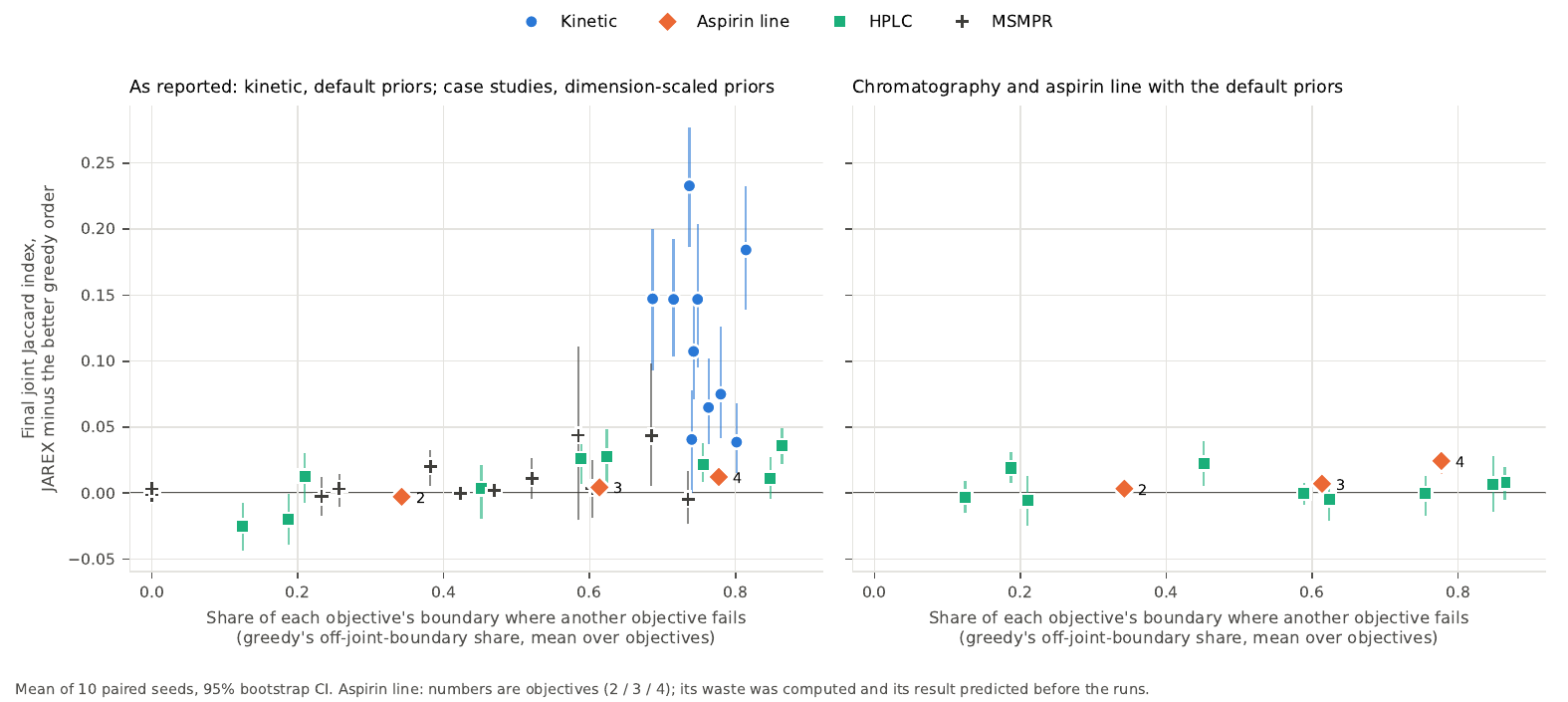}
  \caption{JAREX's paired gain in the final joint Jaccard index over the better greedy order against the waste, the share of each objective's boundary on which another objective fails, for every benchmark setting with both methods (mean of 10 paired seeds, 95\% bootstrap confidence interval); left, each benchmark in its reported setting (kinetic model, default priors; case studies and aspirin line, dimension-scaled priors); right, chromatography and aspirin line with the default priors. Kinetic: the main benchmark, the threshold sets, four objectives, and the dimensional-scaling variants. MSMPR: the 12 crystallization combinations (dimension-scaled priors only). Aspirin line: numbers give the objective count of the test described below.}
  \label{fig:si_pharma_waste}
\end{figure}

\paragraph*{Where the advantage over greedy refinement comes from.}
For objective $i$, let $r_i$ be the share of its boundary on which every other objective passes, measured on the noise-free truth set as the share of the edges between nearest-neighbor points (ten neighbors in the unit cube) that cross objective $i$'s threshold and have both ends inside the other objectives' pass regions.
We define the geometric boundary-irrelevance proxy $1 - \frac{1}{m}\sum_i r_i$ as the waste, averaging equally over objectives to match the greedy strategy's equal budget blocks.
It is not a measured fraction of the experimental budget: the cost of learning a boundary segment also depends on its difficulty, the acquisition's allocation within each block, and surrogate learning from experiments selected for other objectives.
A version based on the 3\% of points nearest each threshold gives lower values for the HPLC settings (0.06--0.59) but the same extremes (kinetic model 0.67--0.83).
Fig.~\ref{fig:si_pharma_waste} plots JAREX's paired gain over the better greedy order against the waste for every benchmark with both methods.
On the kinetic model and its variants, the waste is 0.69--0.81, and JAREX is ahead in every setting (by 0.039 to 0.233 with the default priors).
With low waste (at most 0.21: two HPLC objectives, and the crystallization settings with two objectives at 3 and 6 inputs), the two strategies tie, or the greedy strategy is slightly ahead (by 0.025 and 0.020 with two HPLC objectives at 5 and 9 inputs).
On the HPLC case study, JAREX's gain at a waste of 0.45--0.86 reaches 0.036 and is resolved in four of these six combinations (not with four objectives at 5 inputs and seven at 9).
On the HPLC case study, the mask does not act: in the closed-loop ablation of Section~\ref{sec:si_ablation} repeated on 5 inputs with 7 objectives, every aggregation rule with and without the mask reaches a final joint Jaccard index between 0.508 and 0.534 (softmin with minus without the mask: $0.000$, 95\% CI $-0.011$ to $+0.010$).
The resolution of a peak pair is the absolute value of a smooth function and has a kink where the two peaks change their elution order, which a GP represents poorly, so the surrogates stay uncertain where other objectives fail and the optimistic mask cannot exclude those regions.
A refit test supports this explanation: fitting the surrogates to the same experiments with the signed resolution, which is smooth through the crossover (pass if its absolute value is at least 1.5), raised the final joint Jaccard index of every method and cut the false-acceptance rate about threefold (JAREX on 5 inputs and seven objectives: 0.48 to 0.60, false acceptance 17\% to 6\%).
This test used earlier campaigns of this case study, run with the default priors and before the averaging-then-clamping estimator of the straddle score (Section~\ref{sec:si_softmask}), and refits the surrogates on fixed data.
A closed-loop campaign with signed-resolution surrogates would need a two-sided threshold, which JAREX does not yet support.
A second factor is that about 20\% of the box passes jointly and much of the rest is ambiguous early in a campaign, so the optimistic region covers most of the box and the mask removes few candidates even where the surrogates are accurate.
On the crystallization case, JAREX is ahead by about 0.04 with three and four objectives at 9 inputs (waste 0.58 and 0.68; resolved with four), but with 12 inputs, at as much waste (0.60 and 0.73), the two strategies tie ($+0.004$ and $-0.005$).
At 3 and 6 inputs, where every adaptive method approaches the ceiling, the differences are at most 0.020.

\begin{figure}[H]
  \centering
  \includegraphics[width=\textwidth]{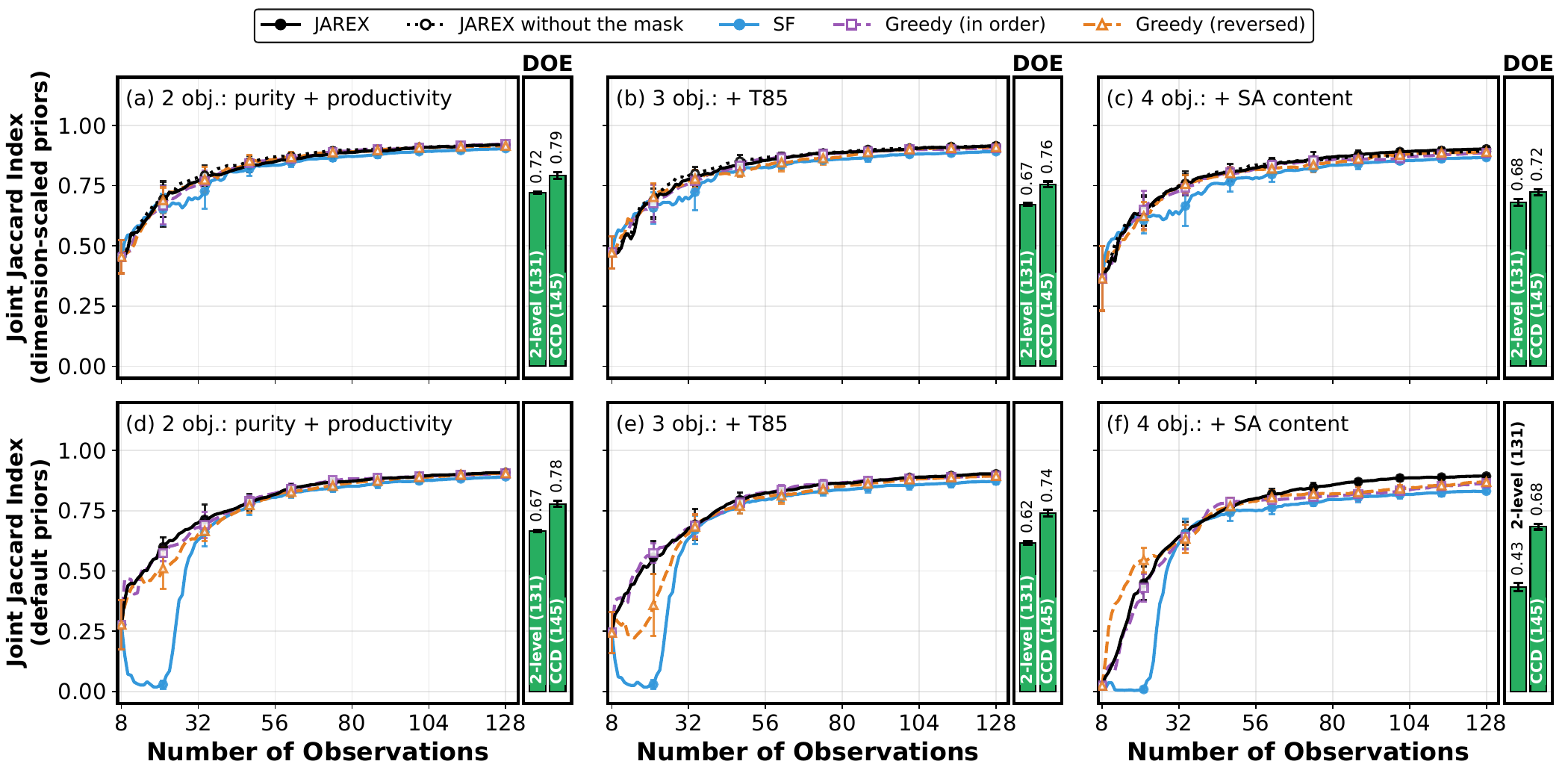}
  \caption{Joint Jaccard index on the integrated aspirin process with two, three, and four objectives; top, dimension-scaled priors; bottom, default priors (mean of 10 campaigns, error bars $\pm 1\sigma$). Side panels (DOE): two-level factorial and CCD.}
  \label{fig:si_aspirin}
\end{figure}

\paragraph*{A test of this explanation on a further model.}
Before running any campaign, we used the waste to predict JAREX's gain over the greedy strategy on a model we had not yet benchmarked: the steady state of the integrated continuous aspirin process of Nagy et al.,~\cite{Nagy2021} which links flow acetylation of salicylic acid, quenching with ethanol, antisolvent crystallization in a mixed-suspension, mixed-product-removal crystallizer, and capsule dissolution.
Our implementation reproduces the paper's simulated optima (conversion and purity within 0.7 percentage points, productivity within 0.5~g/h, mean crystal size within 0.4~$\mu$m, and $T_{85}$ within 0.5~min).
Its seven inputs are the acetic anhydride equivalents (2--8), the acetylation temperature (45--70~$^\circ$C) and residence time (60--180~min), the quench temperature (60--100~$^\circ$C), the ethanol equivalents (3--6), the crystallizer temperature (2.5--35~$^\circ$C, extended by 10~$^\circ$C beyond the measured range so that the dissolution limit binds), and the antisolvent ratio (3--5).
The objectives are the ASA purity of the quenched mixture ($\ge 0.75$), the productivity ($\ge 6$~g/h), the capsule $T_{85} \le 15$~min, the criterion used by Nagy et al., and the salicylic-acid content ($\le 0.15$).
The purity, productivity, and salicylic-acid thresholds were set by the rule used for the other benchmarks (every specification binds, about 20\% joint pass), because the product specifications do not apply to the crude reaction mixture.
Posed with 2 (purity and productivity), 3 (plus $T_{85}$), or 4 objectives, the waste is 0.34, 0.61, and 0.78, and the outputs are smooth.
We predicted that JAREX's gain would be positive and largest with four objectives and absent with two.
The prediction held with both priors (Table~\ref{tbl:si_aspirin} and Fig.~\ref{fig:si_aspirin}), with gains limited by a ceiling that every adaptive method approached: with the dimension-scaled and the default priors, JAREX's difference to the better greedy order is not resolved with two objectives ($-0.003$ and $+0.003$), small with three ($+0.004$, whose confidence interval barely excludes zero, and $+0.007$), and largest with four ($+0.012$ and $+0.024$).
JAREX without the mask was as accurate as JAREX with two and three objectives and slightly less accurate with four (by 0.010), so the mask contributed little here.
Every adaptive method reached a joint Jaccard index near 0.9, which leaves little room for any difference.

\begin{table}[H]
\centering
\caption{Final joint Jaccard index on the integrated aspirin process (7 inputs, 128 experiments, 10 seeds). $\Delta$: paired difference of JAREX to the better greedy order and to JAREX without the mask (dimension-scaled priors only), with 95\% bootstrap confidence intervals. CCD with its run count.}
\label{tbl:si_aspirin}
\scriptsize
\setlength{\tabcolsep}{2.6pt}
\begin{tabular}{lcccccccccc}
\toprule
Priors & Obj. & Waste & JAREX & No mask & Greedy fwd & Greedy rev & SF & CCD (runs) & $\Delta$ greedy & $\Delta$ no mask \\
\midrule
Dimension-scaled & 2 & 0.34 & 0.920 & 0.920 & 0.923 & 0.915 & 0.904 & 0.792 (145) & $-0.003$ [$-0.009$, $+0.003$] & $0.000$ [$-0.003$, $+0.004$] \\
Dimension-scaled & 3 & 0.61 & 0.914 & 0.915 & 0.905 & 0.910 & 0.892 & 0.756 (145) & $+0.004$ [$+0.0001$, $+0.009$] & $-0.001$ [$-0.006$, $+0.004$] \\
Dimension-scaled & 4 & 0.78 & 0.902 & 0.892 & 0.885 & 0.890 & 0.867 & 0.723 (145) & $+0.012$ [$+0.003$, $+0.020$] & $+0.010$ [$+0.001$, $+0.018$] \\
Default & 2 & 0.34 & 0.909 & -- & 0.905 & 0.906 & 0.892 & 0.779 (145) & $+0.003$ [$-0.004$, $+0.011$] & -- \\
Default & 3 & 0.61 & 0.903 & -- & 0.896 & 0.895 & 0.874 & 0.740 (145) & $+0.007$ [$-0.001$, $+0.015$] & -- \\
Default & 4 & 0.78 & 0.894 & -- & 0.865 & 0.869 & 0.832 & 0.684 (145) & $+0.024$ [$+0.015$, $+0.034$] & -- \\
\bottomrule
\end{tabular}
\end{table}

\paragraph*{Two designed tests on the crystallization case.}
On the aspirin process, every boundary is easy for the surrogates (per-objective Jaccard index of space-filling sampling 0.85--0.98), which suggests that wasted boundary segments cost the greedy strategy experiments only where they are hard to learn.
We therefore combined, for each objective $i$, its own waste $1 - r_i$ with its difficulty $1 - J_i$, where $J_i$ is the per-objective Jaccard index of space-filling sampling after 128 experiments, and used the largest product $(1 - r_i)(1 - J_i)$ over the objectives as an index.
This index was defined after the results above, on which its Spearman correlation with JAREX's gain over the better greedy order is 0.84 over the 13 settings of the kinetic, chromatography, and aspirin cases (0.80 without the kinetic model), so these results are not evidence for it.
We then stated predictions before running two sets of campaigns.
First, with 9 inputs and four objectives, we chose by a screen of the noise-free model three threshold sets: the one with the highest index (0.28, with the hardest crystal-size boundary), one with nearly the same waste and joint pass but an index of 0.13, and the benchmark's own thresholds (0.15).
These values use the space-filling campaigns with the default priors.
With the dimension-scaled priors, the index for the benchmark's own thresholds is 0.18.
We predicted that JAREX would be ahead under the highest-index set, by about 0.03 to 0.09, and more than under the other two.
JAREX was ahead under all three sets, by $+0.063$ [$+0.036$, $+0.091$], $+0.043$ [$+0.020$, $+0.065$], and $+0.043$ (Table~\ref{tbl:si_pharma}).
The order between the sets is as predicted, but 10 campaigns per set do not resolve it (highest minus lowest index: $+0.020$ [$-0.015$, $+0.057$]).
Second, across the other 11 combinations of Table~\ref{tbl:si_pharma}, we predicted ties where the index is below 0.05, gains where it exceeds 0.15 (three objectives at 9 inputs, and three and four at 12), with a confidence interval above zero for the highest index (0.21, four objectives at 12 inputs), and a Spearman correlation above 0.5 between the index and the gain.
This test failed: with 12 inputs, the two strategies tie (the gain with four objectives is $-0.005$ [$-0.023$, $+0.016$]), the correlation is $-0.02$, and one low-index combination, four objectives at 3 inputs, shows a small resolved gain ($+0.020$).
The index therefore describes the gains at 9 inputs but does not predict them across this model.
One explanation, found after the results and not tested, is that with 12 inputs, 128 experiments are too few for the surrogates to recognize where the other objectives fail, which the allocation argument requires for a gain.

\subsection{Four Objectives}
\label{sec:si_four_objectives}

Adding a limit on the residual co-reactant concentration, $[\mathrm{Y}]_\mathrm{final} \le 1.50$~M, shrinks the joint pass region from 16.4\% to 10.4\% of the parameter space.
The new objective binds at 24\% of the joint boundary (purity 21\%, conversion 38\%, cost 17\%), and 82\% of the boundary remains single-constraint within three noise standard deviations.
After 160 experiments following the initial design, JAREX reaches a final joint Jaccard index of $0.901 \pm 0.009$, against $0.736 \pm 0.044$ for space-filling sampling and $0.717 \pm 0.078$ for greedy randomized straddle (objective order conversion, purity, cost, $[\mathrm{Y}]_\mathrm{final}$), with joint false-positive rates of 0.60\%, 1.38\%, and 1.53\%.

\subsection{Independent versus Multi-Task Gaussian Process Surrogates}
\label{sec:si_mtgp}

We took the campaigns of JAREX and of space-filling sampling (10 seeds each) at four stages and fitted, to exactly the data the campaign's independent surrogates had seen, multi-task GPs (intrinsic coregionalization model, BoTorch \texttt{KroneckerMultiTaskGP}) with task ranks 1 and 3 and noise-covariance ranks 0, 1, and 3.
All models use a Mat\'ern-5/2 kernel with automatic relevance determination on the same transformed outputs.
The multi-task model shares one input kernel across objectives.
Among the multi-task variants, the one with the highest marginal likelihood was selected, without reference to the ground truth (Table~\ref{tbl:si_mtgp}).
We compare it with the campaign's own independent surrogates.
Independent GPs refitted with the same optimizer as the multi-task models agree with them to within 0.003 in the joint Jaccard index in 78 of the 80 fits and converged to a poor optimum in the other two.

\begin{table}[H]
\centering
\caption{Joint Jaccard index against the noise-free truth of the campaign's independent GPs and of the selected multi-task GP fitted to the same data (10 seeds; $\Delta$: paired difference, multi-task minus independent, with 95\% bootstrap confidence interval).}
\label{tbl:si_mtgp}
\small
\begin{tabular}{llccc}
\toprule
Campaigns & Observations & Independent & Multi-task & $\Delta$ \\
\midrule
JAREX & 24 & 0.458 & 0.456 & $-0.003$ [$-0.043$, $+0.050$] \\
 & 48 & 0.707 & 0.707 & $0.000$ [$-0.016$, $+0.019$] \\
 & 88 & 0.837 & 0.832 & $-0.006$ [$-0.015$, $+0.002$] \\
 & 128 & 0.876 & 0.867 & $-0.010$ [$-0.019$, $-0.002$] \\
Space-filling & 24 & 0.450 & 0.420 & $-0.030$ [$-0.131$, $+0.055$] \\
 & 48 & 0.488 & 0.478 & $-0.009$ [$-0.063$, $+0.055$] \\
 & 88 & 0.620 & 0.565 & $-0.055$ [$-0.111$, $+0.001$] \\
 & 128 & 0.722 & 0.661 & $-0.061$ [$-0.101$, $-0.025$] \\
\bottomrule
\end{tabular}
\end{table}

The selected multi-task model has the higher marginal likelihood in 77 of the 80 fits (median gain 13--26 log units), mostly through its correlated noise term: with independent noise, the gain is 7--12 log units on the JAREX campaigns and below zero late in the space-filling campaigns.
Its fitted noise correlations (median mean absolute value 0.2--0.7) exceed those present in the data, which we measured by evaluating the noise-free model at the observed inputs (0.13--0.28 between purity and cost, at most 0.02 in magnitude for pairs involving conversion).
The two surrogates would also select different experiments: of the ten candidates with the highest JAREX score in a fixed pool of 4,096 points, the multi-task model shares 21--41\%, and the rank correlation of the scores over the pool is 0.66--0.94.

\subsection{Warm Start from an Optimization Campaign}
\label{sec:si_warm_start}

To test whether data from an earlier optimization study help characterization, we started JAREX and space-filling sampling from 20 points of a multi-objective Bayesian optimization campaign (8 Latin hypercube points followed by 12 iterations of noisy expected hypervolume improvement, maximizing conversion and purity and minimizing cost) and, for comparison, from 20 Latin hypercube points.
Both starts were followed by 100 iterations (standard transformation, 10 paired seeds).
Of the optimization points, 83\% lie inside the true joint pass region, against 20\% of the Latin hypercube points.
The warm start changes neither the final accuracy nor the speed of convergence detectably (Table~\ref{tbl:si_warm_start}): points clustered near the optimum carry little information about where the joint boundary lies, so JAREX still has to locate it.
In this benchmark, reusing optimization data causes no statistically resolved loss in final accuracy, but does not shorten characterization.
This result does not guarantee a neutral or beneficial warm start for other processes or initial datasets.

\begin{table}[H]
\centering
\caption{Warm start from an optimization campaign versus Latin hypercube sampling (20 initial points, 100 iterations, 10 paired seeds; mean $\pm$ standard deviation). $\Delta$: paired difference in the final joint Jaccard index, warm start minus Latin hypercube start, with 95\% bootstrap confidence interval. Last column: mean number of experiments, including the initial points, at which the joint Jaccard index first reaches 0.8.}
\label{tbl:si_warm_start}
\footnotesize
\setlength{\tabcolsep}{4pt}
\begin{tabular}{llcccc}
\toprule
Method & Initial points & After 70 experiments & Final (120) & $\Delta$ final & Experiments to 0.8 \\
\midrule
JAREX & Latin hypercube & $0.813 \pm 0.030$ & $0.866 \pm 0.019$ & -- & 66 (10/10) \\
 & Optimization & $0.778 \pm 0.039$ & $0.875 \pm 0.009$ & $+0.008$ [$0.000$, $+0.018$] & 72 (10/10) \\
Space-filling & Latin hypercube & $0.543 \pm 0.073$ & $0.727 \pm 0.039$ & -- & -- \\
 & Optimization & $0.491 \pm 0.084$ & $0.700 \pm 0.047$ & $-0.027$ [$-0.059$, $+0.002$] & -- \\
\bottomrule
\end{tabular}
\end{table}

\subsection{Precision, Recall, and False-Positive Rate}
\label{sec:si_metrics}

Table~\ref{tbl:si_metrics} reports the joint confusion-matrix metrics of the final surrogates.
A condition is classified as passing at the mean level if the posterior mean of every objective clears its threshold, and at the 68\% and 95\% levels if the corresponding lower prediction bound does.
These levels apply to each objective separately.
The joint region intersects the per-objective regions and has no confidence level of its own.

\begin{table}[H]
\centering
\caption{Joint classification metrics of the final surrogates (10 seeds, mean). Experiments include the initial design. FPR: fraction of failing conditions predicted to pass.}
\label{tbl:si_metrics}
\small
\begin{tabular}{llccccc}
\toprule
Transformation & Method & Experiments & Jaccard & Precision & Recall & FPR \\
\midrule
Standard & JAREX, $q=1$ & 128 & 0.878 & 0.969 & 0.903 & 0.65\% \\
 & JAREX, $q=1$, 68\% level & 128 & 0.725 & 0.993 & 0.729 & 0.11\% \\
 & JAREX, $q=1$, 95\% level & 128 & 0.262 & 0.998 & 0.262 & 0.01\% \\
 & JAREX, $q=4$ & 328 & 0.923 & 0.970 & 0.950 & 0.67\% \\
 & JAREX, $q=8$ & 408 & 0.921 & 0.976 & 0.941 & 0.51\% \\
 & Greedy (three orders) & 128 & 0.655--0.731 & 0.84--0.87 & 0.73--0.83 & 2.4--3.3\% \\
 & Space-filling & 128 & 0.720 & 0.875 & 0.801 & 2.63\% \\
 & Central composite design & 27 & 0.356 & 0.717 & 0.415 & 3.69\% \\
Log & JAREX, $q=1$ & 128 & 0.880 & 0.964 & 0.911 & 0.78\% \\
 & JAREX, $q=1$, 68\% level & 128 & 0.804 & 0.992 & 0.809 & 0.15\% \\
 & JAREX, $q=1$, 95\% level & 128 & 0.526 & 0.999 & 0.526 & 0.01\% \\
 & Greedy (three orders) & 128 & 0.853--0.863 & 0.946--0.951 & 0.893--0.908 & 1.1--1.2\% \\
 & Space-filling & 128 & 0.808 & 0.948 & 0.846 & 1.05\% \\
 & Central composite design & 27 & 0.475 & 0.932 & 0.492 & 0.81\% \\
\bottomrule
\end{tabular}
\end{table}

At the 68\% level, no method other than JAREX yields a usable conservative region under the standard transformation: their recall is at most 4\%.

\begin{figure}[H]
  \centering
  \includegraphics[width=\textwidth]{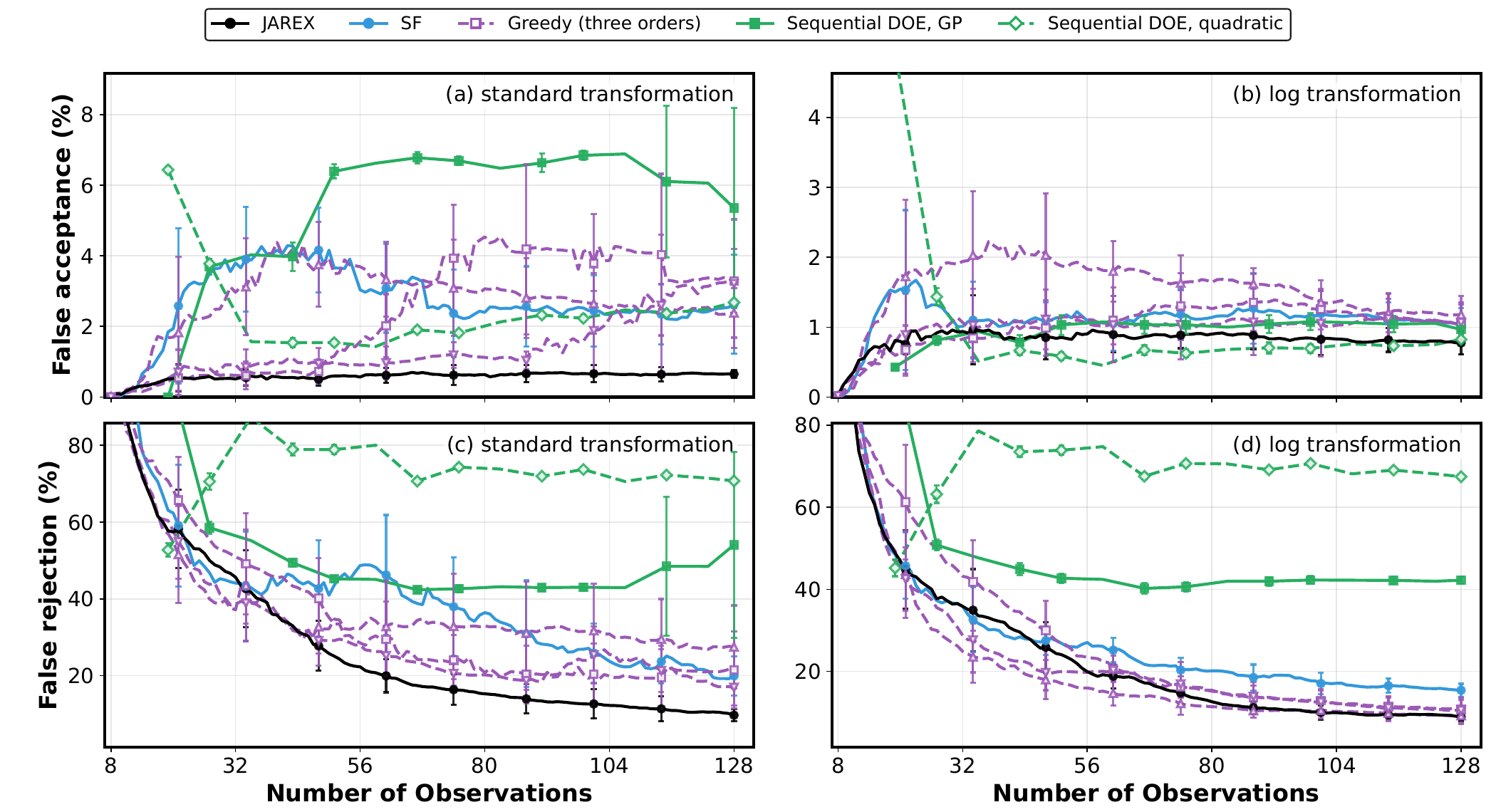}
  \caption{False acceptance and false rejection of the joint pass region over the campaign on the kinetic benchmark (default priors; mean of 10 campaigns, error bars $\pm 1\sigma$; each panel has its own vertical range).}
  \label{fig:si_kinetic_errors}
\end{figure}

Because every campaign records the confusion matrix at every step, the same metrics can be followed over the budget (Fig.~\ref{fig:si_kinetic_errors} and Table~\ref{tbl:si_kinetic_errors}).
Under the standard transformation, JAREX accepts 0.6--0.7\% of the failing conditions from 27 experiments onward, while the other methods accept 1.0--6.8\% from 51 experiments onward.
From 51 experiments onward, JAREX also rejects the fewest passing conditions, 9.7\% at 128 experiments against 19.9\% for space-filling sampling and 16.9--27.5\% for the greedy strategy.
Under the log transformation, all adaptive methods accept 0.8--1.9\% of the failing conditions, and JAREX and the greedy strategy reject passing conditions at similar rates.
The sequential design of Section~\ref{sec:si_baselines} rejects 40--79\% of the passing conditions throughout, whichever model scores it.
The corresponding curves for the chromatography case study are in Fig.~\ref{fig:si_pharma_errors_hplc}.

\begin{table}[H]
\centering
\caption{False acceptance, FP/(FP + TN), and false rejection, FN/(FN + TP), of the joint pass region (\%) on the kinetic benchmark after 27, 51, 75, 99, and 128 experiments (mean of 10 campaigns, mean level, default priors). Greedy: range over the three objective orders. Sequential DOE: the CCD built in stages and augmented with I-optimal blocks (Section~\ref{sec:si_baselines}), scored with the GP or with full quadratic models.}
\label{tbl:si_kinetic_errors}
\scriptsize
\setlength{\tabcolsep}{3pt}
\begin{tabular}{llcccccccccc}
\toprule
 & & \multicolumn{5}{c}{False acceptance (\%)} & \multicolumn{5}{c}{False rejection (\%)} \\
\cmidrule(lr){3-7}\cmidrule(lr){8-12}
Transformation & Method & 27 & 51 & 75 & 99 & 128 & 27 & 51 & 75 & 99 & 128 \\
\midrule
Standard & JAREX & 0.6 & 0.6 & 0.6 & 0.7 & 0.7 & 50.0 & 25.1 & 16.2 & 12.6 & 9.7 \\
 & Greedy (three orders) & 0.6--2.3 & 1.0--3.5 & 1.2--4.0 & 1.6--4.1 & 2.4--3.3 & 39.9--55.4 & 29.1--34.7 & 20.5--32.8 & 20.4--31.5 & 16.9--27.5 \\
 & Space-filling & 3.5 & 3.7 & 2.3 & 2.5 & 2.6 & 46.8 & 43.5 & 37.5 & 26.9 & 19.9 \\
 & Sequential DOE, GP & 3.7 & 6.4 & 6.7 & 6.8 & 5.4 & 58.5 & 45.2 & 42.6 & 43.0 & 54.1 \\
 & Sequential DOE, quadratic & 3.8 & 1.5 & 1.8 & 2.2 & 2.7 & 70.6 & 78.9 & 74.3 & 73.7 & 70.7 \\
Log & JAREX & 0.9 & 0.9 & 0.9 & 0.8 & 0.8 & 37.7 & 24.1 & 14.2 & 10.1 & 8.9 \\
 & Greedy (three orders) & 0.8--1.7 & 1.0--1.9 & 1.0--1.6 & 1.0--1.4 & 1.0--1.2 & 30.0--49.9 & 17.1--26.9 & 11.8--16.3 & 10.4--13.0 & 9.2--10.7 \\
 & Space-filling & 1.3 & 1.1 & 1.2 & 1.2 & 1.0 & 37.3 & 26.9 & 20.4 & 17.7 & 15.4 \\
 & Sequential DOE, GP & 0.8 & 1.0 & 1.0 & 1.1 & 1.0 & 50.8 & 42.7 & 40.6 & 42.3 & 42.2 \\
 & Sequential DOE, quadratic & 1.4 & 0.6 & 0.6 & 0.7 & 0.8 & 63.2 & 73.9 & 70.6 & 70.6 & 67.4 \\
\bottomrule
\end{tabular}
\end{table}

\subsection{Spatial Distribution of Classification Errors}
\label{sec:si_errors}

\begin{figure}[H]
  \centering
  \includegraphics[width=\textwidth]{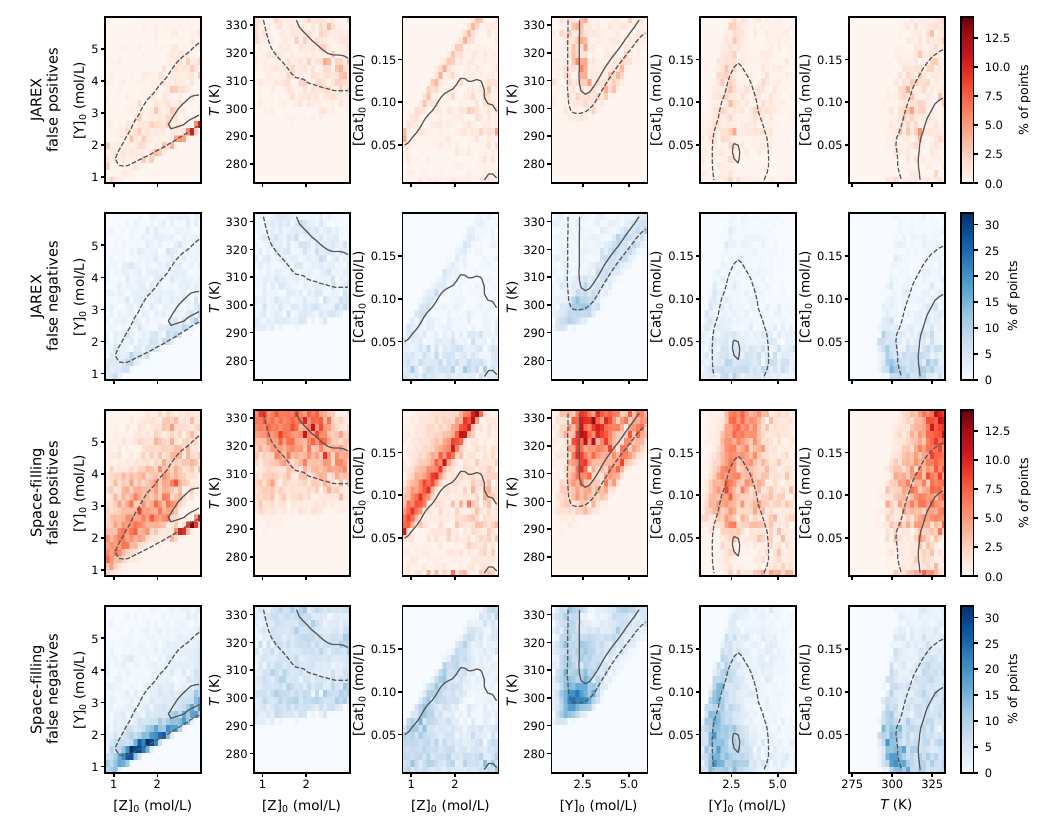}
  \caption{Location of the joint classification errors of the final surrogates on the kinetic benchmark (standard transformation, 128 experiments, 10 seeds); top two rows, JAREX; bottom two rows, space-filling sampling.
  Each panel projects the 32,768 points of the noise-free truth set onto one pair of inputs; color: share of the points in each bin that are false positives (red) or false negatives (blue), averaged over seeds, on a common scale for both methods.
  Gray contours: share of points in the true joint pass region (dashed, 25\%; solid, 50\%).}
  \label{fig:si_error_maps}
\end{figure}

We mapped the false positives and false negatives of the final serial JAREX surrogates (standard transformation) on the 32,768-point truth set and attributed each error to the misclassified objective(s), recording its distance from that objective's threshold in local noise standard deviations.
False rejections cover 1.8\% of the parameter space and false acceptances 0.53\%.
Purity is misclassified at 59\% of the false rejections, a median 7 standard deviations from its threshold, consistent with limited resolution of the steep purity response near 99.3\%.
Cost is misclassified at 40\% of the false acceptances, more often than any other objective, a median 0.49 standard deviations from its threshold (83\% within one), where 5\% measurement noise makes the classification hard to resolve with a finite number of noisy experiments.
By contrast, the false rejections of space-filling sampling caused by conversion lie a median 25 standard deviations from the threshold, in regions the design did not sample.
The noise-free runs of Section~\ref{sec:si_noise} test these interpretations (distances below in units of the 5\% noise standard deviation).
Without measurement noise, the false rejections of JAREX change little: they cover 1.5\% of the parameter space (2.2\% and 3.2\% with 10\% and 20\% noise), and purity causes 61\% of them, a median 6.5 standard deviations from its threshold.
The false acceptances stay at 0.57\% (0.60\% and 0.72\% with 10\% and 20\% noise), but the share caused by cost falls from 40\% to 28\% (48\% and 49\% with 10\% and 20\% noise).
The remaining cost-driven false acceptances lie a median 0.38 standard deviations from the threshold.
Measurement noise thus accounts for part of the cost-driven false acceptances, and more noise adds false acceptances, whereas without noise or with little of it the false rejections reflect the resolution of the purity response.
Figure~\ref{fig:si_error_maps} shows where these errors lie: those of JAREX are sparse and concentrated along the edges of the joint pass region, whereas space-filling sampling (2.1\% false acceptances, 3.7\% false rejections) also misclassifies larger areas, such as false rejections at low $[\mathrm{Z}]_0$ and $[\mathrm{Y}]_0$ and false acceptances at high temperature.

\subsection{Model-Based Baselines}
\label{sec:si_baselines}

As a classical active-learning baseline, we added uncertainty sampling, which selects the point of maximum posterior standard deviation (for several objectives, the sum of the standard deviations of the standardized objectives).
On the kinetic benchmark (128 experiments), JAREX has a higher final joint Jaccard index under both transformations: $0.878 \pm 0.015$ against $0.659 \pm 0.048$ with the standard transformation, and $0.880 \pm 0.010$ against $0.781 \pm 0.019$ with the log transformation.
Its joint false-positive rate is also lower (0.65\% against 3.01\% under the standard transformation).
On the single-objective benchmarks (Table~\ref{tbl:si_baselines_so}), randomized straddle outperforms uncertainty sampling on every function and noise level, with every paired confidence interval above zero except on the noise-free six-hump camel function ($+0.044$, 95\% CI $-0.011$ to $+0.099$).
Uncertainty sampling falls below space-filling sampling on every function, because it concentrates experiments at high-variance corners of the domain.

\begin{figure}[H]
  \centering
  \includegraphics[width=\textwidth]{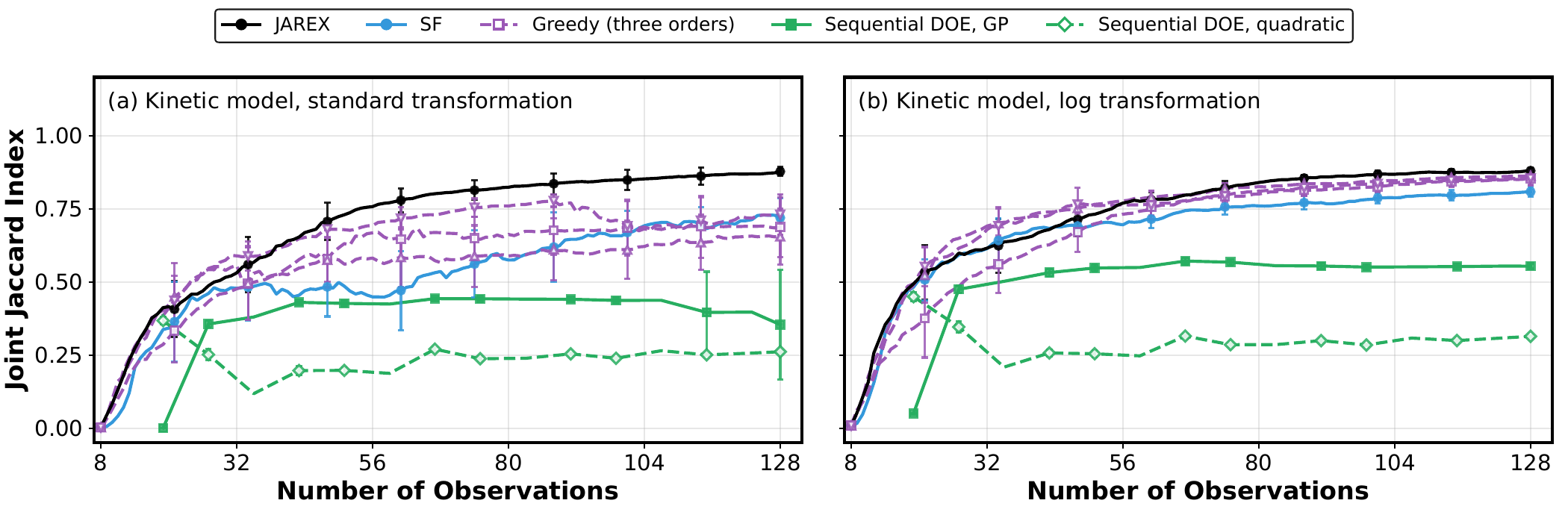}
  \caption{Sequential design (CCD built in stages, then I-optimal blocks of eight up to 128 runs) on the kinetic benchmark, scored with the GP surrogate (default priors) and with full quadratic models, against JAREX, space-filling sampling, and greedy randomized straddle (three objective orders); mean of 10 campaigns, error bars $\pm 1\sigma$. The two-level factorial (19 runs) and CCD (27 runs) are the first two points of the sequential design. Under the standard transformation, the GP fit fails late in the sequence in two campaigns, which lowers the mean and widens the error bars at 115 and 128 runs.}
  \label{fig:si_seqdoe}
\end{figure}

\paragraph*{Sequential, model-based design of experiments.}
As the conventional sequential alternative, we built the design a practitioner would use for the same task.
The CCD of the main text is built in stages, first the two-level factorial with three center runs (19 runs) and then the axial points (27 runs), and is then augmented in blocks of eight runs, each I-optimal for a full quadratic model given all runs before it, up to 128 runs (the last block has five), as in the design-augmentation step of standard DOE software.
An I-optimal design minimizes the average prediction variance of the model over the parameter space.~\cite{Goos2011}
We computed the blocks by coordinate exchange (pyoptex 1.2.1, best of 10 random starts) on the coded unit cube and fixed every block in advance, so the sequence does not depend on the data.
The average prediction variance of the quadratic model (per unit noise variance) falls from 8.53 with the CCD to 7.57 after one block and 7.19 at 128 runs, within 0.3\% of a 128-run I-optimal design built in one step (7.17).
Like the other methods, the design was evaluated in 10 campaigns with the same measurement noise and scored at every block, both with the GP surrogate used for the other methods and with least-squares full quadratic models, as in response surface methodology, under both transformations (Table~\ref{tbl:si_seqdoe} and Fig.~\ref{fig:si_seqdoe}).

Scored with the GP, the joint Jaccard index rises to 0.43--0.44 by 51--75 runs under the standard transformation and then stays flat.
In two of the ten campaigns, the GP fit fails at 115 or 128 runs (joint Jaccard index 0), which lowers the mean at 128 runs to 0.354 (median 0.439).
Under the log transformation, it levels off at 0.55.
Full quadratic models reach 0.26 and 0.31 at 128 runs.
With the 19 runs of the two-level stage, where the four pure quadratic terms cannot be separated and the regression shrinks them toward a common curvature, they score higher (0.37 and 0.45) than after the axial points are added.
At 128 experiments, every variant is below space-filling sampling (0.720 and 0.808), the tested blockwise greedy strategies (0.655--0.731 and 0.853--0.863), and JAREX (0.878 and 0.880).
An I-optimal design for a quadratic places its runs on three levels per input, here 81 distinct points at most, so it leaves the interior of the parameter space, where the boundaries of the kinetic model lie, almost unsampled.
On the pharmaceutical case studies, fixed 128-run I-optimal designs (Section~\ref{sec:si_benchmark_b}) likewise score within 0.07 of the CCD where both were run.
On the chromatography case, they are below space-filling sampling in seven of the nine combinations (paired differences $-0.205$ to $-0.012$, with intervals below zero), not resolved from it with two objectives at 13 inputs and four at 9, and less accurate than JAREX in every combination (by 0.12 to 0.28).
On the crystallization case, they are below space-filling sampling at 3 and 6 inputs, ahead of it with two and three objectives at 9 inputs (paired differences $+0.039$, 95\% CI $+0.009$ to $+0.075$, and $+0.074$, $+0.015$ to $+0.136$), and not resolved from it in the other four combinations.
JAREX is more accurate than the I-optimal design in every combination (by 0.019 to 0.271).

\begin{table}[H]
\centering
\caption{Joint Jaccard index of the sequential design (CCD built in stages, then augmented with I-optimal blocks of eight) on the kinetic benchmark, by number of runs (mean of 10 campaigns), scored with the GP surrogate of the other methods (default priors) or with full quadratic models. Reference at 128 experiments: JAREX 0.878 (standard) and 0.880 (log); space-filling sampling 0.720 and 0.808; greedy randomized straddle 0.655--0.731 and 0.853--0.863.}
\label{tbl:si_seqdoe}
\small
\begin{tabular}{llccccccc}
\toprule
Transformation & Scoring & 19 & 27 & 35 & 51 & 75 & 99 & 128 \\
\midrule
Standard & GP & 0.001 & 0.356 & 0.380 & 0.427 & 0.443 & 0.437 & 0.354$^{a}$ \\
 & Quadratic & 0.368 & 0.252 & 0.118 & 0.198 & 0.238 & 0.239 & 0.262 \\
Log & GP & 0.050 & 0.475 & 0.502 & 0.548 & 0.568 & 0.551 & 0.554 \\
 & Quadratic & 0.450 & 0.346 & 0.209 & 0.255 & 0.286 & 0.285 & 0.314 \\
\bottomrule
\multicolumn{9}{l}{\footnotesize $^{a}$ Median 0.439.
  In two campaigns, the GP fit fails at 115 or 128 runs.} \\
\end{tabular}
\end{table}

\begin{table}[H]
\centering
\caption{Final Jaccard index on the single-objective benchmarks (10 seeds, paired with the runs of the main text; mean $\pm$ standard deviation).}
\label{tbl:si_baselines_so}
\footnotesize
\setlength{\tabcolsep}{4pt}
\begin{tabular}{llcccc}
\toprule
Function & Noise & Rand.\ straddle & Uncertainty & FEI & Space-filling \\
\midrule
Six-hump camel & 0\% & $0.944 \pm 0.074$ & $0.900 \pm 0.044$ & $0.985 \pm 0.006$ & $0.927 \pm 0.013$ \\
 & 5\% & $0.701 \pm 0.063$ & $0.580 \pm 0.054$ & $0.696 \pm 0.047$ & $0.678 \pm 0.051$ \\
Hartmann 4D & 0\% & $0.985 \pm 0.013$ & $0.633 \pm 0.122$ & $0.958 \pm 0.012$ & $0.819 \pm 0.037$ \\
 & 5\% & $0.920 \pm 0.026$ & $0.548 \pm 0.193$ & $0.910 \pm 0.009$ & $0.806 \pm 0.038$ \\
Hartmann 6D & 0\% & $0.837 \pm 0.009$ & $0.461 \pm 0.127$ & $0.801 \pm 0.024$ & $0.672 \pm 0.050$ \\
 & 5\% & $0.759 \pm 0.011$ & $0.311 \pm 0.142$ & $0.725 \pm 0.032$ & $0.639 \pm 0.036$ \\
\bottomrule
\end{tabular}
\end{table}

\bibliographystyle{unsrtnat}
\bibliography{ref}
\end{document}